\documentclass[english, 5pt]{elsarticle}
\usepackage[T1]{fontenc}
\usepackage[latin9]{inputenc}
\usepackage[a4paper]{geometry}
\usepackage{float}
\usepackage{textcomp}
\usepackage{amsthm}
\usepackage{amsmath}
\usepackage{amssymb}
\usepackage{algorithm}
\usepackage{algpseudocode}
\usepackage{csquotes}
\usepackage{multirow}
\usepackage{booktabs}
\usepackage{color}
\usepackage{epstopdf}
\usepackage{graphicx}
\usepackage{subfigure}
\usepackage{bm}
\usepackage{multirow}
\usepackage{multicol}
\usepackage{booktabs}
\usepackage{color}
\usepackage{epstopdf}
\usepackage[colorlinks,linkcolor=red]{hyperref}
\usepackage{appendix}

\algnewcommand\algorithmicforeach{\textbf{for each}}
\algdef{S}[FOR]{ForEach}[1]{\algorithmicforeach\ #1\ \algorithmicdo}

\begin{document}

\begin{frontmatter}

\title{Solve routing problems with a residual edge-graph attention neural network}

\author[mymainaddress]{Kun LEI}

\author[mymainaddress,mysecondaryaddress]{Peng GUO\corref{mycorrespondingauthor}}
\cortext[mycorrespondingauthor]{Corresponding author}
\ead{pengguo318@swjtu.edu.cn}

\author[mythirdaddress]{Yi WANG}

\author[mymainaddress,mysecondaryaddress]{Xiao WU}
	
\author[mymainaddress]{Wenchao ZHAO}

\address[mymainaddress]{Department of Industrial Engineering, School of Mechanical Engineering, Southwest Jiaotong University, Chengdu, 610031 China}
\address[mysecondaryaddress]{Technology and Equipment of Rail Transit Operation and Maintenance Key Laboratory of Sichuan Province, Chengdu, 610031 China}
\address[mythirdaddress]{Department of Mathematics and Computer Science, Auburn University at Montgomery, Montgomery, AL 36124-4023 USA}

\begin{abstract}
For NP-hard combinatorial optimization problems, it is usually difficult to find high-quality solutions in polynomial time. The design of either an exact algorithm or an approximate algorithm for these problems often requires significantly specialized knowledge. Recently, deep learning methods provide new directions to solve such problems. In this paper, an end-to-end deep reinforcement learning framework is proposed to solve this type of combinatorial optimization problems. This framework can be applied to different problems with only slight changes of input (for example, for a traveling salesman problem (TSP), the input is the two-dimensional coordinates of nodes; while for a capacity-constrained vehicle routing problem (CVRP), the input is simply changed to three-dimensional vectors including the two-dimensional coordinates and the customer demands of nodes), masks and decoder context vectors. The proposed framework is aiming to improve the models in literacy in terms of the neural network model and the training algorithm. The solution quality of TSP and the CVRP up to 100 nodes are significantly improved via our framework. Specifically, the average optimality gap is reduced from 4.53\% (reported best \cite{R22}) to 3.67\% for TSP with 100 nodes and from 7.34\% (reported best \cite{R22}) to 6.68\% for CVRP with 100 nodes when using the greedy decoding strategy. Furthermore, our framework uses about 1/3$\sim$3/4 training samples compared with other existing learning methods while achieving better results. The results performed on randomly generated instances and the benchmark instances from TSPLIB and CVRPLIB confirm that our framework has a linear running time on the problem size (number of nodes) during the testing phase, and has a good generalization performance from random instance training to real-world instance testing.

\bigskip

{\bf Keywords:}
	Combinatorial optimization, Deep reinforcement learning, Residual edge-graph attention model, Routing problems

\end{abstract}

\end{frontmatter}

\section{Introduction}
Combinatorial optimization problems as basic problems in computer science and operations research have received extensive attention from the theory and algorithm design communities in the past few decades. TSP and vehicle routing problem (VRP) are two representatives of classic combinatorial optimization problems, and have been studied in the fields of logistics transportation, genetics, express delivery, and dispatching \cite{R1,R2,R3,R4}. Generally, TSP is defined on a graph with a number of nodes, and it is necessary to search among the permutation sequences of nodes for finding an optimal one with the shortest traveling distance. CVRP is a basic variant of VRP, which finds the route with the lowest cost while not violating vehicle capacity constraints and meeting all customer needs \cite{R5}. Even in the case of two-dimensional Euclid, it is difficult to find an optimal solution for a TSP or CVRP due to their intractability of NP-hardness \cite{R6}. In general, such a NP-hard problem can be expressed as sequential decision tasks on a graph due to its highly structured nature \cite{R7}.

Traditional methods for solving graph optimization problems with NP-hardness include exact algorithms, approximate algorithms, and heuristic algorithms \cite{R8}. Exact algorithms with the branch and bound framework can obtain optimal solutions, but they are not suitable for large-sized problems due to their NP-hardness. Polynomial-time approximation algorithms can usually obtain quality-guaranteed solutions, but they possess weaker optimality warrants compared to exact algorithms. In particular, for problems that are not amenable to a polynomial approximation algorithm, the optimality guarantee may not exist at all. In addition, heuristic algorithms are widely used due to their good computational performance, but usually require customizations and domain expertise knowledge for a specific problem. Besides, heuristic algorithms often lack theoretical support. All the above three groups of algorithms seldomly take advantage of the common features among optimization problems, and thus often need to design a new algorithm to solve a different instance of an even similar problem that is based on the same combinatorial structure, of which the coefficient values in the objective function or constraints may be deemed as samples from the same basic distributions \cite{R9}. The idea of using machine learning approaches has cast a silver lining to provide a scalable method to solve combinatorial problems with similar combinatorial structures.

\begin{figure}[htbp]
	\centering
	\includegraphics[width=5in]{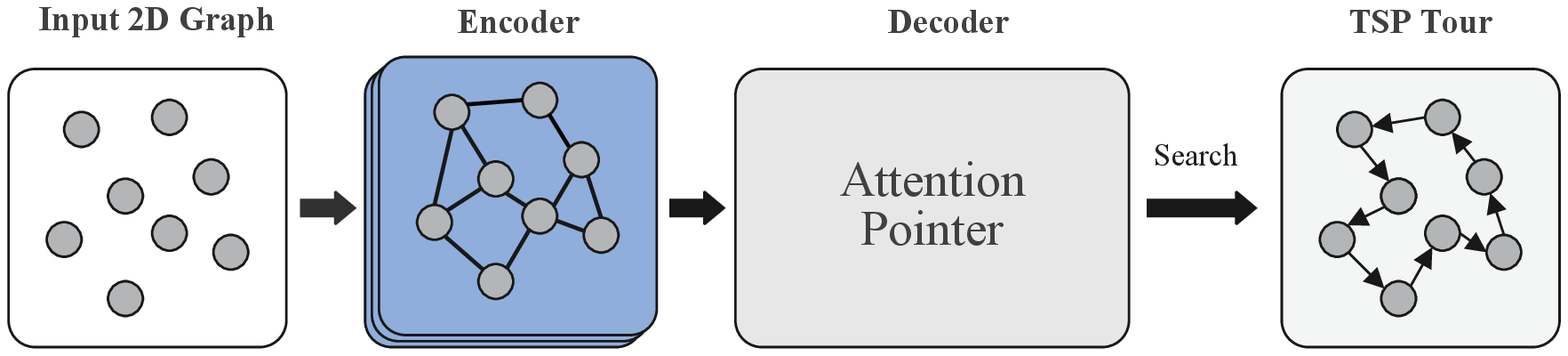}
	\caption{An end-to-end model for solving a TSP.}
	\label{fig:f1}
\end{figure}

Many combinatorial optimization problems, such as a TSP or a VRP,  are based on a graph structure \cite{R10}, which can be easily modeled by the existing graph embedding or network embedding technique.  In such a technique, the graph information is embedded in a continuous node representation. The latest development of graph neural network (GNN) can be used in modeling a graph combinatorial problem due to its strong capabilities in information embedding and belief propagation of graph topology \cite{R11}. This motivates us to adopt a GNN model to solve combinatorial optimization problems, particularly TSP and CVRP. In this paper, we use the GNN model to build an end-to-end deep reinforcement learning (DRL) framework, as schematically shown in Figure \ref{fig:f1}.

In our framework, after the features (node coordinates for TSP) of a 2D graph is entered into the model, the encoder encodes the features with GNN. The encoded features are then passed into the decoder with an attention pointer mechanism to predict the probabilities of unselected nodes. The next node is subsequently selected according to the probability distribution by a search strategy, such as a greedy search or a sampling approach. \textit{Our encoder amends the Graph Attention Network (GAT) \cite{R12} by taking into consideration \textbf{the edge information in the graph structure and residual connections between layers}.} We shall call the designed network a residual edge-graph attention network (residual E-GAT). Our decoder is designed based on a Transformer model, which is used primarily in the field of natural language processing \cite{R39}. The entire network is optimized using either a proximal policy optimization algorithm (PPO) or an improved baseline REINGORCE algorithm.

For demonstrating the performance of the proposed framework, a serial of random instances of TSP and CVRP with nodes of 20, 50, and 100 are used to train and test our algorithm. Besides, the complexity of the model running time during the testing phase is analyzed. To have a fair comparison to the current literature, we have tested the generalization ability of the framework using the standard benchmark instances from TSPLIB and CVRPLIB.

The main contributions of this paper can be summarized below:
\begin{itemize}
	\item An improved DRL framework is proposed for the routing problems. In this framework, a residual E-GAT is introduced for the encoder in which, are taken into consideration the edge information and residual connections between layers. The edge information and residual connections are not considered in GAT in the literature. It is demonstrated that the residual E-GAT is powerful to capture information of graph structures directly. In addition, the transformer model is used to design the decoder.
	\item In the training phase, we adopted two actor-critic algorithms: PPO and improved baseline REINGORCE algorithm. Some techniques of "code-level optimization" are used for further improving the performance of the two algorithms.
	\item The proposed algorithm is efficient and has strong generalization power. The efficiency of the proposed framework is evaluated on randomly generated instance datasets of TSP and CVRP.  Besides, the standard benchmark instances from TSPLIB and CVRPLIB are used to verify the generalization capacity of our framework.
\end{itemize}

The rest of the paper is organized as follows. Section \ref{se:se2} summarizes the relevant literature. Section \ref{se:se3} describes our encoder-decoder model in detail. Section \ref{se:se4} introduces two deep reinforcement learning algorithms and decoding strategies. Section \ref{se:se5} gives the experimental results and discussions are made. Finally, conclusions and prospects are listed in Section \ref{se:se6}. \ref{app:a} analyzes the sensitivities of hyper-parameters. \ref{app:b} shows solution examples of random instances with 100 nodes. \ref{app:c} demonstrates examples of standard benchmark instances.

\section{Literature Review}
\label{se:se2}
Recently, supervised learning (SL) and reinforcement (RL) learning have been widely used to solve combinatorial optimization problems \cite{R10,R13,R14,R15,R16}, especially in routing problems. Table \ref{tab:tab1} summarizes various existing methods for solving routing problems based on supervised learning and reinforcement learning. Reinforcement learning can be further divided into model-based and model-free methods. Finally, model-free reinforcement learning methods can be divided into Value-Based and Policy-Based methods or a combination of both (actor-critic).
\begin{table*}[htbp]
	\centering
\tiny
	\caption{Survey of deep/reinforcement learning methods in solving routing problems}
	\begin{tabular}{llll}
		\toprule
		Routing Problem & \multicolumn{1}{l}{Authors} & \multicolumn{1}{l}{Network Structure} & \multicolumn{1}{l}{Learning Type} \\
		\midrule
		\multirow{15}[2]{*}{TSP} & \multicolumn{1}{l}{[Vinyals et al.,2015]\cite{R17} } & \multicolumn{1}{l}{LSTM+Attention} & \multicolumn{1}{l}{Supervised Learning, Approximation} \\
		\multicolumn{1}{l}{} & \multicolumn{1}{l}{[Bello et al.,2016]\cite{R18}} & \multicolumn{1}{l}{LSTM+Attention } & \multicolumn{1}{l}{RL-Model Free, Actor-Critic} \\
		\multicolumn{1}{l}{} & \multicolumn{1}{l}{[Joshi et al.,2019]\cite{R7}} & \multicolumn{1}{l}{GCN} & \multicolumn{1}{l}{Supervised Learning, Approximation} \\
		\multicolumn{1}{l}{} & \multicolumn{1}{l}{[Dai et al.,2017]\cite{R9}} & \multicolumn{1}{l}{GNN} & \multicolumn{1}{l}{RL-Model Free, Deep Q-Learning} \\
		\multicolumn{1}{l}{} & \multicolumn{1}{l}{[Nazari et al.,2018]\cite{R19}} & \multicolumn{1}{l}{LSTM+Attention } & \multicolumn{1}{l}{RL-Model Free, Actor-Critic} \\
		\multicolumn{1}{l}{} & \multicolumn{1}{l}{[Deudon et al.,2018]\cite{R20}} & \multicolumn{1}{l}{GRU} & \multicolumn{1}{l}{RL-Model Free, Actor-Critic} \\
		\multicolumn{1}{l}{} & \multicolumn{1}{l}{[Emami et al.,2018]\cite{R21}} & \multicolumn{1}{l}{Transformer} & \multicolumn{1}{l}{RL-Model Free, Actor-Critic} \\
		\multicolumn{1}{l}{} & \multicolumn{1}{l}{[Kool et al.,2019]\cite{R22}} & \multicolumn{1}{l}{Transformer} & \multicolumn{1}{l}{RL-Model Free, Actor-Critic} \\
		\multicolumn{1}{l}{} & \multicolumn{1}{l}{[Malazgirt et al.,2019]\cite{R23}} & \multicolumn{1}{l}{NN} & \multicolumn{1}{l}{RL-Model Free, Policy-Based} \\
		\multicolumn{1}{l}{} & \multicolumn{1}{l}{[Ma et al.,2020]\cite{R24}} & \multicolumn{1}{l}{GPN} & \multicolumn{1}{l}{Hierarchical RL-Model Free, Policy-Based} \\
		\multicolumn{1}{l}{} & \multicolumn{1}{l}{[Cappart et al.,2020]\cite{R25}} & \multicolumn{1}{l}{GAT} & \multicolumn{1}{l}{RL-Model Free, Actor-Critic} \\
		\multicolumn{1}{l}{} & \multicolumn{1}{l}{[P. Felix et al.,2020]\cite{R26}} & \multicolumn{1}{l}{GCN} & \multicolumn{1}{l}{RL-Model Based, Given Model} \\
		\multicolumn{1}{l}{} & \multicolumn{1}{l}{[Drori et al.,2020]\cite{R27}} & \multicolumn{1}{l}{GAT+ Attention } & \multicolumn{1}{l}{RL-Model Free, Actor-Critic} \\
		\multicolumn{1}{l}{} & \multicolumn{1}{l}{[Zhang et al.,2020]\cite{R28}} & \multicolumn{1}{l}{Transformer} & \multicolumn{1}{l}{RL-Model Free, Policy-Based} \\
		\multicolumn{1}{l}{} & \multicolumn{1}{l}{[Hu et al., 2020]\cite{R29}} & \multicolumn{1}{l}{GNN} & \multicolumn{1}{l}{RL-Model Free, Policy-Based} \\
		\midrule
		\multirow{9}[2]{*}{VRP} & \multicolumn{1}{l}{[Nazari et al.,2018]\cite{R19}} & \multicolumn{1}{l}{LSTM+Attention } & \multicolumn{1}{l}{RL-Model Free, Actor-Critic} \\
		\multicolumn{1}{l}{} & \multicolumn{1}{l}{[Chen et al.,2019]\cite{R30}} & \multicolumn{1}{l}{LSTM} & \multicolumn{1}{l}{RL-Model Free, Actor-Critic} \\
		\multicolumn{1}{l}{} & \multicolumn{1}{l}{[Kool et al.,2019]\cite{R22}} & \multicolumn{1}{l}{Transformer} & \multicolumn{1}{l}{RL-Model Free, Actor-Critic} \\
		\multicolumn{1}{l}{} & \multicolumn{1}{l}{[Zhao et al. 2020]\cite{R31}} & \multicolumn{1}{l}{LSTM+Attention} & \multicolumn{1}{l}{RL-Model Free, Actor-Critic} \\
		\multicolumn{1}{l}{} & \multicolumn{1}{l}{[Lu et al.,2020]\cite{R32}} & \multicolumn{1}{l}{NN} & \multicolumn{1}{l}{RL-Model Free, Actor-Critic} \\
		\multicolumn{1}{l}{} & \multicolumn{1}{l}{[Gao et al.,2020]\cite{R33}} & \multicolumn{1}{l}{GAT+GRU} & \multicolumn{1}{l}{RL-Model Free, Actor-Critic} \\
		\multicolumn{1}{l}{} & \multicolumn{1}{l}{[Drori et al.,2020]\cite{R27}} & \multicolumn{1}{l}{GAT+ Attention} & \multicolumn{1}{l}{RL-Model Free, Actor-Critic} \\
		\multicolumn{1}{l}{} & \multicolumn{1}{l}{[Chen et al.,2020]\cite{R11}} & \multicolumn{1}{l}{GAT+GRU} & \multicolumn{1}{l}{RL-Model Free, Actor-Critic} \\
		\multicolumn{1}{l}{} & \multicolumn{1}{l}{[zhang et al., 2020]\cite{R34}} & \multicolumn{1}{l}{Transformer} & \multicolumn{1}{l}{RL-Model Free, Actor-Critic} \\
		\bottomrule
		\multicolumn{4}{p{360pt}}{Note: In the Network Structure column, LSTM: long short-term memory; NN: neural networks; GRU: gate recurrent unit; GPN: graph pointer network.} 	
	\end{tabular}%
	\label{tab:tab1}%
\end{table*}%

Vinyals et al. \cite{R17} introduced a supervised learning framework with sequence-to-sequence pointer network (PtrNet) to train and solve a Euclidean TSP and other combinatorial optimization problems. Their model uses the softmax probability distribution as a 'pointer' to select a member from the input sequence as the output. Bello et al. \cite{R18} introduced an actor-critic reinforcement learning algorithm to train PtrNet to solve the TSP in an unsupervised manner. They considered each instance as a training sample and used the cost (tour length) of a sampled solution for an unbiased Monte-Carlo estimate of the policy gradient, and showed the performance of their algorithms on TSP with up to 100 nodes is better than most previous approximate algorithms. Nazari et al. \cite{R19} extended the structure of PtrNet to solve more complex combinatorial optimization problems, such as the VRP with batch delivery and random variables. Gu et al. \cite{R35} used the PtrNet combined with supervised learning and DRL to solve an unconstrained binary quadratic programming problem. \textit{However, the neural network structure used in the above works does not fully consider the relationship of the edges between vertexes in a graph, which is actually very important for many routing problems.}

GNN as a powerful tool for processing non-Euclidean data and capturing graphical information has been widely researched in recent years. Specifically, once a GNN-based approximate solver has been trained, its time complexity to solve a problem is significantly better than that of an OR algorithm. In this case, the trained GNN is very suitable for real-time decision-making problems such as TSP and related vehicle routing problems. Li et al. \cite{R36} applied graph convolutional network (GCN) model proposed by Kipf et al. \cite{R37} and guided tree search algorithm to solve graph-based combinatorial optimization problems, such as maximum independent set and minimum vertex cover problems. Dai et al. \cite{R9} used GNN to encode a problem instance and demonstrated a GNN preserves node order and performs better in reflecting the combinatorial structure of TSP, when compared with the sequence-to-sequence model. They used deep Q-networks (DQN) to train a structure2vec (S2V) graph embedding model \cite{R38}. Motivated by the Transformer architecture \cite{R39}, Kool et al. \cite{R22} proposed an attention model (AM) to solve a serial of combinatorial optimization problems, and used rollout baseline in the policy gradient algorithm to significantly improve the results of small-sized routing problems. Nowak et al. \cite{R40} used deep GCN in a supervised learning manner to construct an effective TSP graph representation and output itineraries through a highly parallel beam search using non-autoregressive approaches. Ma et al. \cite{R24} proposed a Graph Pointer Network (GPN), which trains a network to solve TSP with time window constraints (TSPTW) through hierarchical reinforcement learning. They claimed that the model can be extended from small-sized problems to large-sized problems. Drori et al. \cite{R27} used the GAT to solve many graph combinatorial optimization problems. Their results showed that the GAT framework has a generalization performance from training on random graphs to testing on a real-world graph. Hu et al. \cite{R41} introduced a bidirectional GNN trained by imitation learning to solve an arbitrary symmetric TSP.

\textit{The previous works used GNN and reinforcement learning to solve combinatorial optimization problems without considering the dependency between edges in a graph structure. Motivated by the above researches, our work considers the role of edges in a graph structure and considers connecting residuals between layers to remedy the model degradation due to vanishing gradients in a deep model to further improve the GAT. Besides, most existing methods combine GNN, search methods, and reinforcement learning. Our model is based on an encoder built-upon the improved GAT and a decoder built-upon the Transformer model. The proposed framework uses two deep reinforcement learning algorithms in training, i.e., PPO and improved baseline REINFORCE algorithm. These differences make the proposed model different from the existing ones.}

\section{Graph-attention model}
\label{se:se3}
In this section, we will formally introduce our residual edge-graph-attention model (Residual E-GAT). We define the model through a 2D Euclidean and symmetric TSP. The model is also applicable to other
 graph-based routing problems, and only needs to change accordingly the input, masks and decoder context vectors. TSP is defined on an undirected graph $G=(V,E,W)$  where node $i$ is represented by features
  $\bm{n}_i$, $i\in V=\{1,\cdots,m\}$. Here $m$ is the number of nodes and $\bm{n}_i$ denotes the coordinates of node
   $i$. And $a_{ij}\in E,i,j\in V$ represents the edge from node $i$ to node $j$, further $e_{ij}\in W$ is
    the distance information of $a_{ij}$. Solution $\hat{\pi}=(\hat{\pi}_1,\cdots,\hat{\pi}_m$) is introduced to
     express a permutation of all the nodes, $\hat{\pi}_t\in \{1,\cdots,m\}$ and $\hat{\pi}_t\neq \hat{\pi}_{t'}$,
      $\forall t\neq t'$.
Our goal is to find a solution $\hat{\pi}$ given a problem instance $s$ so that each node can be visited exactly once and the total tour length is minimized. The length of a tour is defined for permutation $\hat{\pi}$  as:
\begin{equation}
L(\hat{\pi}|s)=||\bm{n}_{\hat{\pi}_m}-\bm{n}_{\hat{\pi}_1} ||_2+\sum_{t=1}^{m-1}||\bm{n}_{\hat{\pi}_t}-\bm{n}_{\hat{\pi}_{t+1}}||_2,\label{eq:eq1}
\end{equation}
where $||\cdot||_2$ denotes the L2 norm. Our graph-attention model defines a stochastic policy $p(\hat{\pi}|s)$  for instance $s$. Based on the chain rule of probability, the selection probability for sequence $\hat{\pi}$ can be calculated based on a parameter set $\theta$ of the graph-attention model:
\begin{equation}
	p_{\theta} (\hat{\pi}|s)=\prod_{t=1}^{m}p_{\theta}(\hat{\pi}_t|s,\hat{\pi}_{t'}\;\forall t'<t). \label{eq:eq2}
\end{equation}
The encoder makes embeddings of all input nodes. The decoder produces a permutation $\hat{\pi}$ of input nodes by generating a node at each time step and masks (will be introduced in Section \ref{se:3.2}) that node out to prevent the model from visiting the node again.

Our encoder is designed based on the GAT, which is a neural network architecture that transmits node information through an attention mechanism and has a powerful graph topology representation capability. For the TSP problem with an undirected graph $G=(V,E,W)$, GAT \cite{R12} only updates the information of each node by assigning new weights to its neighbors. However, this update method ignores the information of the edge in the graph structure. Inspired by the work of Gao et al. \cite{R33}, the residual E-GAT proposed in this paper integrates the edge information into the node information and updates them simultaneously, instead of only updating the node information in the original GAT. In addition, each sub-layer of E-GAT adds a residual connection for avoiding vanishing gradient and model degradation \cite{R42}. The decoder is based on an attention mechanism, in which the next node is selected through either sampling or a greedy decoding. We will refine the encoder-decoder model in detail below.

\subsection{The encoder}
\label{se:se3.1}
\begin{figure}[htbp]
	\centering
	\subfigure[Edge and node fusion and update method.]{
		\label{fig:fig2_a}
		\includegraphics[width=7cm]{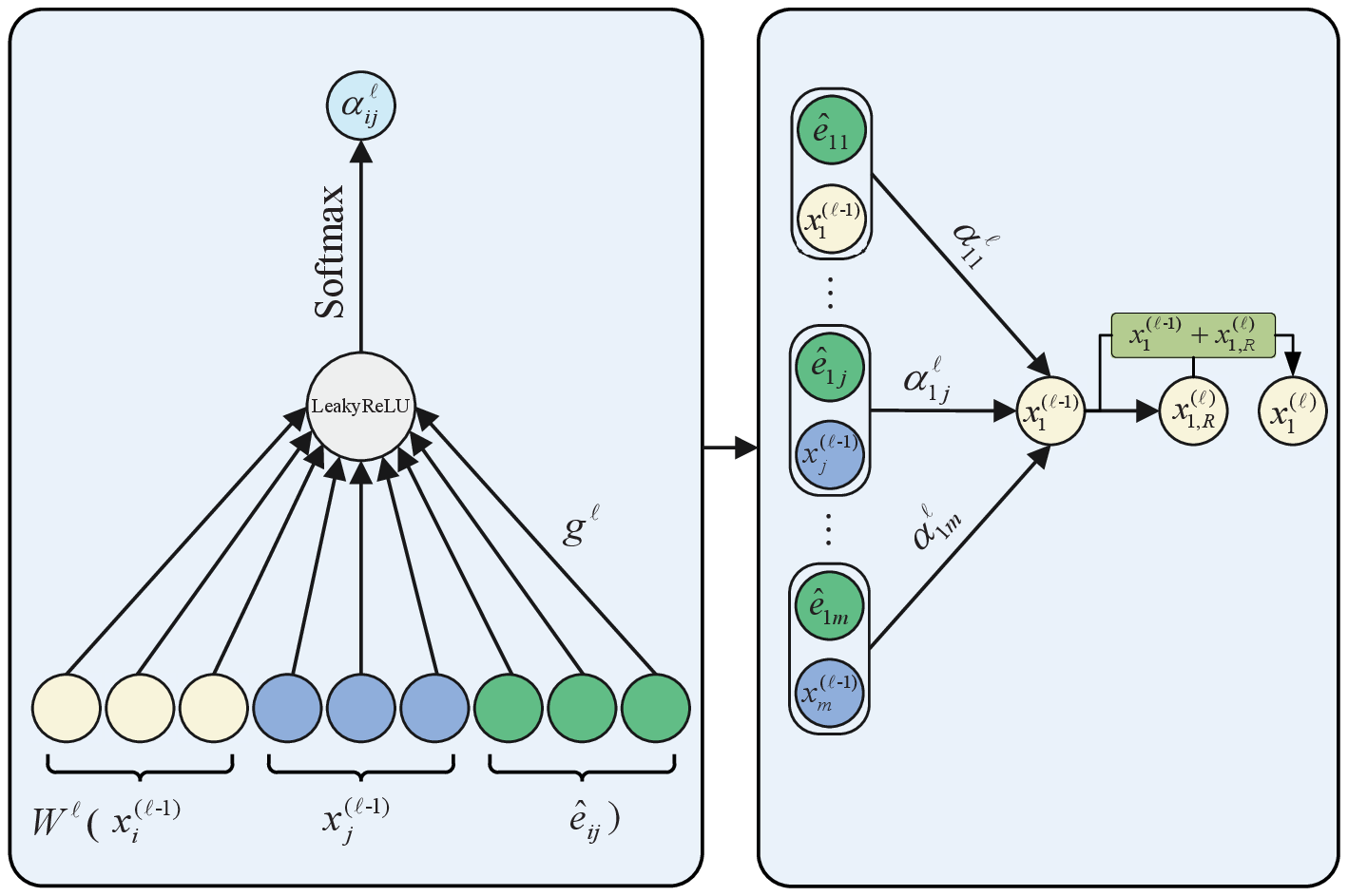}
	}
	\quad
	\subfigure[Encoder network structure.]{
		\label{fig:fig2_b}
		\includegraphics[width=4.75cm]{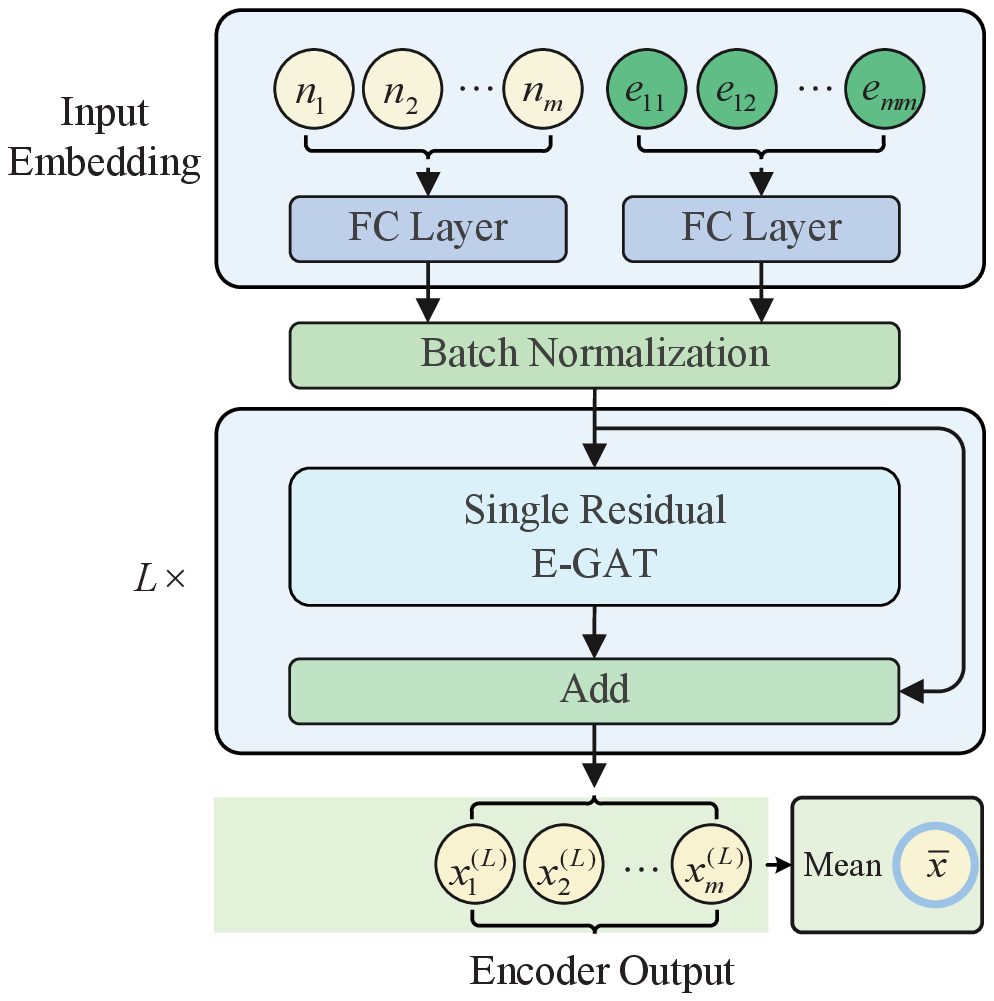}
	}
	\caption{The Encoder structure with multi-layer residual E-GAT.}
	\label{fig:fig2}
\end{figure}

Our encoder model takes a graph $G=(V,E,W)$ as input, as shown in Figure \ref{fig:fig2}. Taking TSP as an example, the input node features are the two-dimensional coordinates $\bm{n}_i$, and the input edge features are the Euclidean distances $e_{ij},i,j\in \{1,\cdots,m\}$.

The above two features are embedded in $d_x$ and $d_e$ dimension features through a fully connected layer (FC Layer in Figure \ref{fig:fig2_b}) before they are fed into the residual E-GAT. Equation \ref{eq:eq3} and Equation \ref{eq:eq4} below describe the node and edge embeddings respectively, where, the superscript 0 indicates that the embedding is before entering the residual E-GAT.
\begin{equation}
	\bm{x}_i^{(0) }=BN(\bm{A}_0 \bm{n}_i+\bm{b}_0 ), \forall i\in \{1,\cdots,m\}, \label{eq:eq3}
\end{equation}
\begin{equation}
	\hat{\bm{{e}}}_{ij}=BN(\bm{A}_{1}e_{ij}+\bm{b}_{1}), \forall i,j \in \{1,\cdots,m\}, \label{eq:eq4}
\end{equation}
where $\bm{{A}}_{0}$  and $\bm{{A}}_{1}$ represent learnable weight matrixes, and $BN(\cdot)$ represents batch normalization \cite{R43}. We next use layer index $\ell \in \{1,\cdots,L\}$ to represent the node embeddings $\bm{{x}}_{i}^{(\ell)}$ of the $\ell$-th layer obtained in the residual E-GAT. The inputs of the first layer of the residual E-GAT are the node features $\bm{x}^{(0)}=\{\bm{x}_1^{(0)},\bm{x}_2^{(0)},\cdots,\bm{x}_m^{(0)}\},\bm{x}_i^{(0)} \in \mathbb{R}^{d_x}$ and edge features $\hat{\bm{e}}=\{\hat{\bm{e}}_{11},\hat{\bm{e}}_{12},\cdots,\hat{\bm{e}}_{mm}\},\hat{\bm{e}}_{ij}\in \mathbb{R}^{d_e}$. The first layer of the residual E-GAT produces a new set of node features $\bm{x}^{(1)}=\{\bm{x}_1^{(1)},\bm{x}_2^{(1)},\cdots,\bm{x}_m^{(1)} \},\bm{x}_i^{(1)}\in \mathbb{R}^{d_x}$ as its output, while the edge features $\hat{\bm{e}}_{ij}$ remain unchanged.

Figure \ref{fig:fig2_a} describes how a single-layer residual E-GAT integrates the information of nodes and edges and updates the information of each node. The attention coefficient $\alpha_{ij}^{\ell}$ indicates the importance of node $i$ and node $j (i,j\in \{1,\cdots,m\})$ at the $\ell$-th layer:
\begin{equation}
\alpha_{ij}^{\ell}=\frac{exp(\sigma(\bm{g}^{\ell^{T}}[\bm{W}^{\ell}(\bm{x}_{i}^{(\ell-1)}||\bm{x}_{j}^{(\ell-1)}||\hat{\bm{e}}_{ij})]))}{\sum_{z=1}^{m}exp(\sigma(\bm{g}^{\ell^{T}}[\bm{W}^{\ell}(\bm{x}_{i}^{(\ell-1)}||\bm{x}_{z}^{(\ell-1)}||\hat{\bm{e}}_{iz})]))}
\label{eq:eq5}
\end{equation}
where $(\cdot)^T$ represents transposition, $\cdot || \cdot$ is the concatenation operation, $\bm{g}^{\ell}$ and $\bm{W}^{\ell}$ are learnable weight vectors and matrices respectively, and $\sigma(\cdot)$ is the LeakyReLU activation function (as adopted in GAT \cite{R12}). Figure \ref{fig:fig2_b} shows the encoder model structure with multi-layer residual E-GAT. Each layer of the residual E-GAT updates the feature vector of every node through the attention mechanism described by Equation \eqref{eq:eq5}. We have employed a residual connection between every two layers. That is, the output of layer $\ell$ is calculated as:
\begin{equation}
\bm{x}_{i}^{\ell}=\bm{x}_{i,R}^{(\ell)}+\bm{x}_{i}^{(\ell-1)},
\label{eq:eq6}
\end{equation}
where $\bm{x}_{i,R}^{(\ell)}$ is a function implemented by layer $\ell$ itself, which can be expressed as:
\begin{equation}
	 \bm{x}_{i,R}^{(\ell)}=\sum_{j=1}^{m}\alpha_{ij}^{\ell}\bm{W}_{1}^{\ell}\bm{x}_{j}^{(\ell-1)}, \label{eq:eq7}
\end{equation}
here $\bm{W}_{1}^{\ell}$ represents a learnable weight matrix. Our residual E-GAT contains $L$ layers. The $L$-th residual E-GAT outputs the node embeddings $\bm{x}_{i}^{(L)}$.  Subsequently, they are used to compute the final graph embedding (as in \cite{R22}) $\bar{\bm{x}}=\{\bar{x}_{1},\cdots,\bar{x}_{d_{x}} \},\bar{x}_{j}\in \mathbb{R}$, $\forall j\in \{1,2,\cdots,d_{x}\}$ by Equation \eqref{eq:eq8}
\begin{equation}
	\bar{x}_{j}=\frac{1}{m}\sum_{i=1}^{m}{(\bm{x}_{i}^{(L)})}_{j}, j=1,\cdots, d_{x} \label{eq:eq8}
\end{equation}
\subsection{The decoder}
\label{se:3.2}
Our decoder uses an attention mechanism similar to Kool et al. \cite{R22}, and is based on the decoder part of a Transformer model \cite{R39}. A Transformer model uses a multi-head attention mechanism rather than the commonly used loop layer in an encoder-decoder architecture.

\begin{figure}[htbp]
	\centering
	\includegraphics[width=5.8in]{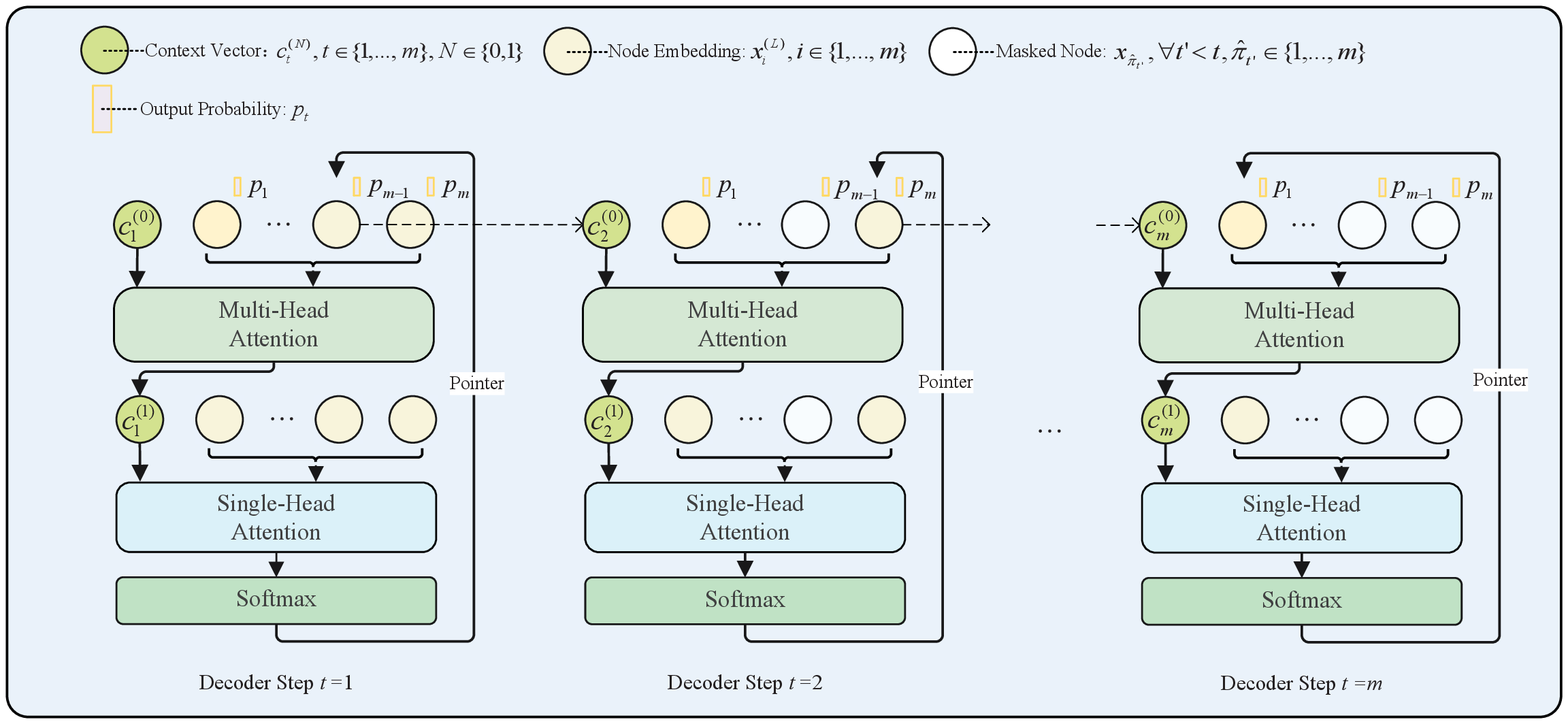}
	\caption{The decoder structure for TSP.}
	\label{fig:f3}
\end{figure}

However, a Transformer cannot directly be applied to solve combinatorial optimization problems, because its output dimension is fixed in advance and cannot be varied according to the input dimension. On the other hand, the PtrNet, proposed by Vinyals et al. \cite{R17}, uses an attention mechanism to select a member from the input sequence as the output at each decoding step based on a softmax probability distribution. The PtrNet enables a Transformer model to apply to combinatorial optimization problems, where the length of an output sequence is determined by the source sequence. With this idea in our mind, our decoder follows the PtrNet way to output nodes, in which each node is related to a probability value as a "pointer" at each decoding time step by using the softmax probability distribution. Different from a Transformer decoder in structure though, we do not use residual connections, batch normalization, and fully connected layers. Instead, our decoder contains two attention sub-layers. Figure \ref{fig:f3} illustrates the decoding process in our decoder.

The first layer computes a context vector through a multi-head attention mechanism \cite{R39}, and the second layer outputs the probability distribution of the nodes to be selected. Decoding happens sequentially, and at time step ${t}\in\{1,\cdots,m\}$ the decoder outputs the node $\hat{{\pi}}_{t}$ based on the embeddings from the encoder and the
previous outputs $\hat{\pi}_{{t}^{'}}$ generated at timestep ${{t}^{'}}<{t}$. The decoder context vector $\bm{c}_{t}^{(0)}$ at time step $t$ is calculated using the graph embedding $\bar{\bm{x}}$, the last selected node $\hat{\pi}_{t-1}$ and the fist selected node $\hat{\pi}_{1}$. For $t=1$, the decoder context vector is obtained from the graph embedding $\bar{\bm{x}}$ and a learnable $d_x$-dimensional parameter vector $\overrightarrow{\bm{v}}$:
\begin{equation}
\label{eq:eq9}
\bm{c}_{t}^{(0)}=\left\{
\begin{aligned}
\bar{\bm{x}}+\bm{W}_x{(\bm{x}_{\hat{\pi}_{1}}^{(L)}||\bm{x}_{\hat{\pi}_{t-1}}^{(L)})} & , & {t}>{1} \\
\bar{\bm{x}}+\overrightarrow{\bm{v}} & , & {t}={1}
\end{aligned}
.
\right.
\end{equation}
where $\bm{W}_{x}$ is a learnable weight matrix. The input of the first layer of the decoder is the context vector $\bm{c}_t^{(0)}$, and this layer produces a new context vector $\bm{c}_t^{(1)}$ which is obtained through a multi-head (with $H$ number of heads) attention mechanism. We next describe in detail the multi-head attention mechanism.

Generally, we define the dimension ${d}_{v}$ to compute the key vectors $\bm{k}_{i}\in\mathbb{R}^{d_{v}}$, value vectors $\bm{v}_{i}\in\mathbb{R}^{d_{v}}$, and query vector $\bm{q}\in\mathbb{R}^{d_{v}}$ through the node embeddings and the context vector $\bm{c}_{t}^{(0)}$:
\begin{equation}
	 \bm{q}=\bm{W}^{Q}\bm{c}_{t}^{(0)},\bm{v}_{i}=\bm{W}^{V}\bm{x}_{i}^{(L)},\bm{k}_{i}=\bm{W}^{K}\bm{x}_{i}^{(L)}, i\in\{1,2,\cdots,m\}. \label{eq:eq10}
\end{equation}
where $\bm{W}^{K}\in{\mathbb{R}^{{d_v}×{d_x}}}, \bm{W}^{Q}\in{\mathbb{R}^{{d_v}×{d_x}}}$ and $\bm{W}^{V}\in{\mathbb{R}^{{d_v}×{d_x}}}{(d_v={d_x/H})}$ are learnable weight matrices.

In this paper, the context vector $\bm{c}_{t}^{(0)}$ is used to compute a single query vector $\bm{q}$. The key vectors $\bm{k}=\{\bm{k}_1,\cdots,\bm{k}_m\}$  and value vectors $\bm{v}=\{\bm{v}_1,\cdots,\bm{v}_m\}$ are computed using the encoder output node embeddings $\bm{x}_{i}^{(L)}, i\in\{1,2,\cdots,m\}$. Then query vector $\bm{q}$ and key vectors $\bm{k}=\{\bm{k}_1,\cdots,\bm{k}_m\}$ are used to compute the attention coefficient $u_{i,t}^{(1)}\in\mathbb{R},i\in\{1,\cdots,m\}$ of the first decoder layer at time step $t$ via Equation \ref{eq:eq11}:
\begin{equation}
\label{eq:eq11}
u_{i,t}^{(1)}=\left\{
\begin{aligned}
\frac{\bm{q}^{T}\bm{k}_{i}}{\sqrt{d_v}}, if\;{i}\neq{\hat{\pi}_{{t}^{'}}} \;\forall{{t}^{'}}<{t}, \\
-\infty ,  otherwise.
\end{aligned}
\right.
\end{equation}
In Equation \ref{eq:eq11}, the nodes that have been selected before timestep $t$ are masked out by setting their attention coefficients to $-\infty$. We then normalize the attention coefficient ${u}_{i,t}^{(1)}$ through a softmax activation function by using Equation \ref{eq:eq12}:
\begin{equation}
\hat{u}_{i,t}^{(1)}=softmax{({u}_{i,t}^{(1)})}
\label{eq:eq12}
\end{equation}

Next, $H$-head independent attentions are performed according to Equation \ref{eq:eq13}, where the $h$-th ${({h}\in\{1,\cdots,H\})}$ normalized attention coefficient is denoted by ${(\hat{u}_{i,t}^{(1)})}^h$ for ${1}\leq{i}\leq{m}$.  The vectors computed by each head are connected in series, and then the final context vector $\bm{c}_{t}^{(1)}$  is obtained through a fully connected layer:
\begin{equation}
\bm{c}_{t}^{(1)}=\bm{W}_f\cdot{(||_{h=1}^{H}\sum_{i=1}^{m}{(\hat{u}_{i,t}^{(1)})}^h\bm{v}_{i}^{h})},
\label{eq:eq13}
\end{equation}
where $\bm{W}_{f}$ is a learnable weight matrix. The multi-head attention mechanism lends itself to enhance the stability of the attention learning process \cite{R12}.

The input to the second decoder layer with a single attention head is the context vector $\bm{c}_{t}^{(1)}$. We then compute the attention coefficient $\hat{u}_{i,t}^{(2)}\in\mathbb{R},{i}\in\{1,\cdots,m\}$ of the second decoder layer at timestep $t$ using Equation \ref{eq:eq14}. Referring to the work proposed by Bello et al. [18], we clipped the result within $[-C, C]$ (we used $C=10$ in this paper) using tanh. Then the probability ${p}_{i,t}$, for each node,$i\in\{1,\cdots,m\}$, is obtained through a softmax activation function using Equation \ref{eq:eq15}:
\begin{equation}
\label{eq:eq14}
{u}_{i,t}^{(2)}=\left\{
\begin{aligned}
{C}\cdot{tanh}{(\frac{{\bm{c}_{t}^{(1)}}^{T}\bm{k}_{i}}{\sqrt{{d}_{v}}})},{if}\;i\neq\hat{{\pi}}_{{t}^{'}}\;\forall{t}^{'}<{t} \\
-\infty ,  otherwise
\end{aligned}
\right.
.
\end{equation}
\begin{equation}
{p}_{i,t}={p}_{\theta}{(\hat{{\pi}}_{t}|{s},\hat{{\pi}}_{{t}^{'}}\;\forall{t}^{'}<{t})}={softmax}({{u}_{i,t}^{(2)})}.
\label{eq:eq15}
\end{equation}

Finally, according to the probability distribution $p_{i,t}$, we shall use either a sampling strategy or a greedy decoding strategy (to be introduced in Section \ref{se:se4.3}) to predict the next node to visit.
\section{DRL algorithm based on policy gradient}
\label{se:se4}
In this section, we present our implementation of two DRL algorithms: PPO and an Improved baseline REINFORCE algorithm. Both algorithms will be used in the proposed model.
\subsection{PPO for solving routing problems}
\begin{figure}[htbp]
	\centering
	\includegraphics[width=5in]{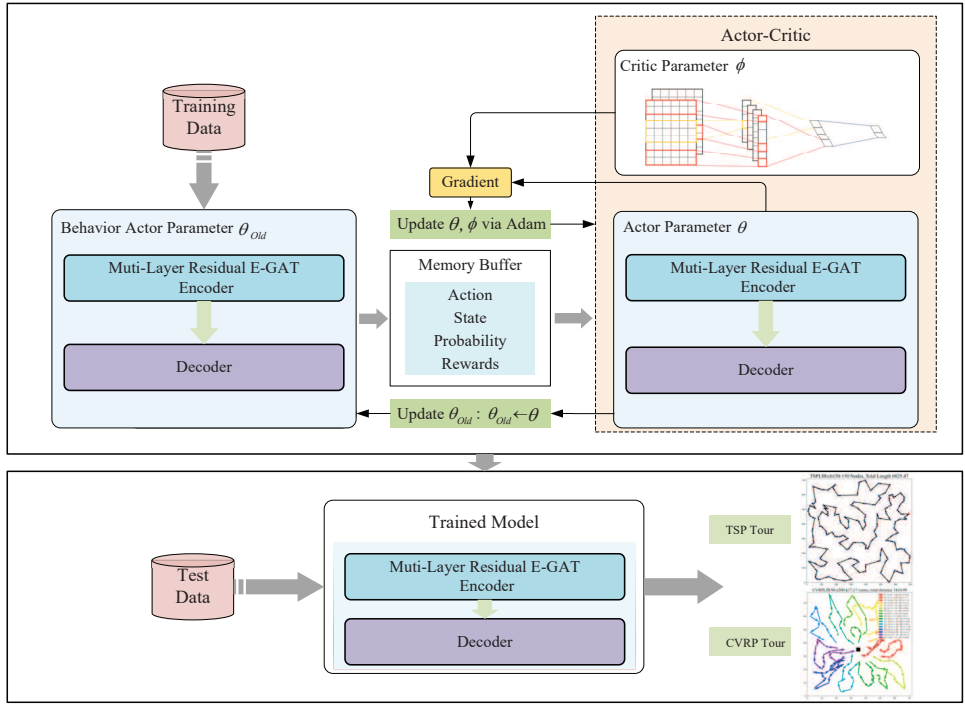}
	\caption{The DRL framework based on PPO.}
	\label{fig:f4}
\end{figure}

PPO \cite{R44,R45}, which is based on an actor-critic algorithm, is used to train our model. Figure \ref{fig:f4} is the PPO framework for routing problems. The actor refers to the graph-attention model mentioned in Section \ref{se:se3}. The critic network is composed of multiple one-dimensional convolutional layers and shares the same encoder network with the actor. The notation $\phi$ is used to represent the parameter set of the critic network.

In the PPO, we update the network after an entire episode ends. Equation \ref{eq:eq1} in Section \ref{se:se3} defines the travel length ${L}{(\hat{\pi}|{s})}$, which is used for obtaining an unbiased Monte Carlo estimation of the policy gradient. The critic network is regarded as the baseline and output a scalar $\hat{v}_{\phi}{(s)}$ to estimate the cumulative rewards, i.e., the travel length ${L}{(\hat{\pi}|{s})}$. Advantage estimation $\hat{A}$  is computed by Equation \ref{eq:eq16}. In Equation \ref{eq:eq17}, ${p}_{\theta}{(\hat{\pi}|{s})}$ is defined by Equation \ref{eq:eq2} in Section \ref{se:se3}, and ${r}{({\theta})}$ represents the probability ratio between the updated and the prior-update policy.
\begin{equation}
\hat{A}={L}{(\hat{\pi}|{s})}-\hat{v}_{\phi}{(s)}
\label{eq:eq16}
\end{equation}
\begin{equation}
{r}{({\theta})}=\frac{{p}_{\theta}{(\hat{\pi}|{s})}}{{{p}_{{\theta}_{old}}{(\hat{\pi}|{s})}}}
\label{eq:eq17}
\end{equation}
The clipped surrogate objective ${L}_{CLIP}{({\theta})}$ and the policy entropy loss function ${L}_{E}{({\theta})}$, computed by Equation \ref{eq:eq18} and Equation \ref{eq:eq19}, respectively, are used to train the actor network. For more detailed information about the clip and clip coefficient $\epsilon$  in Equation \ref{eq:eq18}, please refers to the PPO \cite{R44} (${\epsilon}={0.2}$ is chosen in \cite{R44}).
\begin{equation}
{L}_{CLIP}{({\theta})}=\hat{\mathbb{E}}{[}{min}\{{r(\theta)}\hat{A},{clip}{({r({\theta})},{1}-{\epsilon},{1}+{\epsilon})}\hat{A}\}].
\label{eq:eq18}
\end{equation}
\begin{equation}
{L}_{E}{({\theta})}={Entropy}{({p}_{\theta}{(\hat{\pi}|{s})})}.
\label{eq:eq19}
\end{equation}
The critic network is trained by the Mean Squared Error (MSE) loss ${L}_{MSE}{({\phi})}$  computed by:
\begin{equation}
{L}_{MSE}{({\phi})}={MSE}{(}{L}{(\hat{\pi}|{s})},\hat{v}_{\phi}{(s)}{)}.
\label{eq:eq20}
\end{equation}
Then the total loss function of the actor-critic model can be expressed as:
\begin{equation}
{Loss}{({\theta},{\phi})}={c}_{p}{L}_{CLIP}{({\theta})}+{c}_{v}{L}_{MSE}{({\phi})}-{c}_{e}{L}_{E}{({\theta})},
\label{eq:eq21}
\end{equation}
which is composed of three parts, including ${L}_{CLIP}{({\theta})}$, ${L}_{MSE}{({\phi})}$, and ${L}_{E}{({\theta})}$. And their weights are coordinated by parameters ${c}_{p}$, ${c}_{v}$, and ${c}_{e}$. The objective is to reduce the ${L}_{CLIP}{({\theta})}$ and the ${L}_{MSE}{({\phi})}$ and to increase the ${L}_{E}{({\theta})}$  through gradient descent \cite{R44}. Table \ref{tab:tab2} shows the instantiations of the reinforcement learning framework for the two routing problems.
\begin{table}[htbp]
	\tiny
	\centering
	\caption{Definition of reinforcement learning components for routing problems.}
	\begin{tabular}{lllll}
		\toprule
		Problem & State & Action & Reward & Termination \\
		\midrule
		TSP   & Partial tour & Grow tour by one node & Change in tour cost & Tour includes all nodes \\
		CVRP  & Partial tour and vehicle inventory & Grow tour by one node or depot & Change in tour cost & All needs are met \\
		\bottomrule
	\end{tabular}%
	\label{tab:tab2}%
\end{table}%

However, getting such an actor-critic algorithm to work is non-trivial. Engstrom et al. \cite{R46} believes that the success of the PPO algorithm comes from minor modifications to the core algorithm Trust Region Policy Optimization \cite{R47}. They called these modifications "code-level optimization" and proved the effectiveness of these modifications. This paper adopts some parts of "code-level optimization" when using the PPO algorithm to train our encoder-decoder model. The following tricks sourced from the "code-level optimization" are used in our framework:
\begin{itemize}
\item \textbf{Normalization of the reward in each mini-batch:} In a standard PPO implementation, rather than feeding the reward directly from the environment into the objective, it performs a certain discount-based scaling scheme. We did not perform this discount-based scaling scheme. Instead, we normalize reward ${L}{(\hat{\pi}|{s})}$ by subtracting the mean and dividing by the standard deviation in each mini batch.
\item \textbf{Adam optimizer learning rate annealing:} We used learning rate decay during training, and update the learning rate at the end of each epoch:
\begin{equation}
{l}_{new}={l}_{old}\cdot{({\beta})}^{epoch},
\label{eq:eq22}
\end{equation}
where $l_{new}$ and $l_{old}$ are the learning rate before and after each update, and ${\beta}$ is a learning rate annealing coefficient.
\item \textbf{Orthogonal initialization:} Instead of using the default weight initialization scheme for the actor and critic networks, we used an orthogonal initialization following Engstrom et al. \cite{R46}.
\item \textbf{Global Gradient Clipping:} After computing the loss function gradient with respect to the actor and the critic networks, we clip the concatenated gradient of all parameters such that the "global L2 norm" does not exceed 2 following the recommendation in Bello et al. \cite{R18}.
\end{itemize}

The whole PPO algorithm is listed in Algorithm \ref{alg:alg1}.

\begin{algorithm}[htbp]
	\small
	\caption{PPO for Combinatorial Optimization}
	\label{alg:alg1}
	\begin{algorithmic}[1]
		\Require number of total epochs $E_t$, PPO epochs $E_p$, PPO steps $T_s$, steps per total epoch $P_e$, batch size $B$; actor network $\pi_{\theta}$ and behaviour actor network $\pi_{\theta_{old}}$ with trainable parameters $\theta$ and $\theta_{old}$; critic network $v_{\phi}$ with trainable parameters $\phi$; policy loss coefficient $c_p$; value function loss coefficient $c_v$; entropy loss coefficient $c_e$; clipping ratio $\epsilon$.
		\State \textbf{Initialization}:Orthogonal initialization {$\theta;\phi,\theta_{old}\leftarrow\theta$}
		\ForEach {$total\; epoch =1,{\cdots},E_t$}
			\ForEach {$step=1,\cdots,P_e$}
				\State $s_i\sim$ RandomInstance() $\forall{i}\in\{1,\cdots,B\}$
				\State $\hat{{\pi}}_i\sim$ SampleSolution($p_{{\theta}_{old}}{(\hat{{\pi}}_i{|}s_i)}$) $\forall{i}\in\{1,\cdots,B\}$
				\State Compute $L{(\hat{{\pi}}_i|s_i)}$ via $s_i, \hat{{\pi}}_i$
				\State Put $s_i,\hat{{\pi}}_i,L{(\hat{{\pi}}_i|s_i)},p_{{\theta}_{old}}{(\hat{{\pi}}_i{|}s_i)}$ into Memory Buffer
					\If{$step \% T_s == 0$}
						\ForEach {$PPO\;epoch =1,\cdots,E_p$}
							\ForEach {$PPO\;step=1,\cdots,T_s$}
							\State $\hat{v}_\phi{(s_i)}\leftarrow{V_\phi{(s_i)}}\;\forall{i}\in\{1,\cdots,B\}$
							\State $\bar{L}{(\hat{{\pi}}_i|s_i)}\leftarrow$ Reward Normalization ${L}{(\hat{{\pi}}_i|s_i)}\;\forall{i}\in\{1,\cdots,B\}$
							\State Compute advantage estimates $\hat{A}_i=\bar{L}{(\hat{{\pi}}_i|s_i)}-\hat{v}_\phi{(s_i)}$
							\State ${r}_{i}{({\theta})}=\frac{{p}_{\theta}{(\hat{\pi}_i|{s}_i)}}{{{p}_{{\theta}_{old}}{(\hat{\pi}_i|{s}_i)}}}$
							\State ${L}_{CLIP}^i{({\theta})}=\hat{\mathbb{E}}{[}{min}\{{r_i(\theta)}\hat{A}_i,{clip}{({r_i({\theta})},{1}-{\epsilon},{1}+{\epsilon})}\hat{A}_i\}]$
							\State $L_{MSE}^i{(\phi)}=MSE(\bar{L}{(\hat{{\pi}}_i|s_i)},\hat{v}_\phi{(s_i)})$
							\State $L_E^i{(\theta)}=Entropy(p_{{\theta}}{(\hat{{\pi}}_i{|}s_i)})$
							\State ${Loss}_i{({\theta},{\phi})}=\frac{1}{B}\sum_{i=1}^{B}\{{c}_{p}{L}^i_{CLIP}{({\theta})}+{c}_{v}{L}^i_{MSE}{({\phi})}-{c}_{e}{L}^i_{E}{({\theta})}\}$					 
							\State Update ${\theta,\phi}$ by a gradient method w.r.t. ${Loss}_i{({\theta},{\phi})}$
							\EndFor
						\EndFor
						\State $\theta_{old}\leftarrow\theta$
						\State Clear Memory Buffer
					\EndIf
			\EndFor
		\EndFor
		\Ensure Trained parameters set $\theta$ of the actor
	\end{algorithmic}
\end{algorithm}

\subsection{Improved baseline REINFORCE algorithm}
\label{se:se4.2}
In addition to the PPO algorithm described earlier, we also used an improved baseline REINFORCE algorithm to train our model. To that end, we defined the loss function as ${Loss}{({\theta}|{s})}=\mathbb{E}_{\hat{\pi}\sim{p}_{\theta}{(\hat{\pi}|{s})}}{[{L}{(\hat{\pi}|{s})}]}$. The parameters $\theta$ is optimized by gradient descent using the REINFORCE \cite{R48} algorithm with rollout baseline $b$:
\begin{equation}
\nabla_{\theta}{Loss}{({\theta}|{s})}=\mathbb{E}_{\hat{\pi}\sim{p}_{\theta}{(\hat{\pi}|{s})}}{[{({L}{(\hat{\pi}|{s})}-{b})}}\nabla_{\theta}{log}{{p}_{{\theta}}{(\hat{\pi}|{s})}}].
\label{eq:eq23}
\end{equation}
REINFORCE algorithm with rollout baseline was proposed by Kool et al. \cite{R22} in solving routing problems. The critic network in the actor-critic algorithm was replaced by the so-called baseline actor (policy) network, which can be described as a double actor structure. This replacing strategy freezes the parameters ${\theta}^{BL}$ of the baseline policy network ${\pi}_{{{\theta}^{BL}}}$ in each epoch, somewhat similar to freezing the target Q-network in DQN \cite{R49}. At the end of each epoch, a greedy decoding is used to compare the results of the current training policy and the baseline policy. Then the parameters of the baseline policy network are updated only when an improvement is significant according to a paired t-test (significance level ${\;\alpha}={5}\%$) on 10000 evaluation instances as done in existing studies \cite{R7,R18,R19,R20,R22}. In the training process, "code-level optimization" including Adam optimizer learning rate annealing and normalization of the reward function is used to improve the performance of the baseline REINFORCE algorithm. We dubbed the modified algorithm improved baseline REINFORCE algorithm (Rollout algorithm). The steps are described in Algorithm \ref{alg:alg2}.
\begin{algorithm}[htb]
	\small
	\caption{Improved baseline REINFORCE Algorithm}
	\label{alg:alg2}
	\begin{algorithmic}[1]
		\Require number of epochs $E$, steps per epoch $P_e$, batch size $B$, significance level $\alpha$; actor network $\pi_\theta$  with trainable parameters $\theta$; Baseline policy network $\pi_{{\theta}^{BL}}$ with trainable parameters ${\theta}^{BL}$.
		
		\State \textbf{Initialization}:Xavier initialization {$\theta,\theta^{BL}\leftarrow\theta$}
		\ForEach {$epoch =1,{\cdots},E_t$}
		\ForEach {$step=1,\cdots,P_e$}
		
		\State $s_i\sim$ RandomInstance() $\forall{i}\in\{1,\cdots,B\}$
		\State $\hat{{\pi}}_i\sim$ SampleSolution($p_{{\theta}}{(\hat{{\pi}}_i{|}s_i)}$) $\forall{i}\in\{1,\cdots,B\}$
		\State $\hat{{\pi}}_i^{BL}\sim$ GreedySolution($p_{{\theta}^{BL}}{(\hat{{\pi}}_i{|}s_i)}$) $\forall{i}\in\{1,\cdots,B\}$
		\State Compute $L{(\hat{{\pi}}_i|s_i)},L{(\hat{{\pi}}^{BL}_i|s_i)}$ via $s_i, \hat{{\pi}}_i,\hat{{\pi}}^{BL}_i$
		\State $\bar{L}{(\hat{{\pi}}_i|s_i)}\leftarrow$ Reward Normalization ${L}{(\hat{{\pi}}_i|s_i)}$
		\State $\bar{L}{(\hat{{\pi}}^{BL}_i|s_i)}\leftarrow$ Baseline Reward Normalization $L{(\hat{{\pi}}^{BL}_i|s_i)}$
		\State Compute advantage estimates $\hat{A}_i={\bar{L}{(\hat{{\pi}}_i|s_i)}-\bar{L}{(\hat{{\pi}}^{BL}_i|s_i)})}$
		\State $\nabla_{\theta}{Loss}{({\theta}|{s}_i)}=\frac{1}{B}\sum_{i=1}^{B}\mathbb{E}_{\hat{\pi}\sim{p}_{\theta}{(\hat{\pi}|{s})}}{[\hat{A}_i}\nabla_{\theta}{log}{{p}_{{\theta}}{(\hat{\pi}_i|{s}_i)}}]$
		\State $\theta\leftarrow Adam({\theta},\nabla_{\theta}{Loss}{({\theta}|{s}_i)})$

		\EndFor
		\If{OneSidedPairedTTest$(p_{\theta},p_{{\theta}^{BL}})<{\alpha}$}
		\State ${\theta}^{BL}\leftarrow{\theta}$
		
		\EndIf
		
		\EndFor
		\Ensure Trained parameters set $\theta$ of the actor
	\end{algorithmic}
\end{algorithm}
\subsection{Decoding strategy}
\label{se:se4.3}
Effective search algorithms for combinatorial optimization problems include beam search, neighborhood search and tree search. Bello et al. \cite{R18} proposed search strategies such as sampling and Active Search. We used the following two decoding strategies:
\begin{itemize}
\item \textbf{Greedy Decoding}: Generally, a greedy algorithm selects a local optimal solution and provides a fast approximation of the global optimal solution. In each decoding step, the node with the highest probability is selected greedily, and the visited nodes are masked out. For the TSP problem, the search is terminated when all nodes have been visited. For CVRP, the search is terminated when the requirements of all nodes are satisfied to construct an effective solution.
\item \textbf{Stochastic Sampling}: In each decoding timestep ${t}\in\{1,\cdots,m\}$, the random policy ${p}_{\theta}{(\hat{\pi}_{t}|{s},\hat{\pi}_{{t}^{'}}\;\forall{t}^{'}<{t})}$ samples the nodes to be selected according to the probability distribution to construct an effective solution. During testing, Bello et al. \cite{R18} used temperature hyperparameters $\lambda\in\mathbb{R}$ to modify Equation \ref{eq:eq15} to ensure the diversity of sampling. The modified Equation is as follows:
\begin{equation}
{p}_{i,t}={p}_{\theta}{(\hat{{\pi}}_{t}|{s},\hat{{\pi}}_{{t}^{'}}\;\forall{t}^{'}<{t})}=
{softmax}{(\frac{{u}_{i,t}^{(2)}}{\lambda})}
\label{eq:eq24}
\end{equation}
Through grid search of the temperature hyperparameters, we found that the temperature of 2, 2.5, 1.5 provide best results for TSP20, TSP50, and TSP100, respectively. In addition, the temperature of 2.5, 1.8, 1.2 supply the best results for CVRP20, CVRP50, and CVRP100, respectively.
\end{itemize}

In the training process, stochastic sampling is usually needed to explore the environment to obtain a better model performance. In the testing process, we used the greedy decoding. Moreover, we sampled 1280 solutions (following the existing studies \cite{R7,R18,R19,R20,R22}) by using stochastic sampling with a selected temperature hyperparameters $\lambda$ and listed the best one in Table \ref{tab:ta4} (the row indicated by "\textbf{Ours (Sampling)}") and Table \ref{tab:ta5} (the row indicated by "\textbf{Ours (Sampling)}") .
\section{Computational Experiment}
\label{se:se5}
Our framework is suitable for many combinatorial optimization problems. For different optimization problems, only the input, masks and decoder context vectors need to be adjusted. This paper mainly focuses on the routing problems TSP and CVRP. We trained the proposed graph-attention model for both TSP and CVRP instances with the number of nodes $m=20,50$ and 100, respectively. Instances were generated with batch sizes of 512, 128, 128 (affected by memory) for the three different numbers of nodes, respectively. Each epoch of the PPO algorithm contains 800, 3000, and 3000 batches respectively, and for the Rollout algorithm each contains 1600, 6000, and 6000 batches correspondingly. We trained 100 epochs using instances randomly generated from the unit square $[0,1]×[0,1]$. And 10,000 test instances following the existing studies \cite{R7,R18,R19,R20,R22} are generated with the same data distribution. As the testing process does not update the model parameters, a larger batch size can be used. We used a Nvidia GeForce RTX2070 GPU to complete the training of TSP and CVRP with sizes of $m=20$ and $50$. Since the memory of GPU in the personal computer is limited, the training process with instances of $m=100$ and all testing processes are performed on an Nvidia Tesla V100 GPU. The values of other related hyper-parameters for the training process are listed in Table \ref{tab:ta3}. Furthermore, we analyzed the sensitivities of hyper-parameters in \ref{app:a}. The hyper-parameter values for TSP and CVRP of the same size are identical. The model is constructed by PyTorch \cite{R50}, and the code is implemented by using Python 3.7.
\begin{table}[htbp]
	\tiny
	\centering
	\caption{The values of the hyper-parameters used in our framework.}
	\begin{tabular}{lc|lc}
		\toprule
		\multicolumn{1}{l}{Parameters} & \multicolumn{1}{c|}{Value} & \multicolumn{1}{l}{Parameters} & \multicolumn{1}{c}{Value} \\
		\multicolumn{1}{l}{\multirow{2}[0]{*}{Number of encoder layers $\ell$}} & \multirow{2}[0]{*}{4} & \multicolumn{1}{l}{\multirow{2}[0]{*}{PPO steps $T_s$}} & \multirow{2}[0]{*}{1} \\
		&     &     &  \\
		\multicolumn{1}{l}{\multirow{2}[0]{*}{Learning rate decay $\beta$}} & \multirow{2}[0]{*}{0.96} & \multicolumn{1}{l}{\multirow{2}[0]{*}{Heads of attention $H$}} & \multirow{2}[0]{*}{8} \\
		&     &     &  \\
		\multicolumn{1}{l}{\multirow{2}[0]{*}{PPO learning rate $l$}} & \multirow{2}[0]{*}{$3×10^{-4}(m=20)$} & \multicolumn{1}{l}{\multirow{2}[0]{*}{PPO epochs $E_p$}} & \multirow{2}[0]{*}{3} \\
		&     &     &  \\
		& $1×10^{-4}(m=50,100)$   & \multicolumn{1}{l}{Optimizer} & \multicolumn{1}{c}{Adam \cite{R51}} \\
		\multicolumn{1}{l}{\multirow{2}[0]{*}{Rollout learning rate $l$}} & \multirow{2}[0]{*}{$1×10^{-3}(m=20)$} & \multicolumn{1}{l}{\multirow{2}[0]{*}{Node embedding dimension $d_x$}} & \multirow{2}[0]{*}{128} \\
		&     &     &  \\
		\multirow{2}[1]{*}{} & \multirow{2}[1]{*}{$3×10^{-4}(m=50,100)$} & \multicolumn{1}{l}{\multirow{2}[1]{*}{Edge embedding dimension $d_e$}} & \multirow{2}[1]{*}{64} \\
		&     &     &  \\
		\bottomrule
	\end{tabular}%
	\label{tab:ta3}%
\end{table}%
\subsection{TSP results}
\label{se:se5.1}
The test results of TSP instances for difference sizes are reported in Table \ref{tab:ta4}. The comparison benchmark method is the Gurobi solver. The optimization solver Gurobi \cite{R52} can obtain an exact solution of TSP via mixed integer programming within a reasonable running time. The Google OR tools are developed based on local search and can be generally used to judge non-learned algorithms, such as heuristics with greedy search strategies. The listed results cover three groups of solution techniques, including exact solver, greedy approaches, and sampling/search approaches. Except for Gurobi and our results (both shown in bold) in the Table \ref{tab:ta4}, all others are taken from Table 1 \cite{R22} and Table 1 \cite{R7}. For more detailed information about non-learned algorithms (Nearest Insertion, Random Insertion, Farthest Insertion, Nearest Neighbor), please refer to Appendix B of Kool et al. \cite{R22}. All the results are expressed by the average travel length ${L}{(\hat{\pi}|{s})}$ of ${N}_{test}$ (here $N_{test}=10,000$) test instances. And Opt. Gap denotes the average optimal gap between a result and the optimal solution ${L}{({\pi}|{s})}$ delivered by the optimization solver Gurobi, as shown in the following Equation:
\begin{equation}
{Opt.\;Gap}=\frac{1}{N_{test}}\sum_{i=1}^{N_{test}}{\frac{{L}{(\hat{\pi}|{s})}-{L}{({\pi}|{s})}}{{L}{({\pi}|{s})}}}
\label{eq:eq25}
\end{equation}
\begin{table}[htbp]
	\tiny
	\centering
	\caption{Performance of our framework compared to non-learned algorithms and state-of-the-art methods for TSP instances.}
	\begin{tabular}{lcccccccccc}
		\toprule
		\multirow{2}[2]{*}{Method} & \multicolumn{1}{c|}{\multirow{2}[2]{*}{Type}} & \multicolumn{3}{c|}{TSP20} & \multicolumn{3}{c|}{TSP50} & \multicolumn{3}{c}{TSP100} \\
		\multicolumn{1}{r}{} & \multicolumn{1}{c|}{} & \multicolumn{1}{c}{Length} & \multicolumn{1}{c}{Gap} & \multicolumn{1}{c|}{Time} & \multicolumn{1}{c}{Length} & \multicolumn{1}{c}{Gap} & \multicolumn{1}{c|}{Time} & \multicolumn{1}{c}{Length} & \multicolumn{1}{c}{Gap} & \multicolumn{1}{c}{Time} \\
		\midrule
		\textbf{Gurobi} & \multicolumn{1}{c|}{\textbf{Solver}} & \multicolumn{1}{c}{\textbf{3.829}} & \multicolumn{1}{c}{\textbf{0.00\%}} & \multicolumn{1}{c|}{\textbf{1m}} & \multicolumn{1}{c}{\textbf{5.698}} & \multicolumn{1}{c}{\textbf{0.00\%}} & \multicolumn{1}{c|}{\textbf{20m}} & \multicolumn{1}{c}{\textbf{7.763}} & \multicolumn{1}{c}{\textbf{0.00\%}} & \multicolumn{1}{c}{\textbf{3h}} \\
		\midrule
		Nearest Insertion & \multicolumn{1}{c|}{H,G} & \multicolumn{1}{c}{4.33} & \multicolumn{1}{c}{12.91\%} & \multicolumn{1}{c|}{1s} & \multicolumn{1}{c}{6.78} & \multicolumn{1}{c}{19.03\%} & \multicolumn{1}{c|}{2s} & \multicolumn{1}{c}{9.46} & \multicolumn{1}{c}{21.82\%} & \multicolumn{1}{c}{6s} \\
		Random Insertion & \multicolumn{1}{c|}{H,G} & \multicolumn{1}{c}{4.00} & \multicolumn{1}{c}{4.36\%} & \multicolumn{1}{c|}{0s} & \multicolumn{1}{c}{6.13} & \multicolumn{1}{c}{7.65\%} & \multicolumn{1}{c|}{1s} & \multicolumn{1}{c}{8.52} & \multicolumn{1}{c}{9.69\%} & \multicolumn{1}{c}{3s} \\
		Farthest Insertion & \multicolumn{1}{c|}{H,G} & \multicolumn{1}{c}{3.93} & \multicolumn{1}{c}{2.36\%} & \multicolumn{1}{c|}{1s} & \multicolumn{1}{c}{6.01} & \multicolumn{1}{c}{5.53\%} & \multicolumn{1}{c|}{2s} & \multicolumn{1}{c}{8.35} & \multicolumn{1}{c}{7.59\%} & \multicolumn{1}{c}{7s} \\
		Nearest Neighbor & \multicolumn{1}{c|}{H,G} & \multicolumn{1}{c}{4.50} & \multicolumn{1}{c}{17.23\%} & \multicolumn{1}{c|}{0s} & \multicolumn{1}{c}{7.00} & \multicolumn{1}{c}{22.94\%} & \multicolumn{1}{c|}{0s} & \multicolumn{1}{c}{9.68} & \multicolumn{1}{c}{24.73\%} & \multicolumn{1}{c}{0s} \\
		PtrNet \cite{R17} & \multicolumn{1}{c|}{SL,G} & \multicolumn{1}{c}{3.88} & \multicolumn{1}{c}{1.15\%} & \multicolumn{1}{c|}{} & \multicolumn{1}{c}{7.66} & \multicolumn{1}{c}{34.48\%} & \multicolumn{1}{c|}{} & \multicolumn{3}{c}{-} \\
		PtrNet \cite{R18} & \multicolumn{1}{c|}{RL,G} & \multicolumn{1}{c}{3.89} & \multicolumn{1}{c}{1.42\%} & \multicolumn{1}{c|}{} & \multicolumn{1}{c}{5.95} & \multicolumn{1}{c}{4.46\%} & \multicolumn{1}{c|}{} & \multicolumn{1}{c}{8.30} & \multicolumn{1}{c}{6.90\%} &  \\
		S2V \cite{R9} & \multicolumn{1}{c|}{RL,G} & \multicolumn{1}{c}{3.89} & \multicolumn{1}{c}{1.42\%} & \multicolumn{1}{c|}{} & \multicolumn{1}{c}{5.99} & \multicolumn{1}{c}{5.16\%} & \multicolumn{1}{c|}{} & \multicolumn{1}{c}{8.31} & \multicolumn{1}{c}{7.03\%} &  \\
		GAT \cite{R20} & \multicolumn{1}{c|}{RL,G} & \multicolumn{1}{c}{3.86} & \multicolumn{1}{c}{0.66\%} & \multicolumn{1}{c|}{2m} & \multicolumn{1}{c}{5.92} & \multicolumn{1}{c}{3.98\%} & \multicolumn{1}{c|}{5m} & \multicolumn{1}{c}{8.42} & \multicolumn{1}{c}{8.41\%} & \multicolumn{1}{c}{8m} \\
		GAT \cite{R20} & \multicolumn{1}{c|}{RL,G,2OPT} & \multicolumn{1}{c}{3.85} & \multicolumn{1}{c}{0.42\%} & \multicolumn{1}{c|}{4m} & \multicolumn{1}{c}{5.85} & \multicolumn{1}{c}{2.77\%} & \multicolumn{1}{c|}{26m} & \multicolumn{1}{c}{8.17} & \multicolumn{1}{c}{5.21\%} & \multicolumn{1}{c}{3h} \\
		AM \cite{R22} & \multicolumn{1}{c|}{RL,G} & \multicolumn{1}{c}{3.85} & \multicolumn{1}{c}{0.34\%} & \multicolumn{1}{c|}{0s} & \multicolumn{1}{c}{5.80} & \multicolumn{1}{c}{1.76\%} & \multicolumn{1}{c|}{2s} & \multicolumn{1}{c}{8.12} & \multicolumn{1}{c}{4.53\%} & \multicolumn{1}{c}{6s} \\
		GCN \cite{R7} & \multicolumn{1}{c|}{SL,G} & \multicolumn{1}{c}{3.86} & \multicolumn{1}{c}{0.60\%} & \multicolumn{1}{c|}{6s} & \multicolumn{1}{c}{5.87} & \multicolumn{1}{c}{3.10\%} & \multicolumn{1}{c|}{55s} & \multicolumn{1}{c}{8.41} & \multicolumn{1}{c}{8.38\%} & \multicolumn{1}{c}{6m} \\
		\textbf{Ours (Greedy)} & \multicolumn{1}{c|}{\textbf{RL,G}} & \multicolumn{1}{c}{\textbf{3.832}} & \multicolumn{1}{c}{\textbf{0.06\%}} & \multicolumn{1}{c|}{\textbf{1s}} & \multicolumn{1}{c}{\textbf{5.755}} & \multicolumn{1}{c}{\textbf{0.98\%}} & \multicolumn{1}{c|}{\textbf{4s}} & \multicolumn{1}{c}{\textbf{8.048}} & \multicolumn{1}{c}{\textbf{3.67\%}} & \multicolumn{1}{c}{\textbf{11s}} \\
		\midrule
		OR Tools & \multicolumn{1}{c|}{H,S} & \multicolumn{1}{c}{3.85} & \multicolumn{1}{c}{0.37\%} & \multicolumn{1}{c|}{} & \multicolumn{1}{c}{5.80} & \multicolumn{1}{c}{1.83\%} & \multicolumn{1}{c|}{} & \multicolumn{1}{c}{7.99} & \multicolumn{1}{c}{2.90\%} &  \\
		Chr.f. + 2OPT & \multicolumn{1}{c|}{H,2OPT} & \multicolumn{1}{c}{3.85} & \multicolumn{1}{c}{0.37\%} & \multicolumn{1}{c|}{} & \multicolumn{1}{c}{5.79} & \multicolumn{1}{c}{1.65\%} & \multicolumn{1}{c|}{} & \multicolumn{3}{c}{-} \\
		PtrNet \cite{R18} & \multicolumn{1}{c|}{RL,S} & \multicolumn{3}{c|}{-} & \multicolumn{1}{c}{5.75} & \multicolumn{1}{c}{0.95\%} & \multicolumn{1}{c|}{} & \multicolumn{1}{c}{8.00} & \multicolumn{1}{c}{3.03\%} &  \\
		GNN \cite{R40} & \multicolumn{1}{c|}{SL,BS} & \multicolumn{1}{c}{3.93} & \multicolumn{1}{c}{2.46\%} & \multicolumn{1}{c|}{} & \multicolumn{3}{c|}{-} & \multicolumn{3}{c}{-} \\
		GAT \cite{R20} & \multicolumn{1}{c|}{RL,S} & \multicolumn{1}{c}{3.84} & \multicolumn{1}{c}{0.11\%} & \multicolumn{1}{c|}{5m} & \multicolumn{1}{c}{5.77} & \multicolumn{1}{c}{1.28\%} & \multicolumn{1}{c|}{17m} & \multicolumn{1}{c}{8.75} & \multicolumn{1}{c}{12.70\%} & \multicolumn{1}{c}{56m} \\
		GAT \cite{R31} & \multicolumn{1}{c|}{RL,S,2OPT} & \multicolumn{1}{c}{3.84} & \multicolumn{1}{c}{0.09\%} & \multicolumn{1}{c|}{6m} & \multicolumn{1}{c}{5.75} & \multicolumn{1}{c}{1.00\%} & \multicolumn{1}{c|}{32m} & \multicolumn{1}{c}{8.12} & \multicolumn{1}{c}{4.64\%} & \multicolumn{1}{c}{5h} \\
		AM \cite{R22} & \multicolumn{1}{c|}{RL,S} & \multicolumn{1}{c}{3.84} & \multicolumn{1}{c}{0.08\%} & \multicolumn{1}{c|}{5m} & \multicolumn{1}{c}{5.73} & \multicolumn{1}{c}{0.52\%} & \multicolumn{1}{c|}{24m} & \multicolumn{1}{c}{7.94} & \multicolumn{1}{c}{2.26\%} & \multicolumn{1}{c}{1h} \\
		GCN \cite{R7} & \multicolumn{1}{c|}{SL,BS} & \multicolumn{1}{c}{3.84} & \multicolumn{1}{c}{0.10\%} & \multicolumn{1}{c|}{20s} & \multicolumn{1}{c}{5.71} & \multicolumn{1}{c}{0.26\%} & \multicolumn{1}{c|}{2m} & \multicolumn{1}{c}{7.92} & \multicolumn{1}{c}{2.11\%} & \multicolumn{1}{c}{10m} \\
		GCN \cite{R40} & \multicolumn{1}{c|}{SL,BS*} & \multicolumn{1}{c}{3.84} & \multicolumn{1}{c}{0.01\%} & \multicolumn{1}{c|}{12m} & \multicolumn{1}{c}{5.70} & \multicolumn{1}{c}{0.01\%} & \multicolumn{1}{c|}{18m} & \multicolumn{1}{c}{7.87} & \multicolumn{1}{c}{1.39\%} & \multicolumn{1}{c}{40m} \\
		\textbf{Ours (Sampling)} & \multicolumn{1}{c|}{\textbf{RL,S}} & \multicolumn{1}{c}{\textbf{3.830}} & \multicolumn{1}{c}{\textbf{0.01\%}} & \multicolumn{1}{c|}{\textbf{10m}} & \multicolumn{1}{c}{\textbf{5.738}} & \multicolumn{1}{c}{\textbf{0.70\%}} & \multicolumn{1}{c|}{\textbf{55m}} & \multicolumn{1}{c}{\textbf{7.896}} & \multicolumn{1}{c}{\textbf{1.71\%}} & \multicolumn{1}{c}{\textbf{2h}} \\
		\bottomrule
		\multicolumn{11}{p{56em}}{Note: DL/DRL approaches are named according to the type of neural network used. In the Type column, H: heuristic method; SL: supervised learning; S: sample search; G: greedy search; BS: beam search; BS*: beam search and shortest path heuristics; 2OPT: 2OPT local search.} \\
	\end{tabular}%
	\label{tab:ta4}%
\end{table}%

In Table \ref{tab:ta4}, our results are obtained by applying our trained graph attention model to the TSP testing instances. The applied model is trained by using the Rollout algorithm introduced in Section \ref{se:se4.2}, as the PPO algorithm performs slightly less well. We point out that, for our trained model, the greedy decoding strategy can obtain almost the same solution quality as that by the sampling strategy, for example, $3.67\%$ vs. $1.71\%$ Opt. Gap for TSP100 instances. Furthermore, for our model, the running time of the greedy decoding strategy is much faster compared to the sampling strategy, for example, 11s vs. 2h for TSP100 instances.

Compared with other learning-based methods, our method has a significant improvement in the quality of solutions when adopting the greedy decoding strategy. This is clearly shown in the row indicated by "\textbf{Ours (Greedy)}" in Table \ref{tab:ta4}. Our results using the sampling strategy (indicated by "\textbf{Ours (Sampling)}" in Table \ref{tab:ta4} are at least compatible to other results. Table \ref{tab:ta4} lists also the solution time (from test instances) of some methods. Due to the difference in implementation language and hardware, it is not meaningful for direct comparison of running time.

\begin{figure}[htbp]
	\centering
	\subfigure[TSP20]{
		\label{fig:fig5_a}
		\includegraphics[width=4cm]{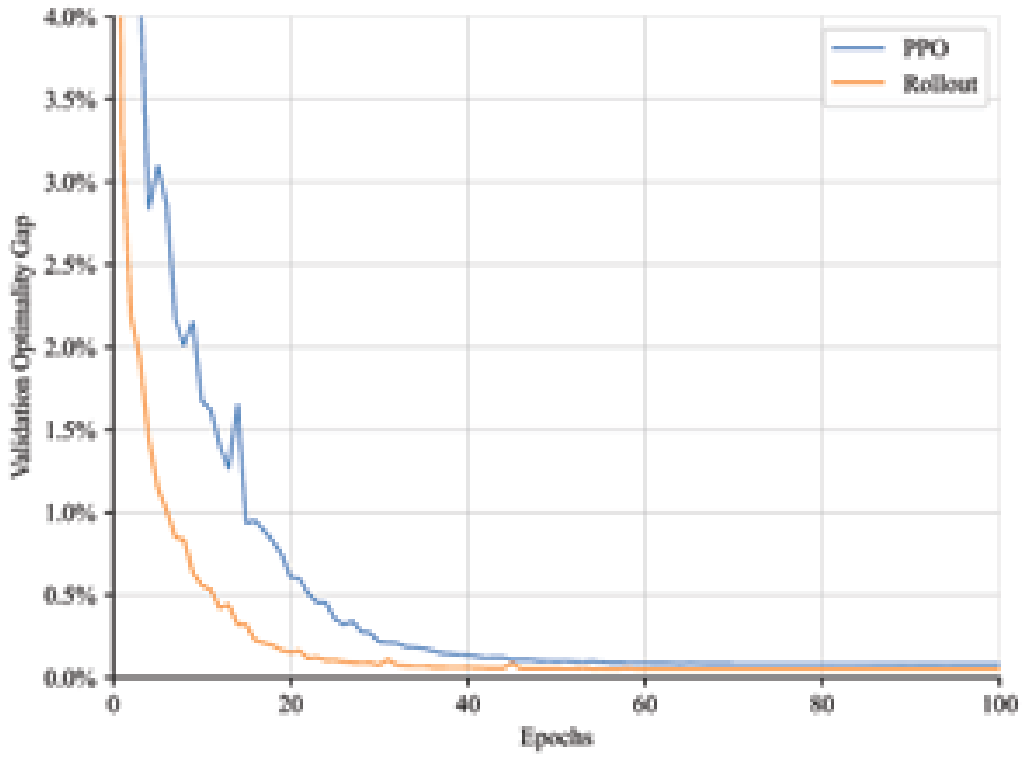}
	}
	\quad
	\subfigure[TSP50]{
		\label{fig:fig5_b}
		\includegraphics[width=4cm]{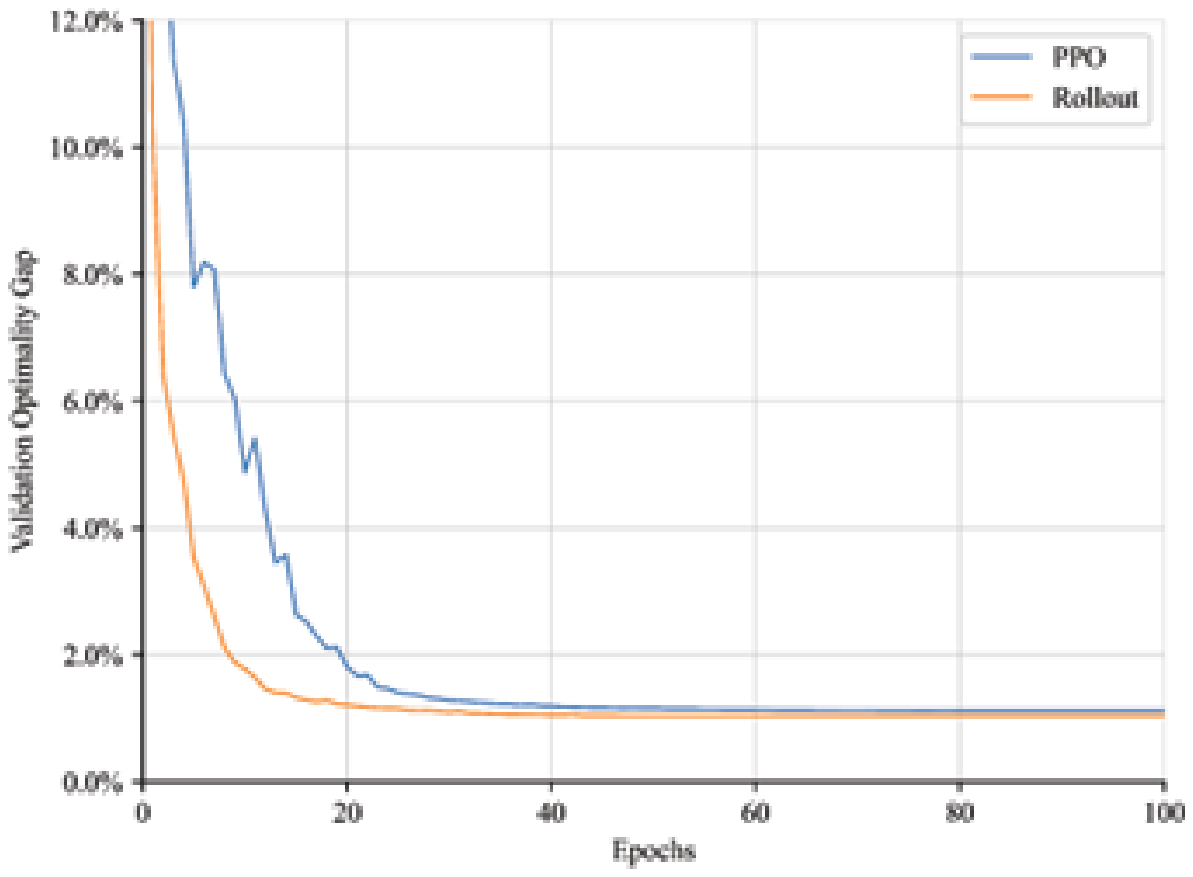}
	}
	\quad
	\subfigure[TSP100]{
		\label{fig:fig5_c}
		\includegraphics[width=4cm]{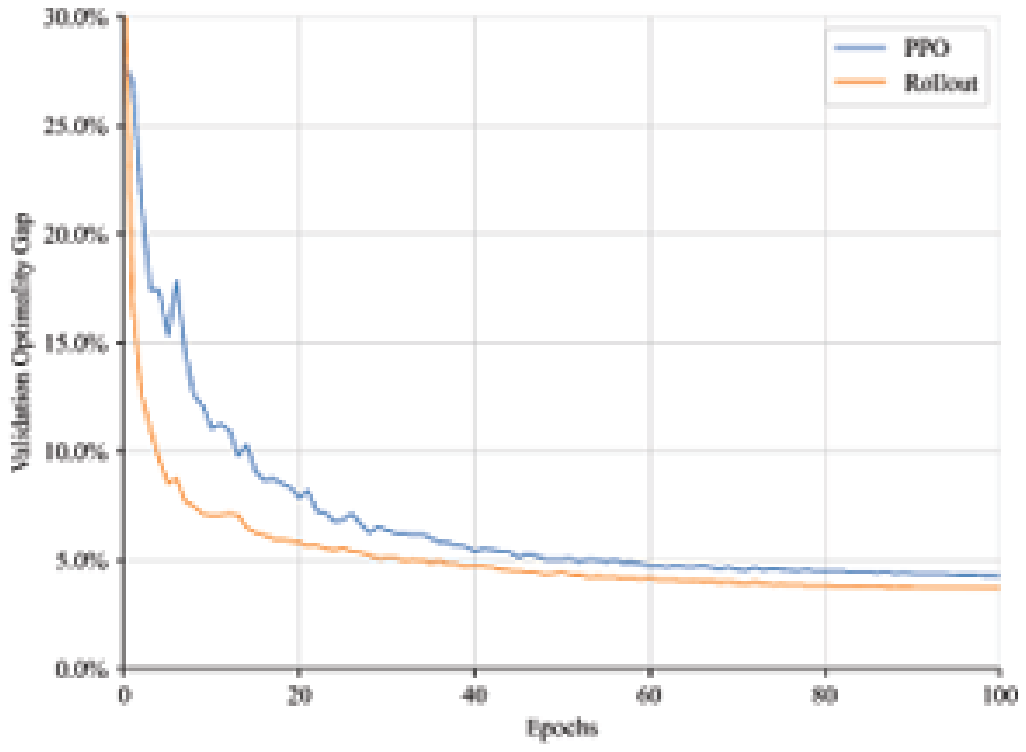}
	}
	\caption{The Opt. Gap convergency curves of PPO and Rollout on the validation set.}
	\label{fig:fig5}
\end{figure}

Figure \ref{fig:fig10} in \ref{app:b} shows the visualization of the solutions to the random TSP instances with 100 nodes through sampling and greedy decoding, respectively. Figure \ref{fig:fig5} shows the convergency of the PPO and Rollout two algorithms training the graph-attention model. Both use a validation set of size 10000 with greedy decoding. It can be seen from Figure \ref{fig:fig5} that the convergence trends are almost the same for the two methods. \textit{It is worth noting that the PPO only uses half of the training data used by the Rollout, but the Rollout converges faster.} In the TSP50 and TSP100 problems, the model proposed by Kool et al. \cite{R22} needs more than one million training samples. Our PPO and Rollout only use 384,000 and 768,000 training samples respectively, nevertheless the performance of our trained model is better than that of Kool et al. \cite{R22}.

\subsection{CVRP results}
CVRP is defined through an undirected graph $G=(V,E,W)$ where node $i=\{0,1,\cdots,m\}$ is represented by features $\bm{n}_i$. The
index $i=0$ represents the depot node, and ${i}{(i>0)}$ represents the i-th customer node. The vehicle has capacity $D>0$ and
each customer node $i=\{1,\cdots,m\}$  has a demand ${\delta}_i$, $0<{\delta}_i<D$. It is assumed that the depot demand ${\delta}_0=0$. Both depot and customer nodes are randomly generated from the unit square $[0,1]×[0,1]$. We generated 21, 51,
and 101 nodes (the first node is the depot) for problems with a size of $m=20,50,$ and 100, and the corresponding vehicle capacities are 30, 40, and 50, respectively. The customer node demands are sampled uniformly from $\{1,\cdots,9\}$. And we normalized the customer node demand to $[0,1]$ through ${\delta}_i^{'}=\frac{{\delta}_i}{10}$, so the vehicle capacity $D$ is transformed into 3, 4, and 5 accordingly. The input, masks and decoder context vectors are adjusted for CVRP:
\begin{itemize}
	\item 	Input: Switching to CVRP only needs to expand the node feature $\bm{n}_i$ to a three-dimensional input $\bm{n}_i^{'}$ that includes the normalized demand ${\delta}_i^{'}$ and node feature $\bm{n}_i$:
	\begin{equation}
	\bm{n}_i^{'}=\bm{n}_i{||}{\delta}_i^{'}
	\label{eq:eq26}
	\end{equation}
	\item 	Vehicle remaining capacity update: We masked out the customer nodes that have been served, so there is no need to update the needs of served customer nodes. The decoder selects the customer node $\hat{\pi}_{t}$ at timestep $t$, and the remaining capacity of the vehicle is represented by $D_t^{'}$. It is assumed that the vehicle starts at the depot when the decoding timestep $t=1$ and the vehicle is fully loaded (with vehicle remaining capacity $D_1^{'}=D$). The remaining vehicle capacity is updated by using the following Equation:
	\begin{equation}
	\label{eq:eq27}
	D_t^{'}=\left\{
	\begin{aligned}
	D_{t-1}^{'}-{\delta}_{\hat{\pi}_t}^{'},\hat{\pi}_t=i,i\in\{1,\cdots,m\} \\
	D ,  \hat{{\pi}}_{t}=0
	\end{aligned}
	.
	\right.
	\end{equation}
	\item 	Decoder context vector: The context vector $\bm{c}_t^{(0)}$ of the decoder at timestep ${t}\in\{1,\cdots,m\}$ consists of three parts: graph embedding $\bar{\bm{x}}$, embedding of node $\hat{\pi}_{t}$ and vehicle remaining capacity $D_{t-1}^{'}$:
	\begin{equation}
	\label{eq:eq28}
	\bm{c}_t^{(0)}=\left\{
	\begin{aligned}
	\bar{\bm{x}}+\bm{W}_x{(x_{\hat{\pi}_{t-1}}^{(L)}||D_{t-1}^{'})},t>1 \\
	\bar{\bm{x}}+\bm{W}_x{(x_{0}^{(L)}||D_{t}^{'})},  t=1
	\end{aligned}
	\right.
	.
	\end{equation}
	\item Mask update: For CVRP, our mask consists of two parts: the customer node mask and the depot node mask. For the customer node mask, we mask out a customer node that has been served or when the demand of a customer node is greater than the remaining capacity of the vehicle. That is, customer node $i$'s attention coefficients $u_{i,t}^{(0)},u_{i,t}^{(1)}=-\infty$ (introduced in Section \ref{se:3.2}) when ${\delta}_i^{'}>D_{t-1}^{'}$ or $i\neq\hat{\pi}_{t^{'}}\;\forall{t}^{'}<t,i\in\{1,\cdots,m\}$. For the depot mask, as the depot will not be the next chosen node when the vehicle leaves the depot, the depot node $0$'s attention coefficients $u_{0,t}^{(0)},u_{0,t}^{(1)}=-\infty$ for $t=1$ or $\hat{\pi}_{t-1}=0$, when the vehicle is at the depot.
\end{itemize}

The test results of CVRP instances with various sizes are listed in Table \ref{tab:ta5}. Our results again are obtained by applying our trained graph attention model to the CVRP testing instances using the Rollout Algorithm due to the same reason. Except for ours, the results in this Table are taken from Table 1 \cite{R22}. The structure and symbol description in the Table are identical to those introduced in Table \ref{tab:ta4} in Section \ref{se:se5.1}. \textit{Our solution approaches outperform other listed learning based algorithms in terms of solution quality.} Figure \ref{fig:fig11} in \ref{app:b} shows the visualization of the solutions to the random CVRP instances with 100 nodes through sampling and greedy decoding, respectively.
\begin{table}[htbp]
	\tiny
	\centering
	\caption{Performance of our framework versus state-of-the-art methods for CVRP instances with various sizes.}
	\begin{tabular}{lc|ccc|ccc|ccc}
		\toprule
		\multirow{2}[2]{*}{Method} & \multirow{2}[2]{*}{Type} & \multicolumn{3}{c|}{VRP20} & \multicolumn{3}{c|}{VRP50} & \multicolumn{3}{c}{VRP100} \\
		\multicolumn{1}{r}{} & \multicolumn{1}{r|}{} & \multicolumn{1}{c}{Length} & \multicolumn{1}{c}{Gap} & Time & \multicolumn{1}{c}{Length} & \multicolumn{1}{c}{Gap} & Time & \multicolumn{1}{c}{Length} & \multicolumn{1}{c}{Gap} & Time \\
		\midrule
		Gurobi & Solver & 6.10 & 0.00\% & \multicolumn{1}{c|}{} &     & \multicolumn{1}{c}{-} & \multicolumn{1}{c|}{} &     & \multicolumn{1}{c}{-} & \multicolumn{1}{c}{} \\
		LKH3 & Solver & 6.14 & 0.58\% & 2h  & 10.38 & 0.00\% & 7h  & 15.65 & 0.00\% & 13h \\
		\midrule
		PtrNet \cite{R19} & RL, G & 6.59 & 8.03\% & \multicolumn{1}{c|}{} & 11.39 & 9.78\% & \multicolumn{1}{c|}{} & 17.23 & 10.12\% & \multicolumn{1}{c}{} \\
		AM \cite{R22} & RL, G & 6.40 & 4.97\% & 1S  & 10.98 & 5.86\% & 3S  & 16.80 & 7.34\% & 8s \\
		\textbf{Ours (Greedy)} & \textbf{RL, G} & \textbf{6.26} & \textbf{2.60\%} & \textbf{2s} & \textbf{10.80} & \textbf{4.05\%} & \textbf{7s} & \textbf{16.69} & \textbf{6.68\%} & \textbf{17s} \\
		\midrule
		OR Tools & H, S & 6.43 & 5.41\% & \multicolumn{1}{c|}{} & 11.31 & 9.01\% & \multicolumn{1}{c|}{} & 17.16 & 9.67\% & \multicolumn{1}{c}{} \\
		PtrNet \cite{R19} & SL, BS & 6.40 & 4.92\% & \multicolumn{1}{c|}{} & 11.15 & 7.46\% & \multicolumn{1}{c|}{} & 16.96 & 8.39\% & \multicolumn{1}{c}{} \\
		AM \cite{R22} & RL, S & 6.25 & 2.49\% & 6m  & 10.62 & 2.40\% & 28m & 16.23 & 3.72\% & 2h \\
		\textbf{Ours (Sampling)} & \textbf{RL, S} & \textbf{6.19} & \textbf{1.47\%} & \textbf{14m} & \textbf{10.54} & \textbf{1.54\%} & \textbf{1h} & \textbf{16.16} & \textbf{3.25\%} & \textbf{4h} \\
		\bottomrule
		\multicolumn{11}{p{52em}}{Note: DL/DRL approaches are named according to the type of neural network used. In the Type column, H: heuristic method; SL: supervised learning; S: sample search; G: greedy search; BS: beam search; BS*: beam search and shortest path heuristics; 2OPT: 2OPT local search.}\\
	\end{tabular}%
	\label{tab:ta5}%
\end{table}%
\subsection{Testing running time complexity analysis}
\begin{figure}[htbp]
	\centering
	\includegraphics[width=3.5in]{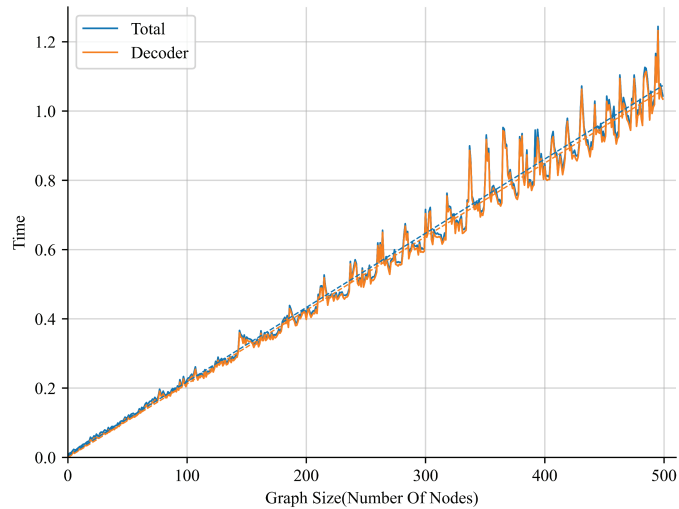}
	\caption{The running time vs. TSP size (number of nodes) increased from 1 to 500.}
	\label{fig:f6}
\end{figure}

We next report the running times of the entire encoder-decoder model (graph-attention model) and the decoder model only when the graph size (number of nodes) increases from 1 to 500. Each test is run with a single batch (batch size=1). Figure \ref{fig:f6} shows that our framework has linear testing time complexity on graph size (number of nodes). The running time mainly is consumed by the decoder. The encoder generally completes the embedding of nodes quickly (within 1ms). Table \ref{tab:ta6} summarizes the running time complexity, running time (average), acceleration factor, and optimal gap of exact algorithms, heuristic algorithms, and learning-based methods for TSP with 100 nodes.  All those results except ours are taken from Table 1 of Drori et al. \cite{R27}. Our results are listed in bold. The running time of exact algorithms, approximate algorithms, and heuristic algorithms increases at least squarely in the graph size. S2V-DQN \cite{R9} and GPN \cite{R24} are both reinforcement learning methods, and their running time complexity is $O{(n^2 )}$ and $O(nlogn)$ respectively. In particular, they have a large Opt. Gap (8.4$\%$ and 8.6$\%$ respectively). \textit{The running time complexity of our method is $O(n)$, and both the running time and the Opt. Gap are significantly improved.} GAT \cite{R27} has the same running time complexity as our method, but its Opt. Gap is larger.
\begin{table}[htbp]
	\tiny
	\centering
	\caption{Comparison of running time for TSP.}
	\begin{tabular}{l|c|c|c|c}
		\toprule
		Method & \multicolumn{1}{c|}{Runtime Complexity} & \multicolumn{1}{c|}{Runtime (ms)} & \multicolumn{1}{c|}{Speedup} & \multicolumn{1}{c}{Opt. Gap} \\
		\midrule
		Gurobi (Exact) & \multicolumn{1}{c|}{NA} & \multicolumn{1}{c|}{3, 220} & \multicolumn{1}{c|}{2, 752.1} & 0.00\% \\
		Concorde (Exact) & \multicolumn{1}{c|}{NA} & 254.1 & 217.2 & 0.00\% \\
		\midrule
		Christofifides & $O{(n^3)}$   & \multicolumn{1}{c|}{5, 002} & \multicolumn{1}{c|}{4, 275.2} & 2.90\% \\
		LKH & $O{(n^{2.2})}$   & \multicolumn{1}{c|}{2, 879} & 2460.7 & 0.00\% \\
		2-opt & $O{(n^2)}$   & 30.08 & 25.7 & 9.70\% \\
		Farthest & $O{(n^2)}$   & 8.35 & 7.1 & 7.50\% \\
		Nearest & $O{(n^2)}$   & 9.35 & 8   & 24.50\% \\
		\midrule
		S2V-DQN [Dai et al., 2017]\cite{R9} & $O{(n^2)}$   & 61.72 & 52.8 & 8.40\% \\
		GPN [Ma et al., 2019]\cite{R24} & $O{(nlogn)}$   & 1.537 & 1.3 & 8.60\% \\
		GAT [Drori et al., 2020]\cite{R27} & $O{(n)}$   & 1.17 & 1   & 7.40\% \\
		\textbf{Ours (Greedy)} & \textbf{$\textbf{O{(n)}}$} & \textbf{1.06} & \textbf{1} & \textbf{3.70\%} \\
		\bottomrule
	\end{tabular}%
	\label{tab:ta6}%
\end{table}%
\subsection{Results comparison among our framework, Attention Model and Pointer Networks }
\begin{figure}[htbp]
	\centering
	\includegraphics[width=3.5in]{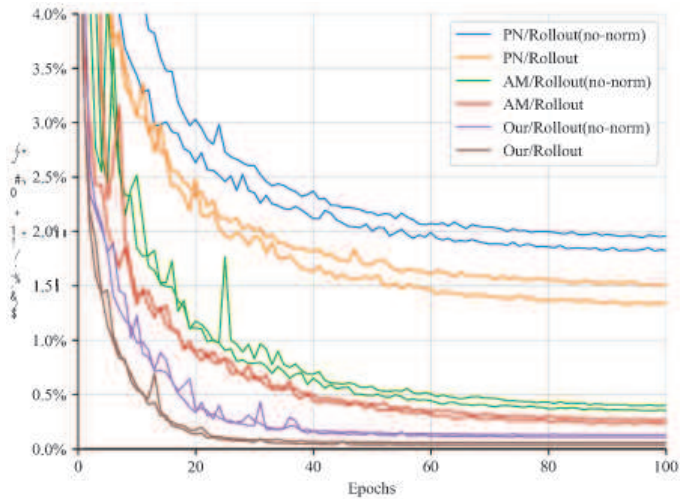}
	\caption{The validation Opt. Gap convergency curves of the attention model (AM), PtrNet (PN) and our model (Our).}
	\label{fig:f7}
\end{figure}

Figure \ref{fig:f7} compares the results of the TSP with 20 nodes on different models, including PtrNet (PN), attention model (AM) and our residual E-GAT (Our) during the training process. We used both the Rollout that employs a normalization of the reward function and its original version (Rollout (no-norm)) which does not normalize the reward function. We ran each experiment twice as did by Kool et al \cite{R22}, so there are two curves for each method in Figure \ref{fig:f7}. We reproduced the results of the PN and the AM with the Rollout (no-norm) reported by Kool et al \cite{R22}. All experiments are performed on a single Nvidia GeForce RTX2070 GPU. The training time for one epoch on a single Nvidia GeForce RTX2070 GPU are 503s for PN, 525s for AM, and 636s for Ours, respectively (For the same model, Rollout and Rollout (no-norm) have the same training time). For each training the annealing learning rate ${\eta}=10^{-3}×{0.96}^{epoch}$.

We reported the Opt. Gap of each epoch for each model on a validation set of 10000 with greedy decoding. The results show that our residual E-GAT is better than AM and PN for both Rollout and Rollout(no-norm) algorithms, in terms of convergence speed and solution quality. We also overserve that the Rollout with normalization of the reward function is better than Rollout (no-norm).

\subsection{Performance analysis of GAT, Residual GAT and Residual E-GAT}

\begin{figure}[htbp]
	\centering
	\includegraphics[width=3.5in]{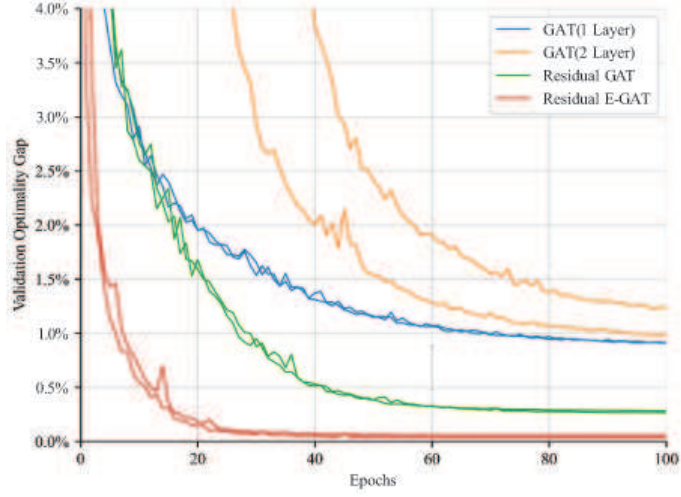}
	\caption{The validation Opt. Gap convergency curves of one-layer GAT, two-layer GAT, Residual GAT and Residual E-GAT during training.}
	\label{fig:f8}
\end{figure}

Table \ref{tab:ta7} shows the results of TSP with 20 nodes of GAT as the number of layers increases. It can be found that the reward cannot converge when the number of GAT layers is more than 2. To remedy this problem, we employed a residual connection (as described in Section \ref{se:se3.1}) for each GAT sub-layers. The resulting variant of GAT is called Residual GAT. Figure \ref{fig:f8} compared the performance of one-layer GAT, two-layer GAT, Residual GAT and Residual E-GAT during training. All these models were implemented on the same machine. Residual GAT and Residual E-GAT contain four layers, respectively. We used the annealing learning rate ${\eta}=3×{10}^{-4}×{0.96}^{epoch}$. This performs best for the one-layer GAT and the two-layers GAT, while ${\eta}=10^{-3}×{0.96}^{epoch}$ are used for the Residual GAT and Residual E-GAT. There is an obvious improvement when adding residual connection to the GAT. This improvement could be caused by the fact that a deeper GAT is powerful to capture the information among the graphs but more difficult to train. Therefore, we introduced the residual connection to ease the training of the deeper GAT, a different approach than those used previously \cite{R42}. Furthermore, the Residual E-GAT has a further improvement relative to Residual GAT. Compared to the best result of the GAT, the Residual E-GAT reduces the Opt. Gap from 0.85$\%$ to 0.06$\%$.
\begin{table}[htbp]
	\tiny
	\centering
	\caption{The results of GAT using different layers.}
	\begin{tabular}{cccccc}
		\toprule
		\multicolumn{1}{p{8em}}{Number of Layers} & \multicolumn{1}{r}{1} & \multicolumn{1}{r}{2} & \multicolumn{1}{r}{3} & \multicolumn{1}{r}{5} & \multicolumn{1}{r}{8} \\
		\midrule
		\multicolumn{1}{c}{\multirow{2}[4]{*}{Length (Opt. Gap)}} & 3.86(0.86\%) & 3.87(1.09\%) & 10.41(171\%) & 10.41(171\%) & 10.41(171\%) \\
		\cmidrule{2-6}        & 3.86(0.85\%) & 3.88(1.31\%) & 10.41(171\%) & 10.41(171\%) & 10.41(171\%) \\
		\bottomrule
	\end{tabular}%
	\label{tab:ta7}%
\end{table}%

\subsection{The generalization performance of our framework }
\label{se:se5.6}
We used small and medium-sized instances from the public libraries TSPLIB \cite{R53} and CVRPLIB \cite{R54} to verify the generalization performance from random instance training to real-world instance testing. The instance suffix for an instance in TSPLIB indicates the number of nodes. Table \ref{tab:ta8} shows the results of exact algorithm, reinforcement learning methods and approximate algorithms. Except for ours, the results in this Table are taken from Table C.2 \cite{R8}. S2V-DQN \cite{R9} is a reinforcement learning method that uses the Active Search proposed by Bello et al. \cite{R18} to solve the instance problems in TSPLIB. Its generalization performance does not depend on the distribution of the training data. Different than greedy decoding with a fixed model, Active Search can refine the parameters of the stochastic policy $p_{\theta}$ during testing on a single test data. Therefore, solving through Active Search consumes a lot of time. Instead of using Active Search, we used greedy decoding in training our model with the 100-node random instances for the purpose of testing the instances in TSPLIB. We saved the model parameters at the end of each epoch, which constitute a set of 100 models ($epoch=100$). For each instance, the tour found by each individual model is collected and the shortest tour is chosen. We refer to this approach as \textit{trained-greedy100}.

\begin{table}[htbp]
	\tiny
	\centering
	\caption{Comparison of different approaches to solve the instances in TSPLIB.}
	\begin{tabular}{cl|cc|ccccccc}
		\toprule
		TSPLIB & \multicolumn{1}{c|}{Exact} & \multicolumn{2}{c|}{RL} & \multicolumn{7}{c}{Approx.} \\
		Instance & \multicolumn{1}{c|}{Concord} & \multicolumn{1}{p{2.5em}}{Ours} & \multicolumn{1}{c|}{S2V-DQN} & \multicolumn{1}{p{2.5em}}{Farthest} & \multicolumn{1}{p{2.5em}}{2-opt} & \multicolumn{1}{p{2.5em}}{Cheapest} & \multicolumn{1}{p{2.5em}}{Closet} & \multicolumn{1}{p{4em}}{Christofides} & \multicolumn{1}{p{2.5em}}{Nearest} & \multicolumn{1}{p{2.5em}}{MST} \\
		\midrule
		eil51 & 426 & \textbf{428} & 439 & 467 & 446 & 494 & 488 & 527 & 511 & 614 \\
		berlin52 & 7,542 & 7,802 & \textbf{7,542} & 8,307 & 7,788 & 9,013 & 9,004 & 8,822 & 8,980 & 10,402 \\
		st70 & 675 & \textbf{681} & 696 & 712 & 753 & 776 & 814 & 836 & 801 & 858 \\
		eil76 & 538 & \textbf{555} & 564 & 583 & 591 & 607 & 615 & 646 & 705 & 743 \\
		pr76 & 108,159 & 111,036 & \textbf{108,446} & 119,692 & 115,460 & 125,935 & 128,381 & 137,258 & 153,462 & 133,471 \\
		rat99 & 1,211 & 1,304 & \textbf{1,280} & 1,314 & 1,390 & 1,473 & 1,465 & 1,399 & 1,558 & 1,665 \\
		kroA100 & 21,282 & 22,372 & \textbf{21,897} & 23,356 & 22,876 & 24,309 & 25,787 & 26,578 & 26,854 & 30,516 \\
		kroB100 & 22,141 & 23,961 & \textbf{22,692} & 23,222 & 23,496 & 25,582 & 26,875 & 25,714 & 29,158 & 28,807 \\
		kroC100 & 20,749 & 22,002 & \textbf{21,074} & 21,699 & 23,445 & 25,264 & 25,640 & 24,582 & 26,327 & 27,636 \\
		kroD100 & 21,294 & 22,642 & 22,102 & \textbf{22,034} & 23,967 & 25,204 & 25,213 & 27,863 & 26,947 & 28,599 \\
		kroE100 & 22,068 & 23,162 & 22,913 & 23,516 & \textbf{22,800} & 25,900 & 27,313 & 27,452 & 27,585 & 30,979 \\
		rd100 & 7,910 & \textbf{7,930} & 8,159 & 8,944 & 8,757 & 8,980 & 9,485 & 10,002 & 9,938 & 10,467 \\
		eil101 & 629 & \textbf{652} & 659 & 673 & 702 & 693 & 720 & 728 & 817 & 847 \\
		lin105 & 14,379 & \textbf{14,932} & 15,023 & 15,193 & 15,536 & 16,930 & 18,592 & 16,568 & 20,356 & 21,167 \\
		pr107 & 44,303 & 46,415 & \textbf{45,113} & 45,905 & 47,058 & 52,816 & 52,765 & 49,192 & 48,521 & 55,956 \\
		pr124 & 59,030 & \textbf{60,487} & 61,623 & 65,945 & 64,765 & 65,316 & 68,178 & 64,591 & 69,297 & 82,761 \\
		bier127 & 118,282 & 124,387 & \textbf{121,576} & 129,495 & 128,103 & 141,354 & 145,516 & 135,134 & 129,333 & 153,658 \\
		ch130 & 6,110 & \textbf{6,203} & 6,270 & 6,498 & 6,470 & 7,279 & 7,434 & 7,367 & 7,578 & 8,280 \\
		pr136 & 96,772 & \textbf{98,481} & 99,474 & 105,361 & 110,531 & 109,586 & 105,778 & 116,069 & 120,769 & 142,438 \\
		pr144 & 58,537 & 60,844 & \textbf{59,436} & 61,974 & 60,321 & 73,032 & 73,613 & 74,684 & 61,652 & 77,704 \\
		ch150 & 6,528 & \textbf{6,825} & 6,985 & 7,210 & 7,232 & 7,995 & 7,914 & 7,641 & 8,191 & 9,203 \\
		kroA150 & 26,524 & 28,077 & \textbf{27,888} & 28,658 & 29,666 & 29,963 & 32,631 & 31,341 & 33,612 & 38,763 \\
		kroB150 & 26,130 & 27,431 & \textbf{27,209} & 27,404 & 29,517 & 31,589 & 33,260 & 31,616 & 32,825 & 35,289 \\
		pr152 & 73,682 & 77,125 & \textbf{75,283} & 75,396 & 77,206 & 88,531 & 82,118 & 86,915 & 85,699 & 90,292 \\
		u159 & 42,080 & \textbf{44,273} & 45,433 & 46,789 & 47,664 & 49,986 & 48,908 & 52,009 & 53,641 & 54,399 \\
		\bottomrule
		Opt. Gap & 0.00\% & 4.10\% & \textbf{3.50\%} & 7.70\% & 8.40\% & 15.50\% & 17.70\% & 17.80\% & 21.40\% & 31.40\% \\
	\end{tabular}%
	\label{tab:ta8}%
\end{table}%

For small sized instances (customer $m<75$) and medium size instances (customer $m\ge75$) in CVRLIB, we used the trained models by 50 and 100 nodes random instances to test them, respectively. Table \ref{tab:ta9} and Table \ref{tab:ta10} show the optimal solutions and our graph-attention model results for instances in CVRPLIB. We listed the Opt. Gap for each instance and the average of all instances.

The average Opt. Gap are 4.1$\%$, 6.39$\%$ and 10.45$\%$ for TSP, small and medium sized CVRP benchmark instances, respectively. The results verify that our framework has generalization performance from random instance training to real-world instance testing. Finally, Figure \ref{fig:fig12} and Figure \ref{fig:fig13} in \ref{app:c} show the visualized solutions of instances in TSPLIB and CVRPLIB with different sizes.

\begin{table}[htbp]
	\tiny
	\centering
	\caption{Trained by random instances with 50 nodes and solved the instances with different sizes in CVRPLIB by trained-greedy100.}
	\begin{tabular}{ccccccccc}
		\toprule
		\multirow{2}[3]{*}{No.} & \multicolumn{1}{c}{\multirow{2}[3]{*}{Problem}} & \multicolumn{1}{c}{\multirow{2}[3]{*}{Nodes}} & \multicolumn{1}{c}{\multirow{2}[3]{*}{Capacity}} & \multicolumn{1}{c}{\multirow{2}[3]{*}{Routes}} & \multicolumn{1}{c}{\multirow{2}[3]{*}{Optimal}} & \multicolumn{2}{c}{Ours (Greedy)} & \multicolumn{1}{c}{\multirow{2}[3]{*}{Opt. Gap}} \\
		\cmidrule{7-8}    \multicolumn{1}{l}{} &     &     &     &     &     & \multicolumn{1}{c}{Routes} & \multicolumn{1}{c}{Distance} &  \\
		\midrule
		\multicolumn{1}{l}{1} & \multicolumn{1}{c}{A-n32-k5} & 31  & 100 & 5   & 784 & 5   & 789 & 0.63\% \\
		\multicolumn{1}{l}{2} & \multicolumn{1}{c}{A-n36-k5} & 35  & 100 & 5   & 799 & 5   & 839 & 5.00\% \\
		\multicolumn{1}{l}{3} & \multicolumn{1}{c}{A-n37-k5} & 36  & 100 & 5   & 669 & 5   & 710 & 6.12\% \\
		\multicolumn{1}{l}{4} & \multicolumn{1}{c}{A-n38-k5} & 37  & 100 & 5   & 730 & 5   & 751 & 2.87\% \\
		\multicolumn{1}{l}{5} & \multicolumn{1}{c}{A-n39-k5} & 38  & 100 & 5   & 822 & 5   & 838 & 1.94\% \\
		\multicolumn{1}{l}{6} & \multicolumn{1}{c}{A-n44-k6} & 43  & 100 & 6   & 937 & 6   & 984 & 5.01\% \\
		\multicolumn{1}{l}{7} & \multicolumn{1}{c}{A-n45-k6} & 44  & 100 & 6   & 944 & 6   & 984 & 4.23\% \\
		\multicolumn{1}{l}{8} & \multicolumn{1}{c}{A-n46-k7} & 45  & 100 & 7   & 914 & 7   & 999 & 9.29\% \\
		\multicolumn{1}{l}{9} & \multicolumn{1}{c}{A-n48-k7} & 47  & 100 & 7   & 1,073 & 7   & 1,123 & 4.65\% \\
		\multicolumn{1}{l}{10} & \multicolumn{1}{c}{A-n63-k10} & 62  & 100 & 10  & 1,314 & 12  & 1,519 & 15.60\% \\
		\multicolumn{1}{l}{11} & \multicolumn{1}{c}{A-n64-k9} & 63  & 100 & 9   & 1,401 & 12  & 1,659 & 18.40\% \\
		\multicolumn{1}{l}{12} & \multicolumn{1}{c}{A-n65-k9} & 64  & 100 & 9   & 1,174 & 10  & 1,246 & 6.13\% \\
		\multicolumn{1}{l}{13} & \multicolumn{1}{c}{A-n69-k9} & 68  & 100 & 9   & 1,159 & 10  & 1,264 & 9.05\% \\
		\multicolumn{1}{l}{14} & \multicolumn{1}{c}{B-n34-k5} & 33  & 100 & 5   & 788 & 5   & 812 & 3.04\% \\
		\multicolumn{1}{l}{15} & \multicolumn{1}{c}{B-n35-k5} & 34  & 100 & 5   & 955 & 5   & 986 & 3.24\% \\
		\multicolumn{1}{l}{16} & \multicolumn{1}{c}{B-n45-k6} & 44  & 100 & 6   & 678 & 7   & 729 & 7.52\% \\
		\multicolumn{1}{l}{17} & \multicolumn{1}{c}{B-n50-k7} & 49  & 100 & 7   & 741 & 7   & 798 & 7.69\% \\
		\multicolumn{1}{l}{18} & \multicolumn{1}{c}{B-n51-k7} & 50  & 100 & 7   & 1,032 & 8   & 1,045 & 1.25\% \\
		\multicolumn{1}{l}{19} & \multicolumn{1}{c}{B-n52-k7} & 51  & 100 & 7   & 747 & 7   & 762 & 2.00\% \\
		\multicolumn{1}{l}{20} & \multicolumn{1}{c}{B-n56-k7} & 55  & 100 & 7   & 707 & 7   & 755 & 6.79\% \\
		\multicolumn{1}{l}{21} & \multicolumn{1}{c}{B-n57-k9} & 56  & 100 & 9   & 1,598 & 9   & 1,638 & 2.50\% \\
		\multicolumn{1}{l}{22} & \multicolumn{1}{c}{B-n63-k10} & 62  & 100 & 10  & 1,496 & 12  & 1,667 & 11.43\% \\
		\multicolumn{1}{l}{23} & \multicolumn{1}{c}{B-n64-k9} & 63  & 100 & 9   & 861 & 11  & 989 & 14.86\% \\
		\multicolumn{1}{l}{24} & \multicolumn{1}{c}{B-n66-k9} & 65  & 100 & 9   & 1,316 & 9   & 1,381 & 4.93\% \\
		\multicolumn{1}{l}{25} & \multicolumn{1}{c}{B-n68-k9} & 67  & 100 & 9   & 1,272 & 9   & 1,397 & 9.82\% \\
		\multicolumn{1}{l}{26} & \multicolumn{1}{c}{E-n30-k3} & 29  & 4500 & 3   & 534 & 5   & 585 & 9.55\% \\
		\multicolumn{1}{l}{27} & \multicolumn{1}{c}{E-n51-k5} & 50  & 160 & 5   & 521 & 6   & 553 & 6.14\% \\
		\multicolumn{1}{l}{28} & \multicolumn{1}{c}{P-n50-k8} & 49  & 120 & 8   & 631 & 9   & 655 & 3.80\% \\
		\multicolumn{1}{l}{29} & \multicolumn{1}{c}{P-n51-k10} & 50  & 80  & 10  & 741 & 10  & 811 & 9.44\% \\
		\multicolumn{1}{l}{30} & \multicolumn{1}{c}{P-n55-k10} & 56  & 115 & 10  & 694 & 10  & 726 & 4.61\% \\
		\multicolumn{1}{l}{31} & \multicolumn{1}{c}{P-n60-k10} & 59  & 120 & 10  & 744 & 10  & 776 & 4.30\% \\
		\multicolumn{1}{l}{32} & \multicolumn{1}{c}{P-n65-k10} & 64  & 130 & 10  & 792 & 10  & 826 & 4.29\% \\
		\multicolumn{1}{l}{33} & \multicolumn{1}{c}{P-n70-k10} & 69  & 135 & 10  & 827 & 11  & 865 & 4.59\% \\
		\bottomrule
		Average  &     &     &     &     &     &     &     & 6.39\% \\
	\end{tabular}%
	\label{tab:ta9}%
\end{table}%
\begin{table}[htbp]
	\tiny
	\centering
	\caption{Trained by random instances 100 nodes and solved the instances with different sizes in CVRPLIB by trained-greedy100.}
	\begin{tabular}{p{4.19em}rcccccccc}
		\toprule
		\multirow{2}[4]{*}{No.} & \multicolumn{2}{c}{\multirow{2}[4]{*}{Problem}} & \multicolumn{1}{c}{\multirow{2}[4]{*}{Nodes}} & \multicolumn{1}{c}{\multirow{2}[4]{*}{Capacity}} & \multicolumn{1}{c}{\multirow{2}[4]{*}{Routes}} & \multicolumn{1}{c}{\multirow{2}[4]{*}{Optimal}} & \multicolumn{2}{c}{Ours (Greedy)} & \multicolumn{1}{c}{\multirow{2}[4]{*}{Opt. Gap}} \\
		\cmidrule{8-9}    \multicolumn{1}{l}{} & \multicolumn{2}{c}{} &     &     &     &     & \multicolumn{1}{p{4.19em}}{Routes} & \multicolumn{1}{p{4.19em}}{Distance} &  \\
		\midrule
		\multicolumn{1}{l}{1} & \multicolumn{2}{c}{A-n80-k10} & 79  & 100 & 10  & 1,763 & 10  & 1,890 & 7.20\% \\
		\multicolumn{1}{l}{2} & \multicolumn{2}{c}{B-n78-k10} & 77  & 100 & 10  & 1,221 & 10  & 1,333 & 9.17\% \\
		\multicolumn{1}{l}{3} & \multicolumn{2}{c}{E-n76-k10} & 75  & 140 & 10  & 830 & 10  & 870 & 4.81\% \\
		\multicolumn{1}{l}{4} & \multicolumn{2}{c}{E-n101-k14} & 100 & 112 & 14  & 1,067 & 14  & 1,163 & 8.96\% \\
		\multicolumn{1}{l}{5} & \multicolumn{2}{c}{M-n101-k10} & 100 & 200 & 10  & 820 & 10  & 873 & 6.46\% \\
		\multicolumn{1}{l}{6} & \multicolumn{2}{c}{M-n121-k7} & 120 & 200 & 7   & 1,034 & 8   & 1,144 & 10.63\% \\
		\multicolumn{1}{l}{7} & \multicolumn{2}{c}{M-n151-k12} & 150 & 200 & 12  & 1,015 & 12  & 1,106 & 8.96\% \\
		\multicolumn{1}{l}{8} & \multicolumn{2}{c}{M-n200-k17} & 199 & 200 & 17  & 1,275 & 17  & 1,414 & 10.90\% \\
		\multicolumn{1}{l}{9} & \multicolumn{2}{c}{P-n76-k5} & 75  & 280 & 5   & 627 & 5   & 690 & 10.04\% \\
		\multicolumn{1}{l}{10} & \multicolumn{2}{c}{P-n101-k4} & 100 & 400 & 4   & 681 & 4   & 820 & 20.41\% \\
		\multicolumn{1}{l}{11} & \multicolumn{2}{c}{X-n106-k14} & 105 & 600 & 14  & 26,362 & 14  & 28,085 & 6.53\% \\
		\multicolumn{1}{l}{12} & \multicolumn{2}{c}{X-n125-k30} & 124 & 188 & 30  & 55,539 & 30  & 59,147 & 6.49\% \\
		\multicolumn{1}{l}{13} & \multicolumn{2}{c}{X-n134-k13} & 133 & 643 & 13  & 10,916 & 13  & 12,342 & 13.06\% \\
		\multicolumn{1}{l}{14} & \multicolumn{2}{c}{X-n148-k46} & 147 & 18  & 46  & 43,448 & 46  & 50,462 & 16.14\% \\
		\multicolumn{1}{l}{15} & \multicolumn{2}{c}{X-n157-k13} & 156 & 12  & 13  & 16,876 & 13  & 20,273 & 20.12\% \\
		\multicolumn{1}{l}{16} & \multicolumn{2}{c}{X-n181-k23} & 180 & 8   & 23  & 25,569 & 23  & 28,566 & 11.72\% \\
		\multicolumn{1}{l}{17} & \multicolumn{2}{c}{X-n200-k36} & 199 & 402 & 36  & 58,578 & 36  & 63,276 & 8.02\% \\
		\multicolumn{1}{l}{18} & \multicolumn{2}{c}{X-n223-k34} & 222 & 836 & 34  & 40,437 & 37  & 43,508 & 7.59\% \\
		\multicolumn{1}{l}{19} & \multicolumn{2}{c}{X-n251-k28} & 250 & 69  & 28  & 38,684 & 28  & 42,965 & 11.06\% \\
		\multicolumn{1}{l}{20} & \multicolumn{2}{c}{X-n266-k58} & 265 & 35  & 58  & 75,478 & 62  & 83,155 & 10.17\% \\
		\multicolumn{1}{l}{21} & \multicolumn{2}{c}{X-n298-k31} & 297 & 55  & 31  & 34,231 & 32  & 38,023 & 11.07\% \\
		\midrule
		\multicolumn{2}{c}{Average } &     &     &     &     &     &     &     & 10.45\% \\
	\end{tabular}%
	\label{tab:ta10}%
\end{table}%
\section{Conclusion}
\label{se:se6}
In this research, a general deep reinforcement learning framework for solving routing problems is introduced. An encoder based on an improved GAT, which forms a graph-attention model with the Transformer decoder, is proposed. Two deep reinforcement learning algorithms: PPO and improved baseline REINFORCE algorithm, are used to train the model. The PPO has higher efficiency of sample utilization compared with the improved baseline REINFORCE algorithm, and both algorithms can achieve better performance with less than one million training data. Our framework significantly improves the results of solving TSP and CVRP problems using fewer training data compared with existing studies. Moreover, it can be found that our framework has linear running time complexity during the testing process. For real-world instances of TSPLIB and CVRPLIB, our framework exhibits similar performance as it does in random instances. This indicates our framework has generalization performance from random instance training to actual problem testing, even using fewer instances for training.

For the future research, PPO and improved baseline REINFORCE algorithm may be combined for finding a better DRL framework. Furthermore, the proposed framework is also possible to solve more complex VRP problems with extra constraints or other complex combinatorial optimization problems, such as job shop scheduling problems and flexible job shop scheduling problems.

\section*{CRediT authorship contribution statement}
\textbf{Kun LEI}: Conceptualization, Methodology, Investigation, Writing-original draft, Editing. \textbf{Peng GUO}: Conception and design of study, Analysis and interpretation of data, Writing - original draft, Writing - review $\&$ editing. \textbf{Yi WANG}: Validation, Writing - review $\&$ editing. \textbf{Xiao WU}: Writing - review $\&$ editing. \textbf{Wenchao ZHAO}: Methodology, Writing-original draft.
\section*{Declaration of Competing Interest}
The authors declare that they have no known competing financial interests or personal relationships that could have appeared to influence the work reported in this paper.
\section*{Acknowledgments}
The research of this paper is supported by the National Key Research and Development Plan (grant number 2020YFB1712200) and the Fundamental Research Funds for the Central Universities (grant number 2682018CX09).
\section*{Data availability}
The associated source code and the computational results are available at the Github website: \url{https://github.com/pengguo318/RoutingProblemsGANN}.

\section*{References}
\bibliographystyle{apalike}

\begin{appendices}
	\appendix
	\section{Sensitivity Analyses}
	\label{app:a}
	 \begin{figure}[htbp]
	 	\centering
	 	\subfigure[Dimension of initial node/edge embedding]{
	 		\label{fig:fig9_a}
	 		\includegraphics[width=6cm]{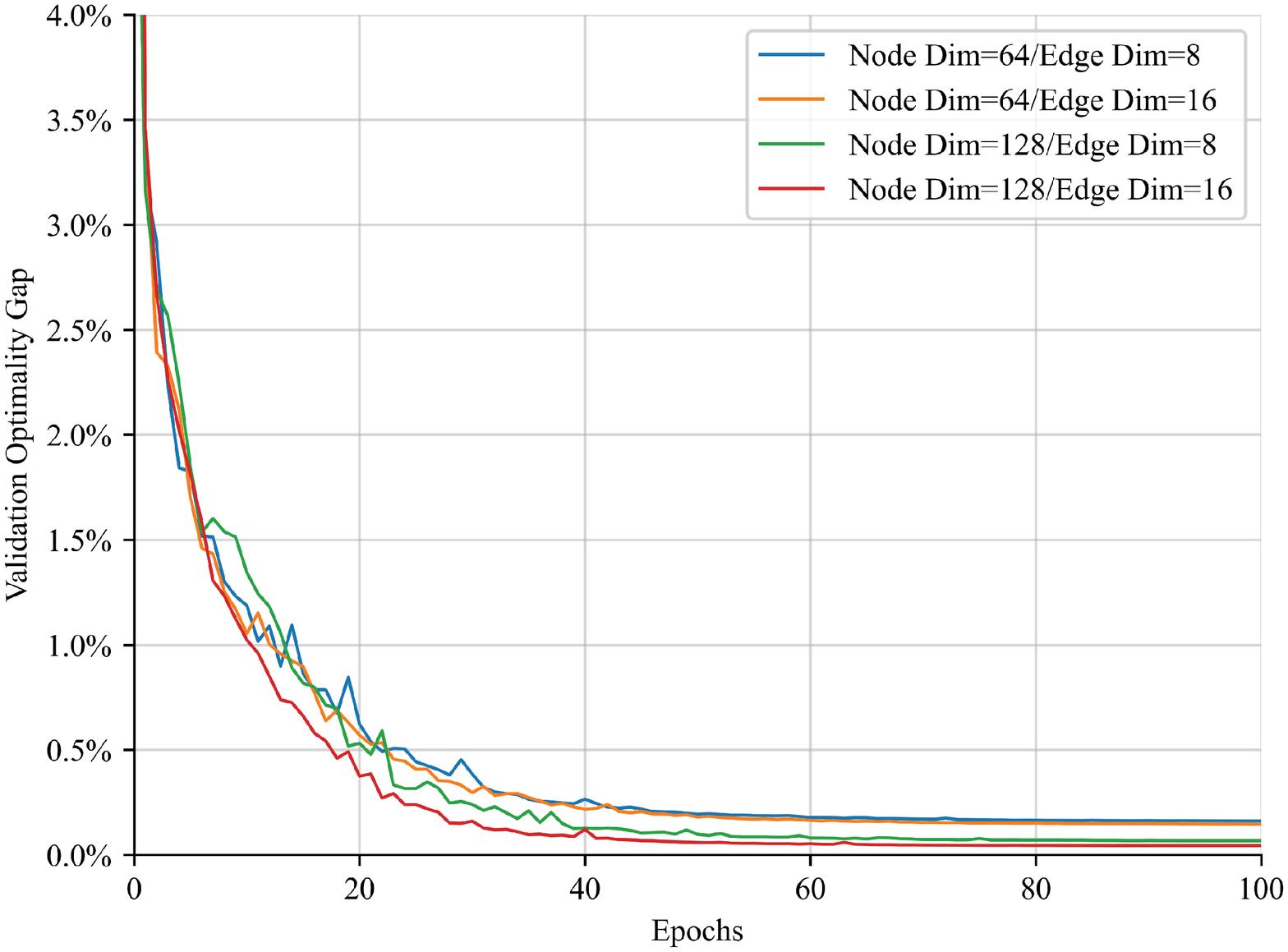}
	 	}
	 	\quad
	 	\subfigure[Number of encoder layers]{
	 		\label{fig:fig9_b}
	 		\includegraphics[width=6cm]{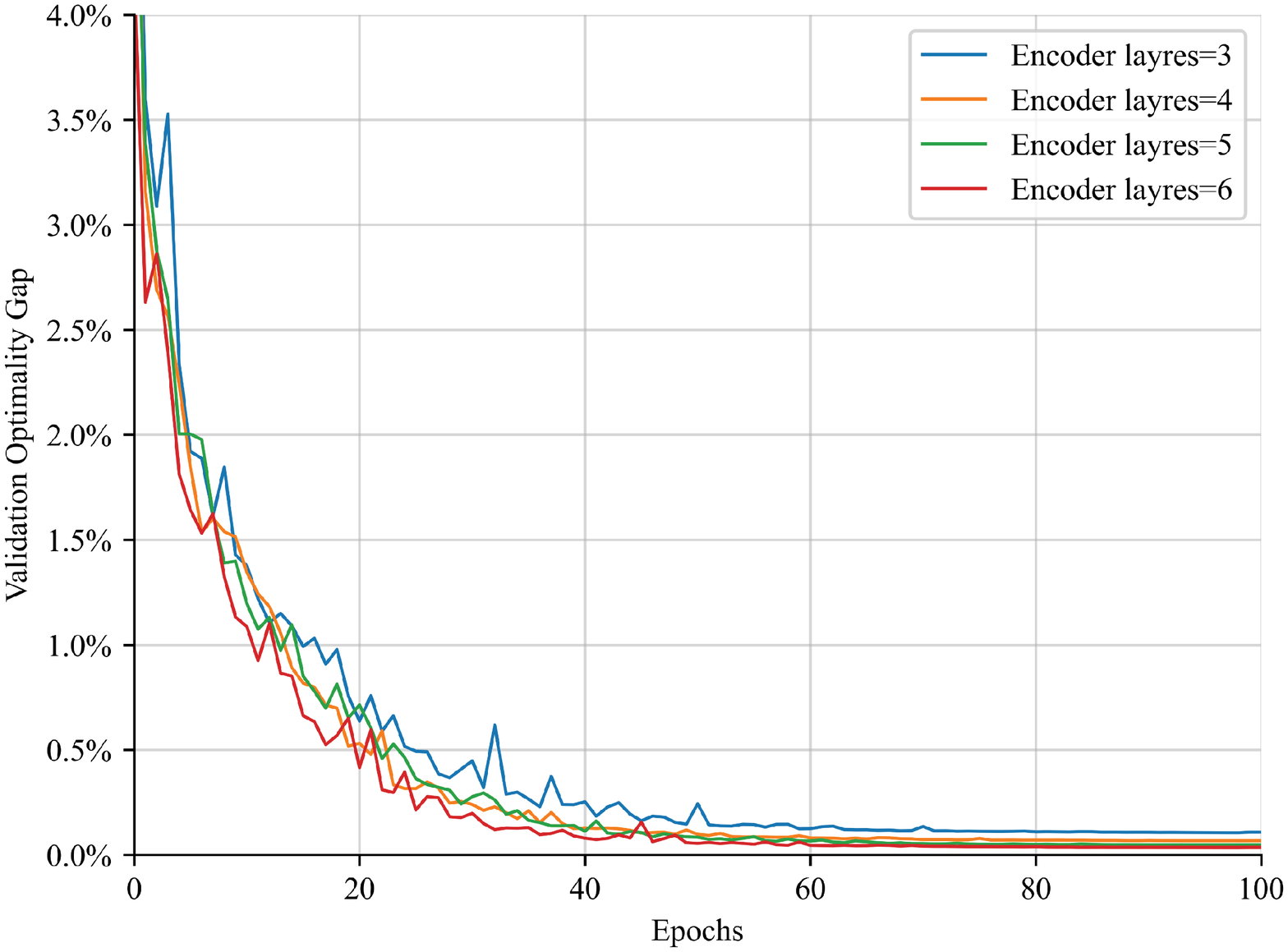}
	 	}
	 	\caption{Sensitivity analyses of hyperparameters in our graph-attention model.}
	 	\label{fig:fig9}
	 \end{figure}
	 In this section, we analyzed the sensitivities of hyper-parameters in the proposed framework, which could greatly influence the solution quality of our graph-attention model. Studied are three hyper-parameters including the dimension of the initial node embedding ($d_x$), the dimension of the initial edge embedding ($d_e$) and the number of the encoder layers ($L$). First, we re-trained the network by setting the embedding dimension of the node to be 64 or 128, and the embedding dimension of the edge as 8 or 16, respectively. Figure \ref{fig:fig9_a} shows the Opt. Gap convergency curves for TSP with 20 nodes. It can be found that the results tend to get better with increased embedding dimensions of the node and the edge. This finding arises because some useful information will be neglected in lower dimensional space, which leads to deteriorated algorithm output. Similarly, we tested the sensitivity of the encoder layers $L{(L={[3,4,5,6]})}$, and the Opt. Gap convergency curves are demonstrated in Figure \ref{fig:fig9_b}. The training time of an epoch on a single Nvidia GeForce RTX2070 GPU are 536s, 632s, 712s, and 805s, respectively. It can be seen that the model with three encoder layers achieved the worst performance. This finding confirmed that a shallow model makes it difficult to capture the information among the nodes and the edges in a graph structure. We found $L = 4$ is a good trade-off between quality of the results and computational complexity (training time) of the model.
	
	\section{Solutions of random instances}
	\label{app:b}
\begin{figure}[htbp]
		\centering
		\subfigure{
			\label{fig:fig101}
			\includegraphics[width=4.2cm]{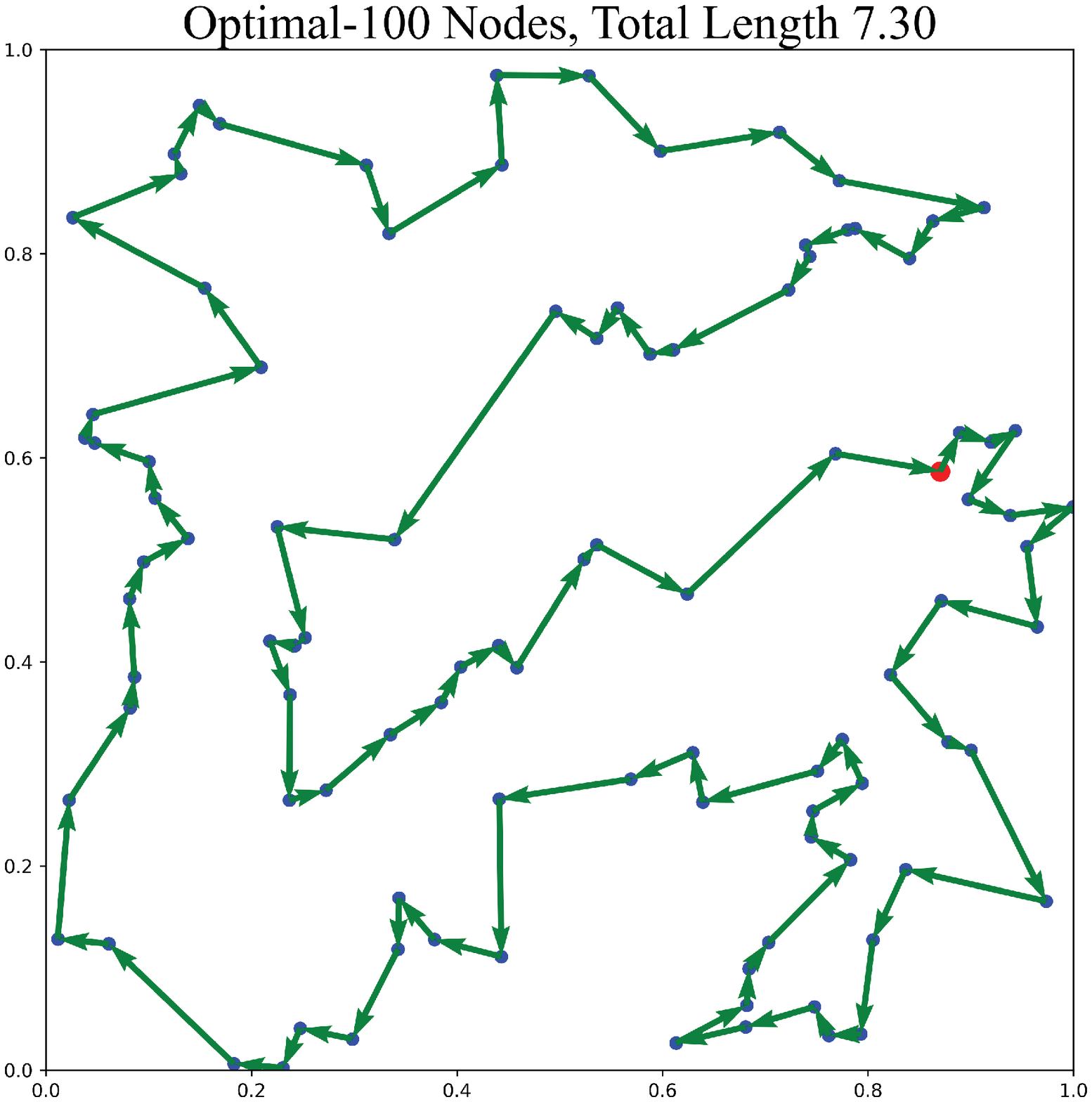}
		}
		\quad
		\subfigure{
			\label{fig:fig102}
			\includegraphics[width=4.2cm]{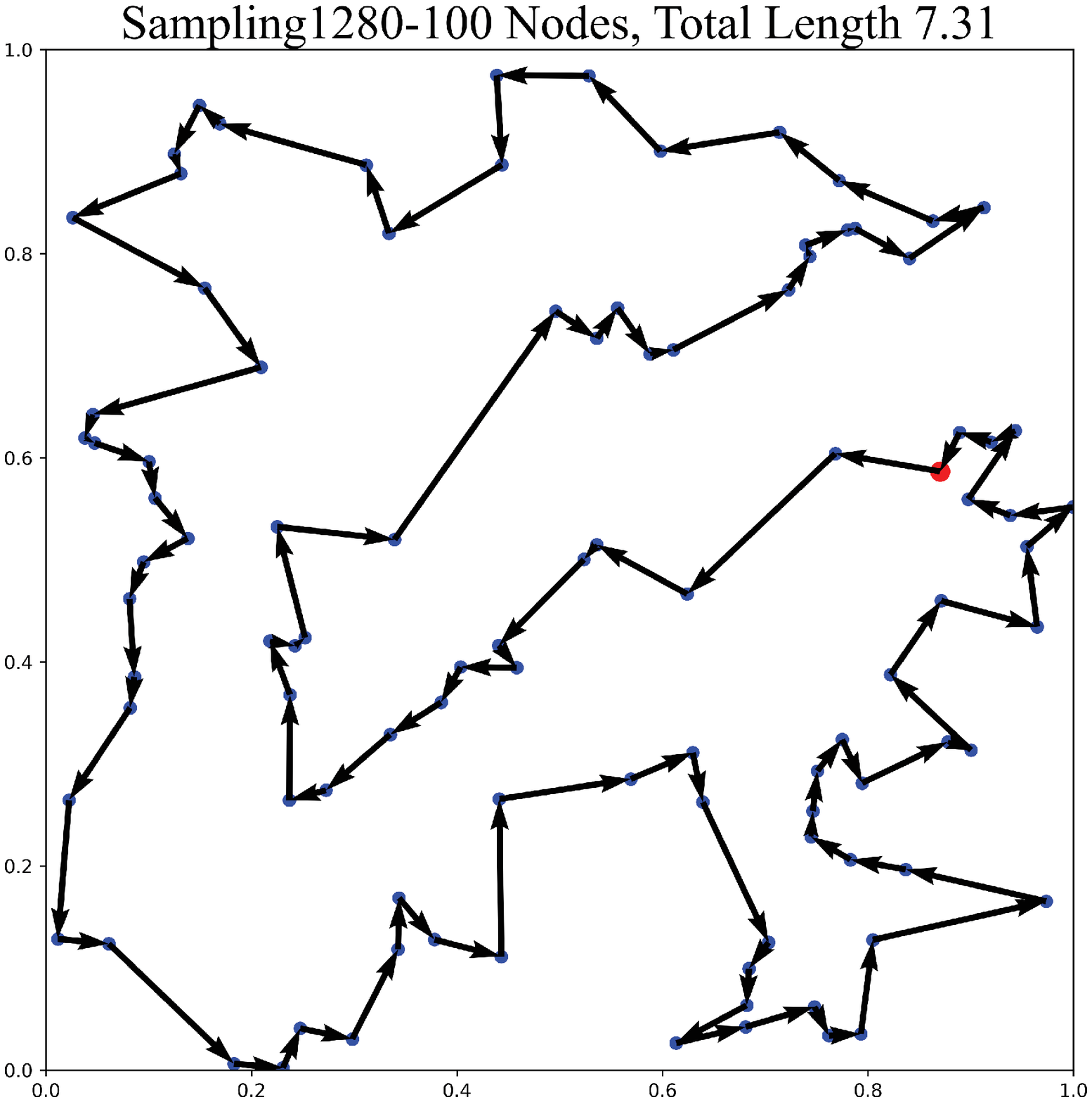}
		}
		\quad
		\subfigure{
			\label{fig:fig103}
			\includegraphics[width=4.2cm]{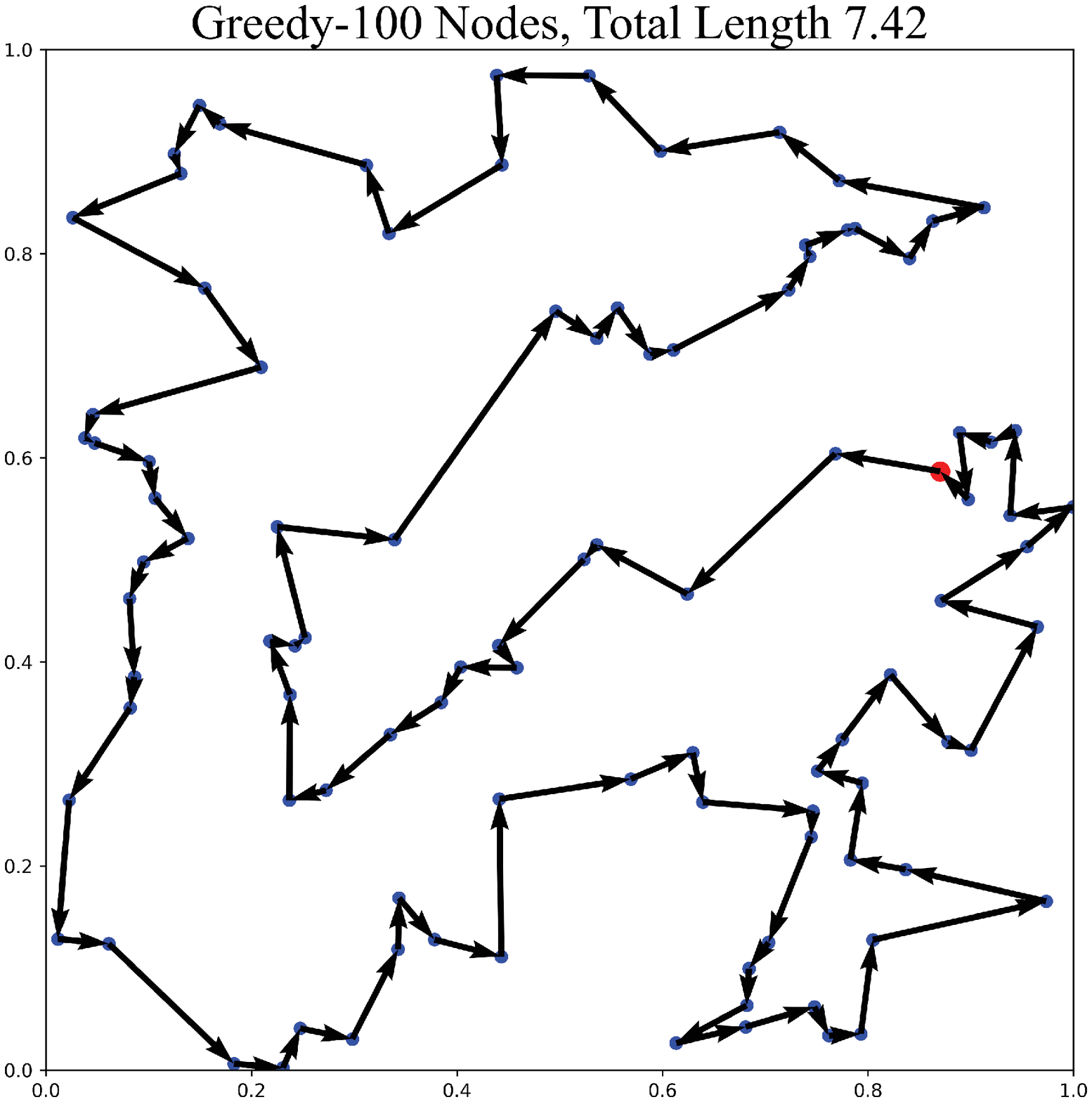}
		}
		\quad
		\subfigure{
			\label{fig:fig104}
			\includegraphics[width=4.2cm]{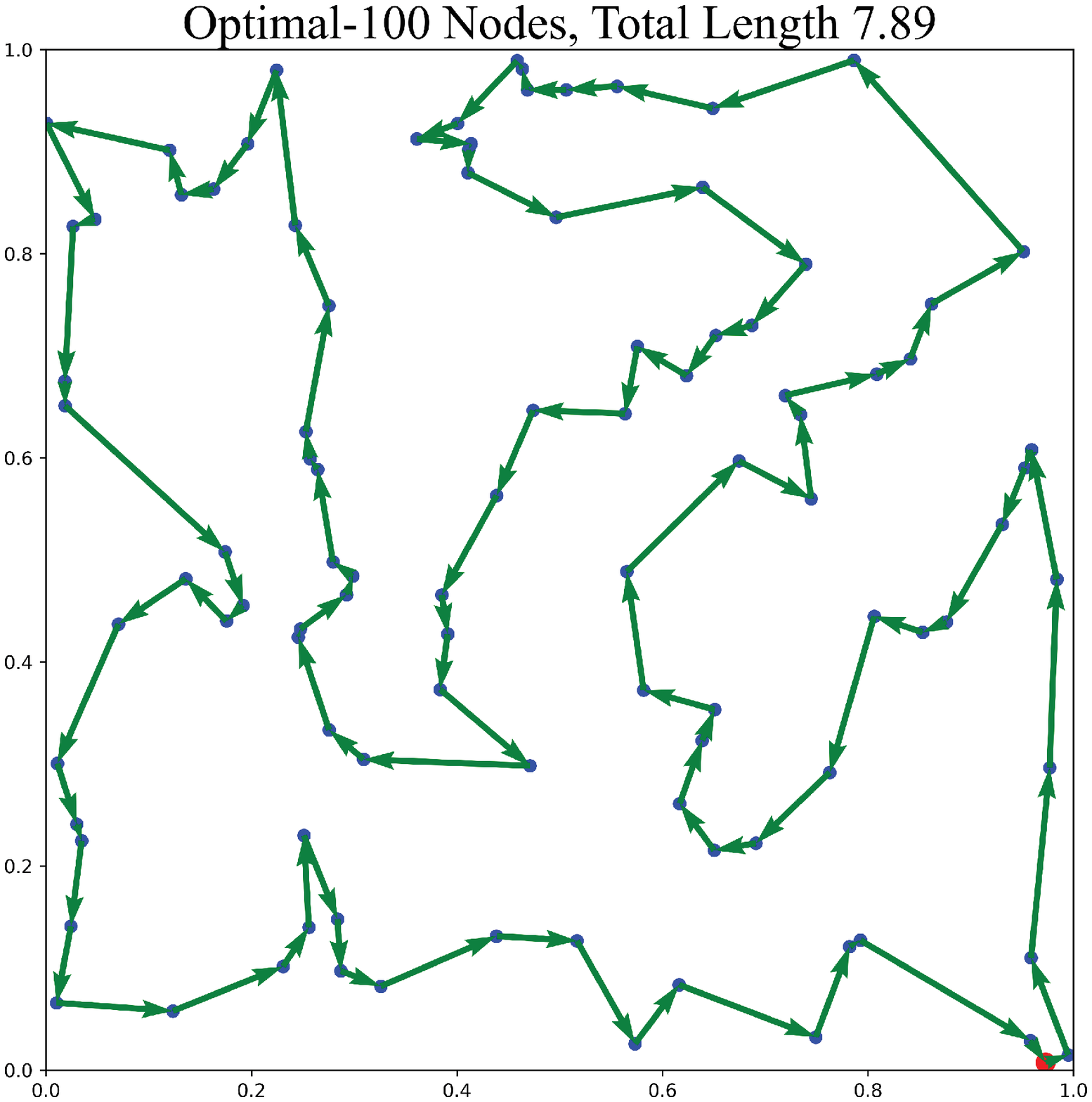}
		}
		\quad
		\subfigure{
			\label{fig:fig105}
			\includegraphics[width=4.2cm]{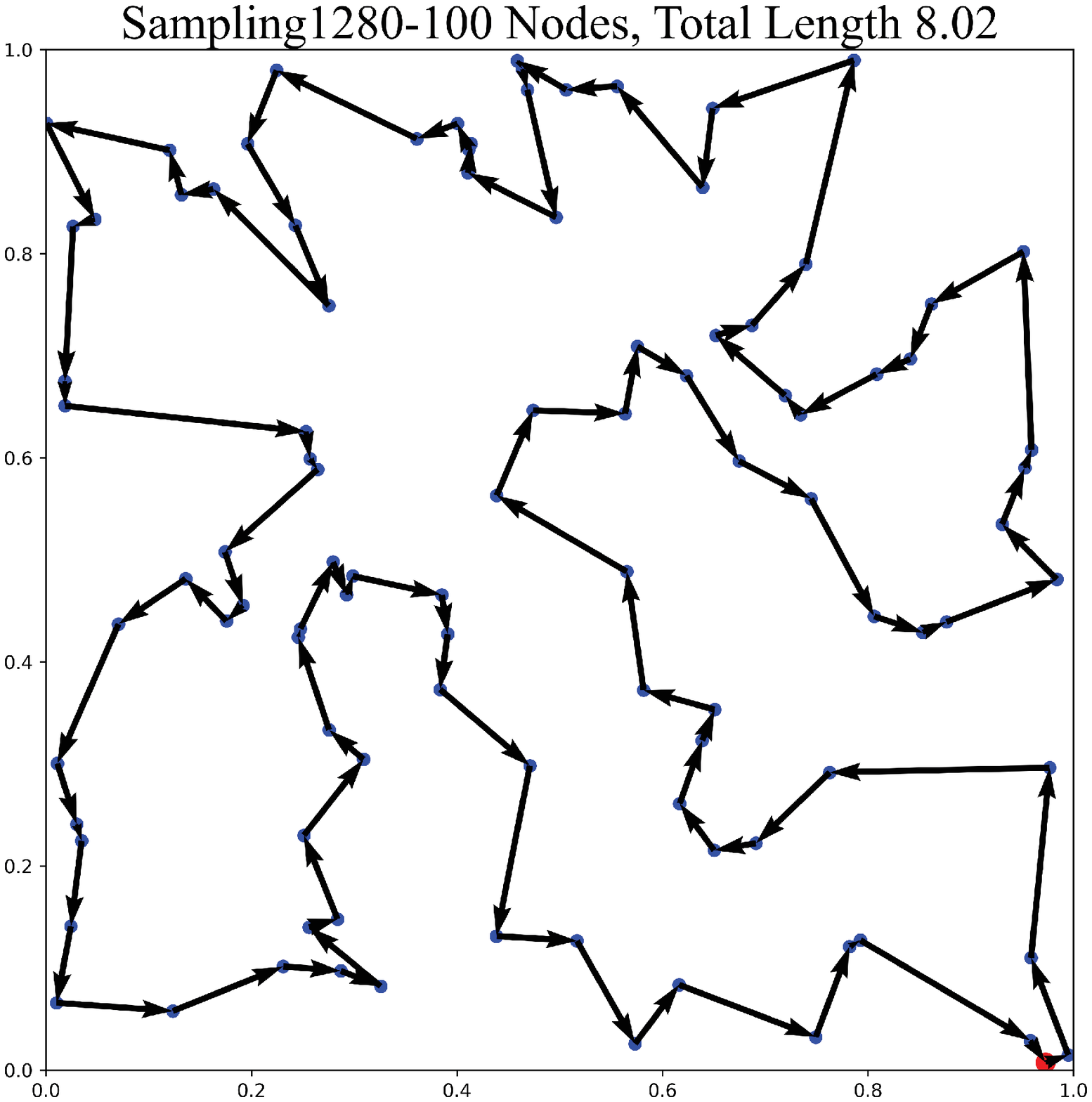}
		}
		\quad
		\subfigure{
			\label{fig:fig106}
			\includegraphics[width=4.2cm]{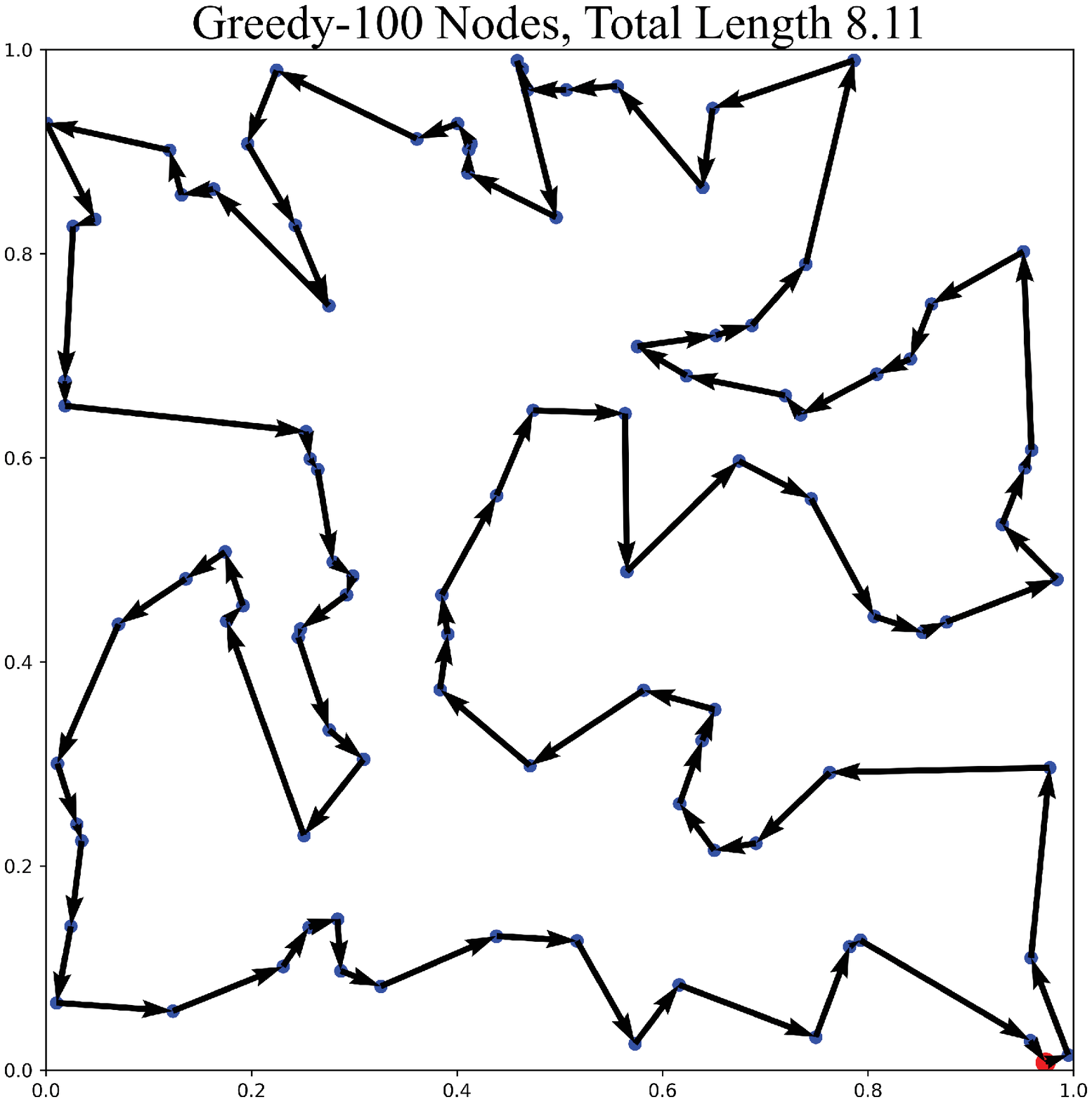}
		}
		\caption{testing a random TSP instance problem with 100 nodes through sampling and greedy decoding.}
		\label{fig:fig10}
	\end{figure}

\begin{figure}[htbp]
		\centering
		\subfigure{
			\label{fig:fig111}
			\includegraphics[width=4.2cm]{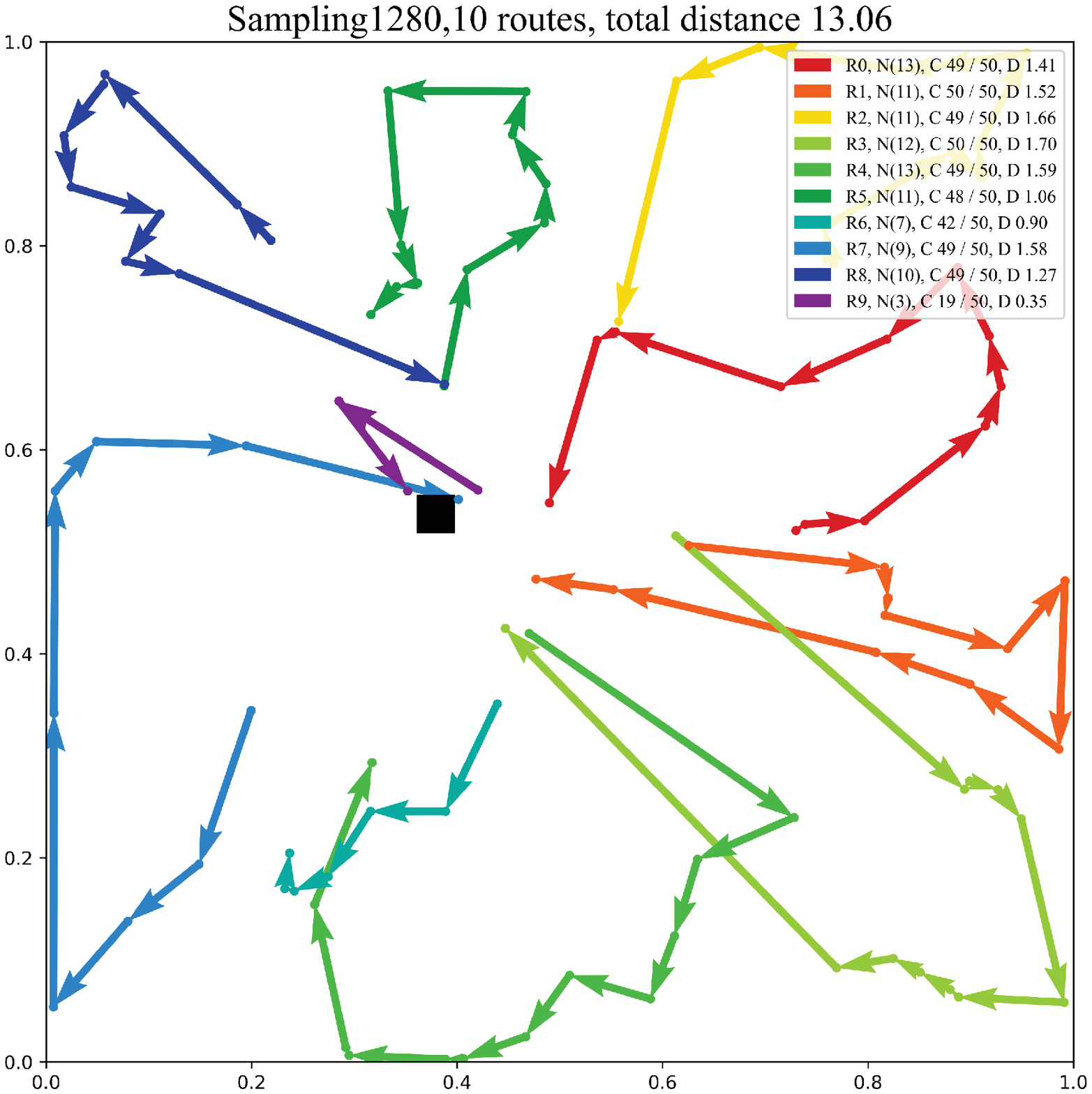}
		}
		\quad
		\subfigure{
			\label{fig:fig112}
			\includegraphics[width=4.2cm]{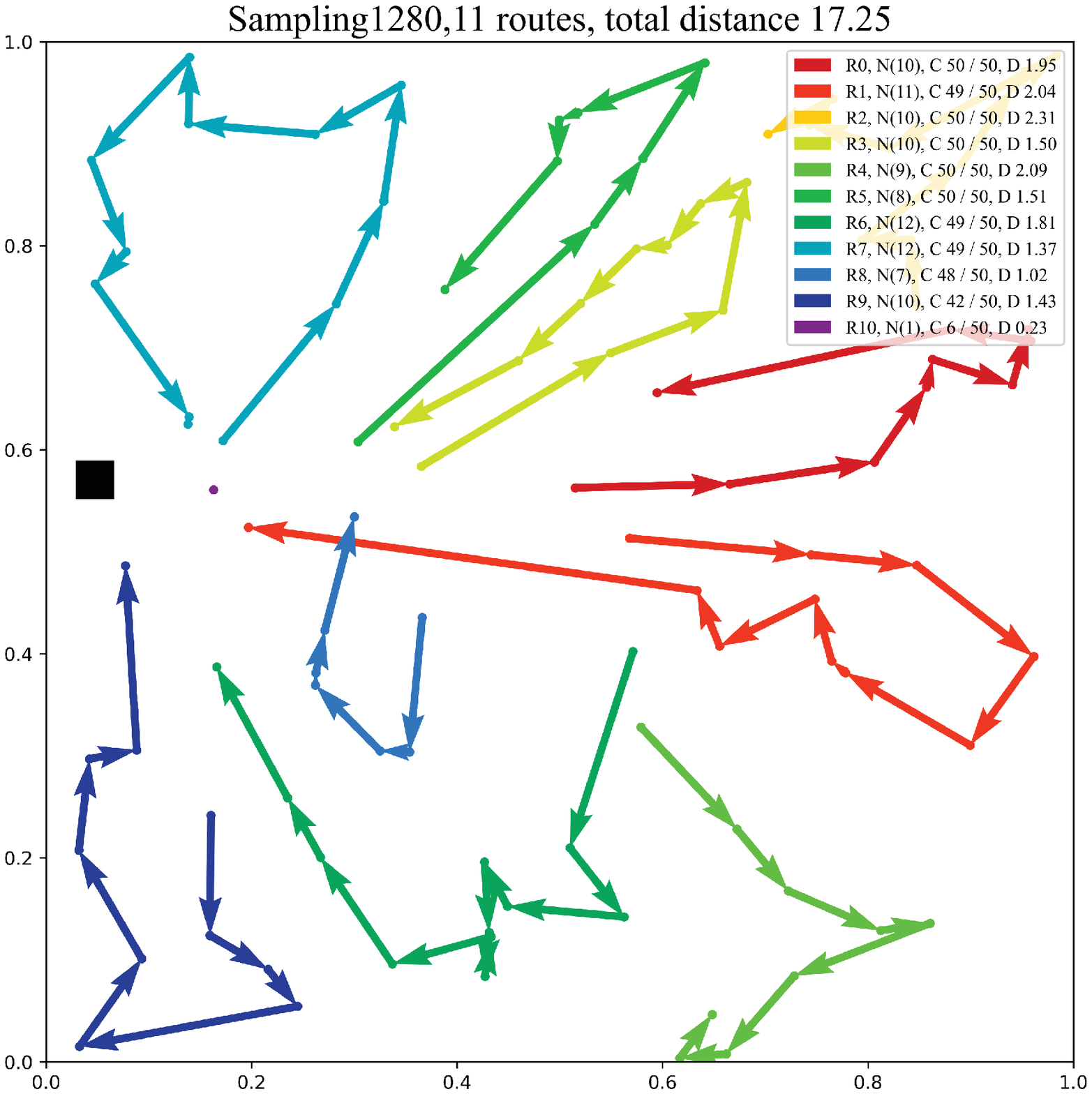}
		}
		\quad
		\subfigure{
			\label{fig:fig113}
			\includegraphics[width=4.2cm]{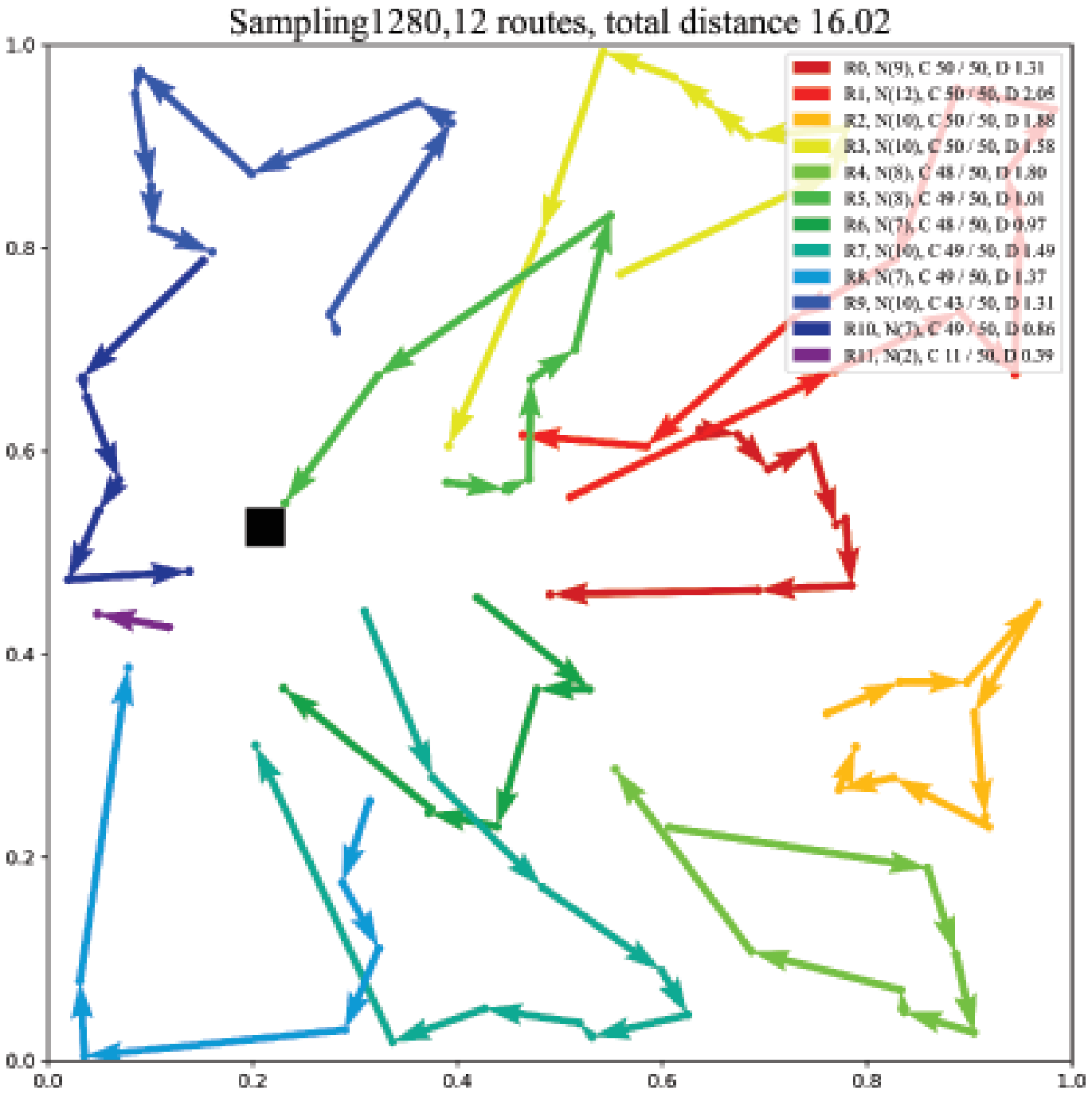}
		}
		\quad
		\subfigure{
			\label{fig:fig114}
			\includegraphics[width=4.2cm]{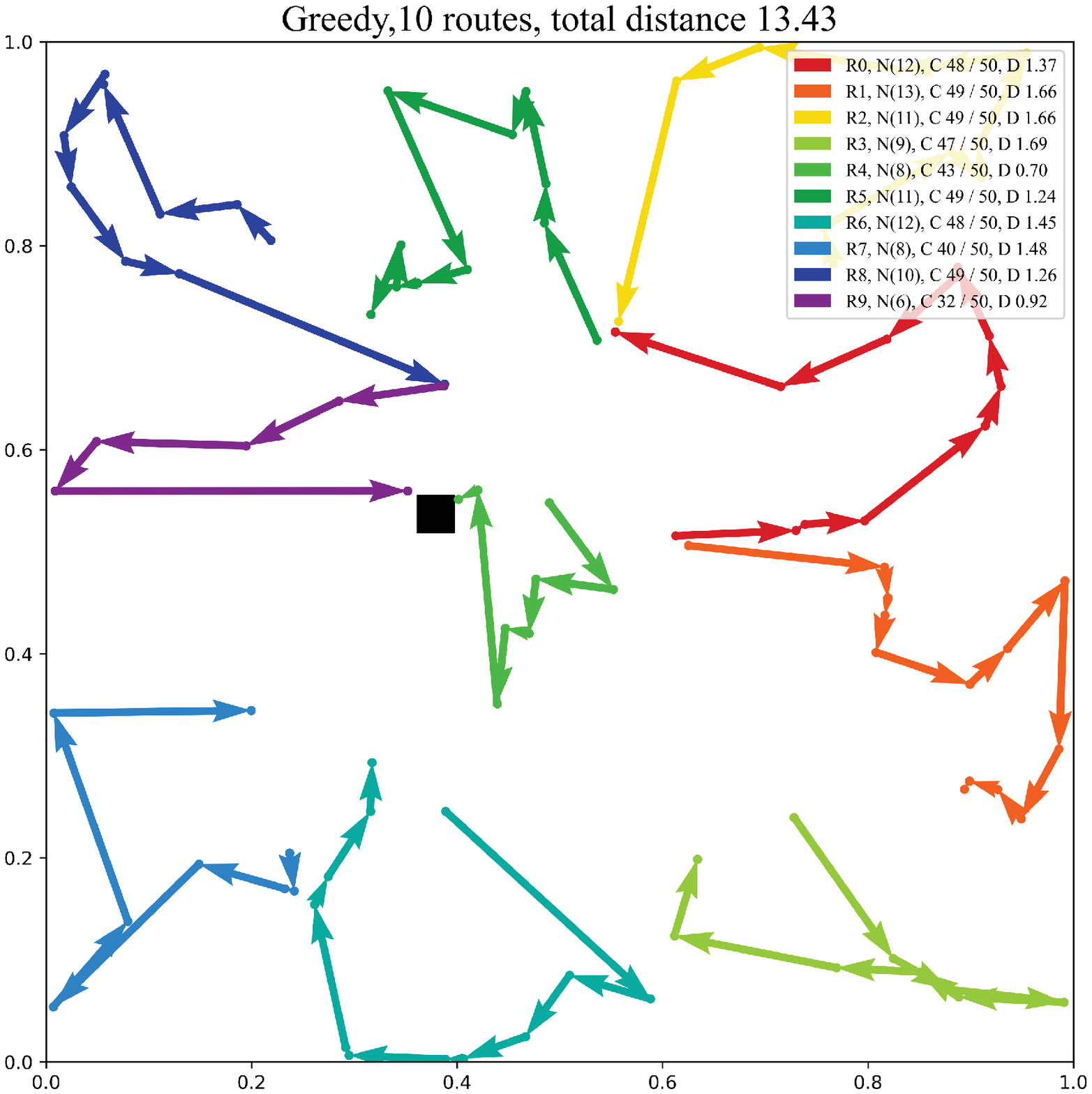}
		}
		\quad
		\subfigure{
			\label{fig:fig115}
			\includegraphics[width=4.2cm]{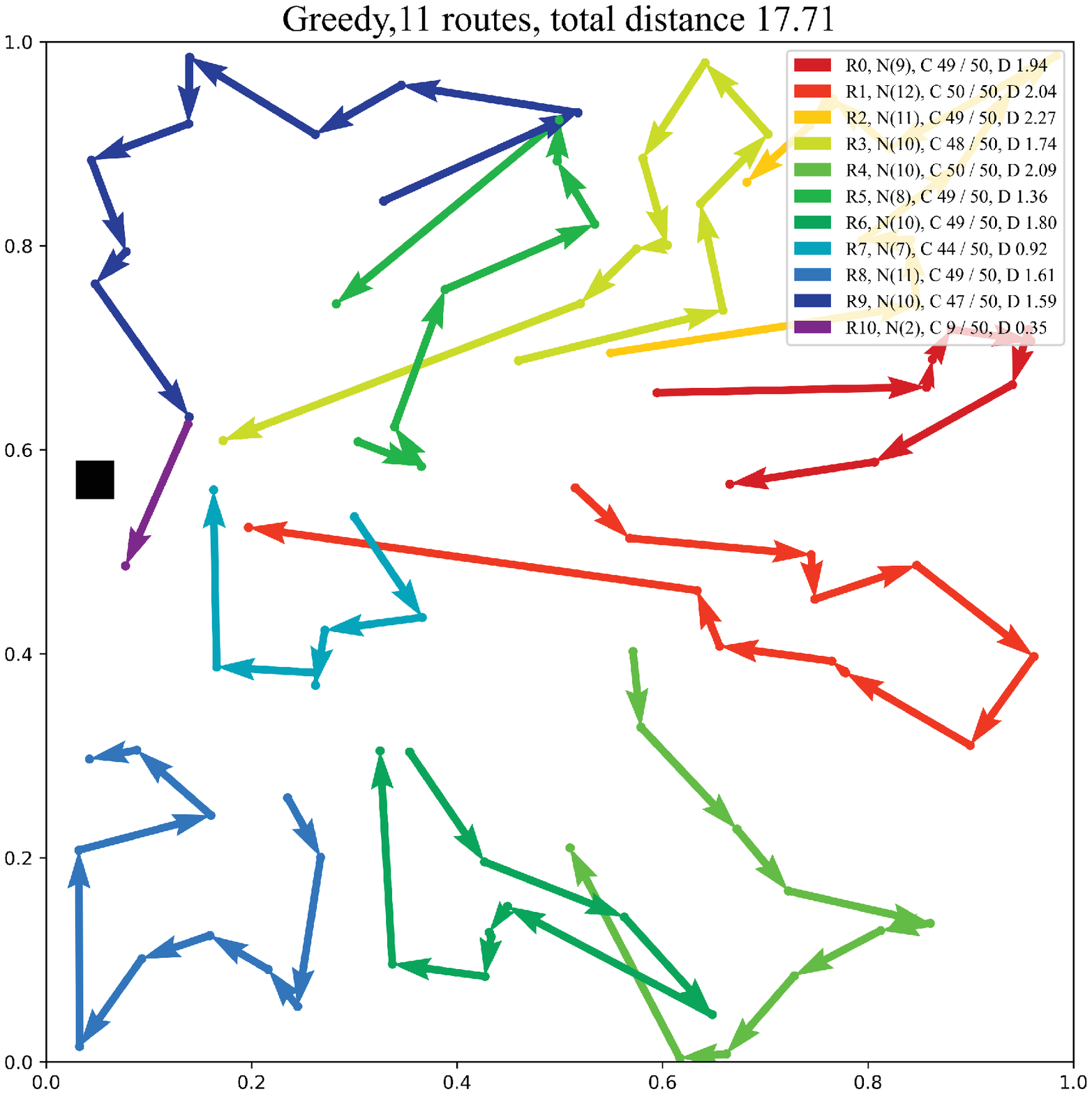}
		}
		\quad
		\subfigure{
			\label{fig:fig116}
			\includegraphics[width=4.2cm]{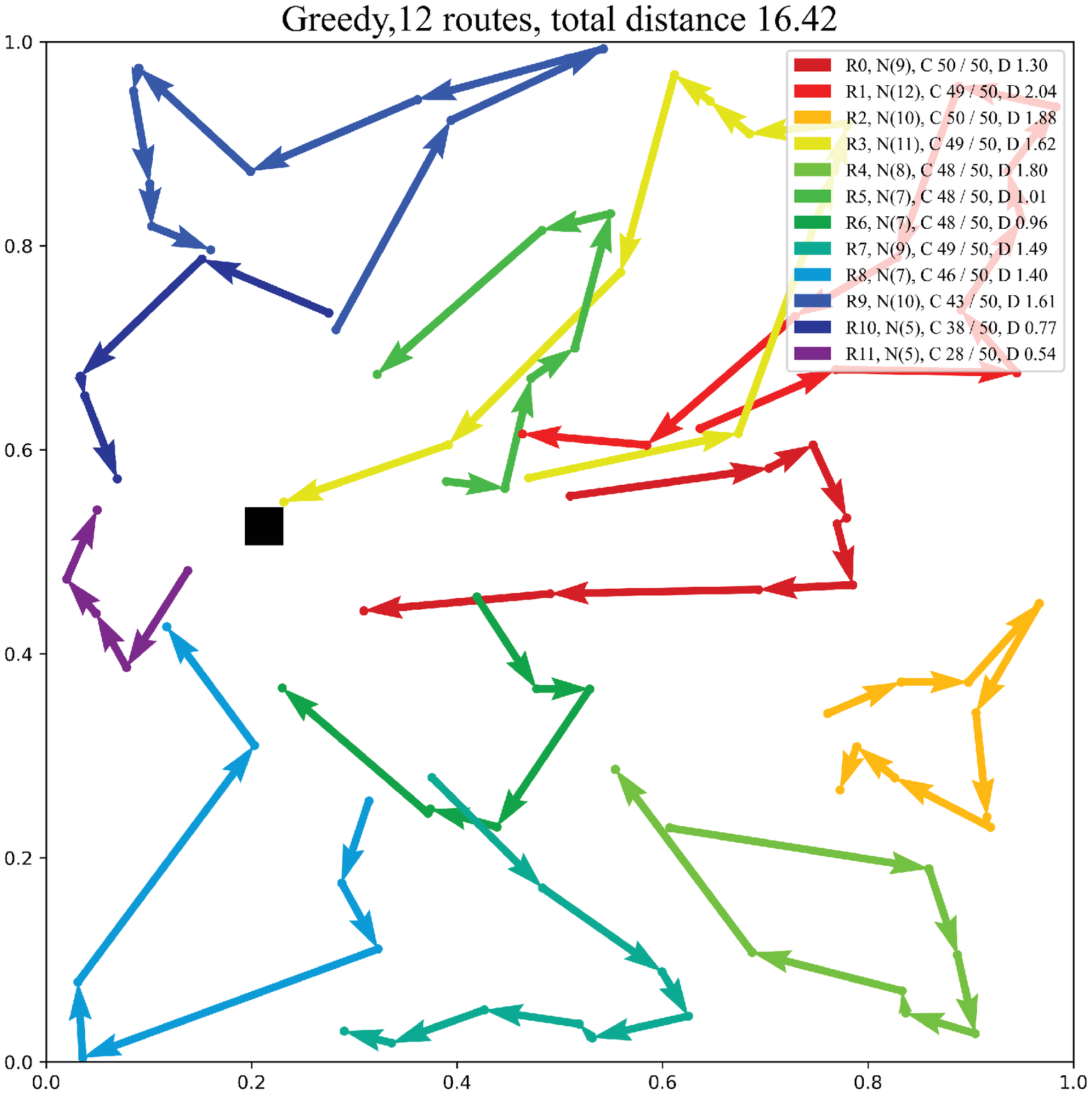}
		}
		\caption{Solving a random CVRP instance problem with 100 nodes through sampling and greedy decoding, where, R: Route; N: nodes included in the path; C: all demand of the path/vehicle capacity; and D: the length of the path. Edges from and to depot omitted for clarity.}
		\label{fig:fig11}
	\end{figure}

	Figure \ref{fig:fig10} and \ref{fig:fig11} show instance solutions for the TSP and the CVRP with 100 nodes obtained by our model compared with the optimal solutions. Our solutions were obtained by a single construction using the trained model with either greedy decoding or sampling 1280 solutions (following the existing studies \cite{R7,R18,R19,R20,R22}) with the best one reported.
	\section{Solutions of standard benchmark instances}
	\label{app:c}
\begin{figure}[htbp]
		\centering
		\subfigure{
			\label{fig:fig121}
			\includegraphics[width=2.85cm]{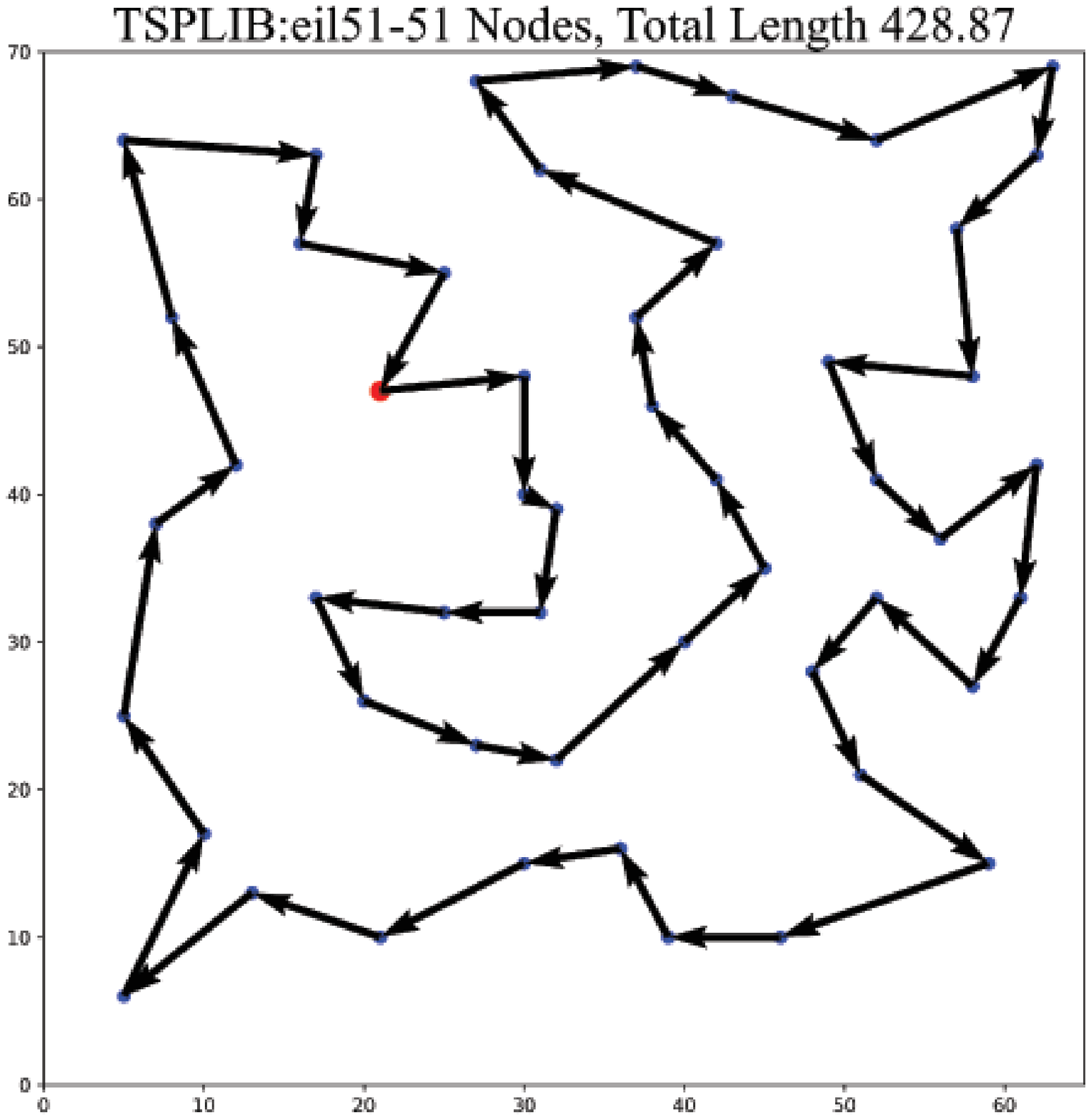}
		}
		\quad
		\subfigure{
			\label{fig:fig122}
			\includegraphics[width=3cm]{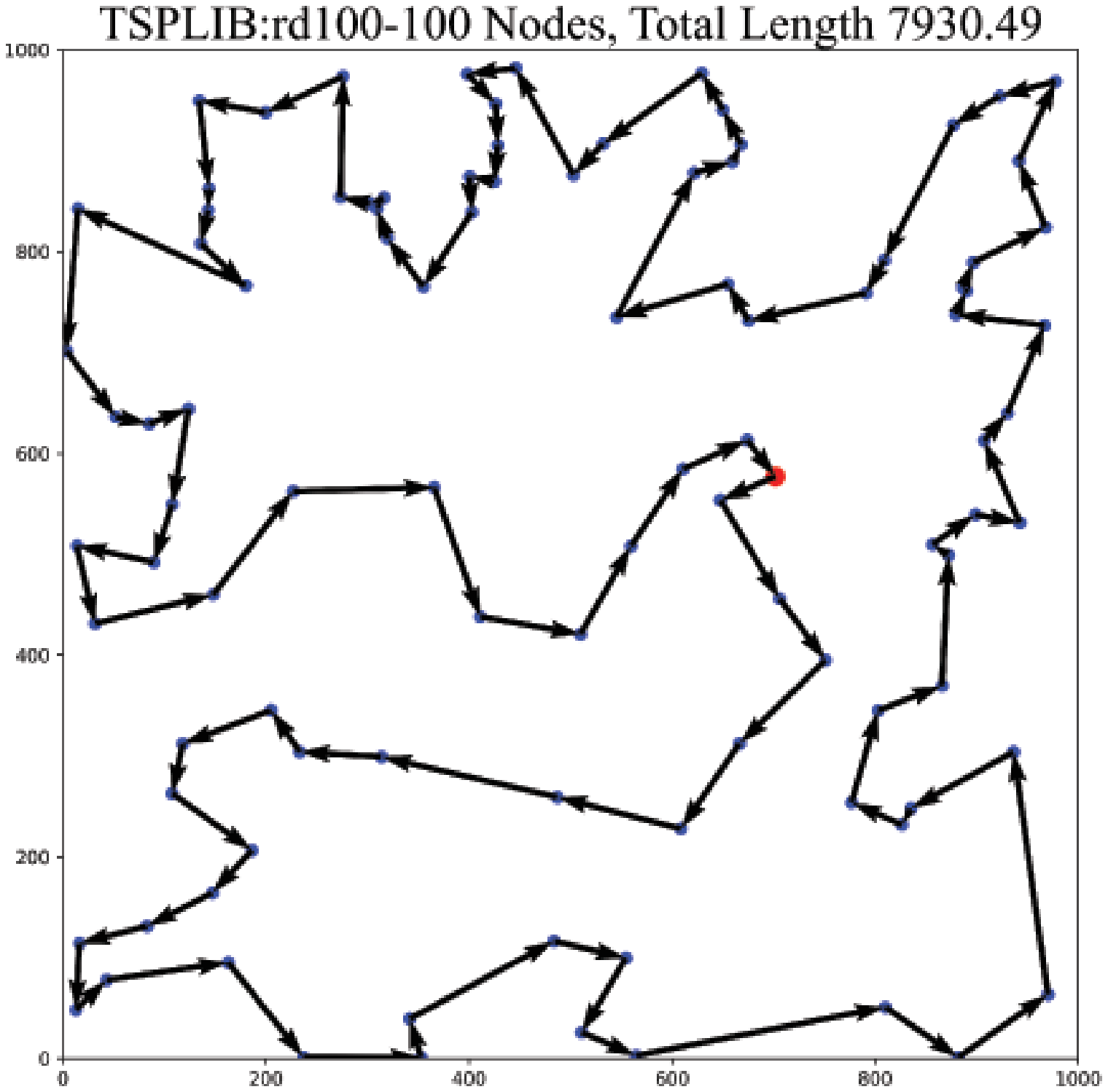}
		}
		\quad
		\subfigure{
			\label{fig:fig123}
			\includegraphics[width=3cm]{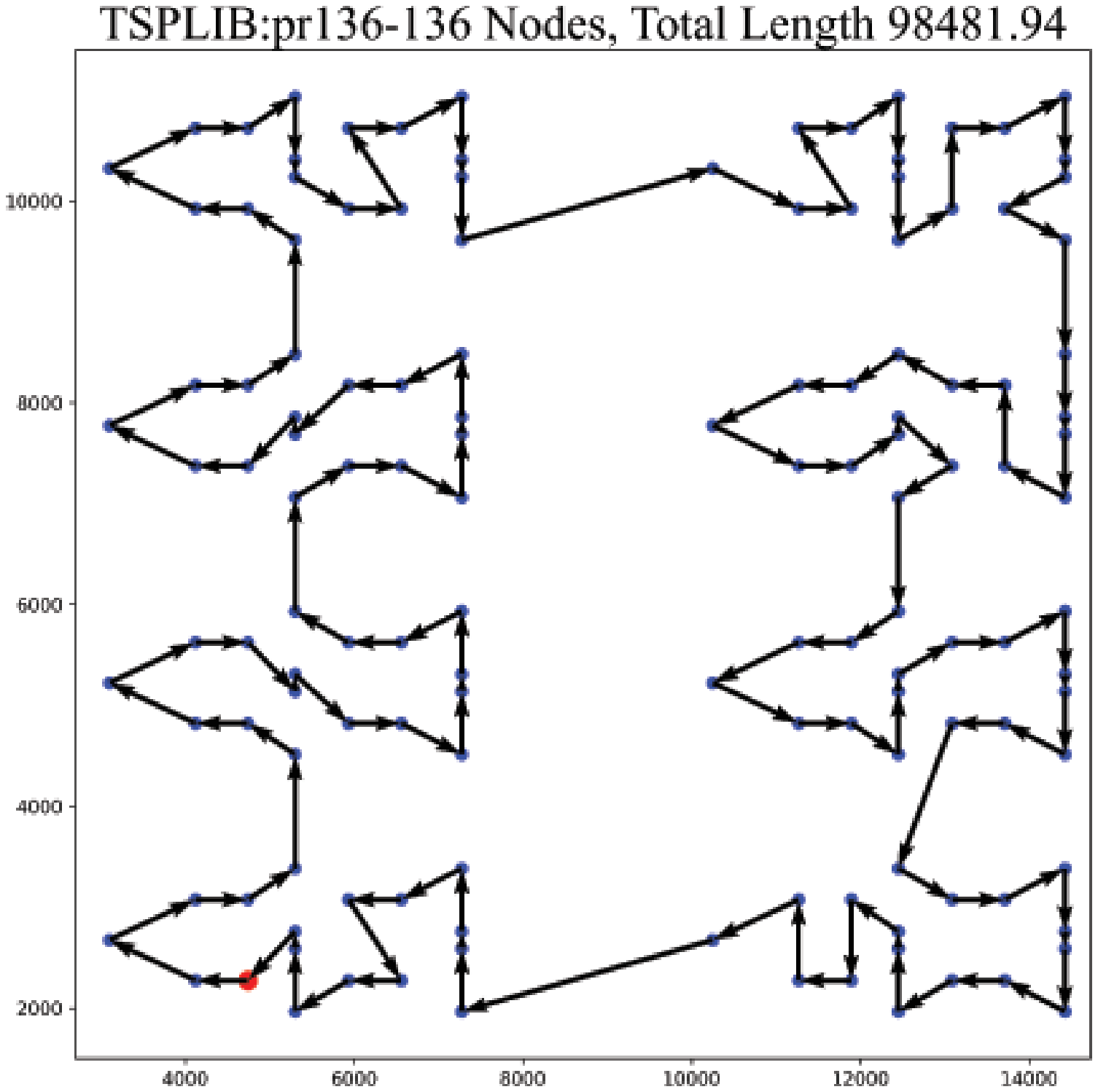}
		}
		\quad
		\subfigure{
			\label{fig:fig124}
			\includegraphics[width=3cm]{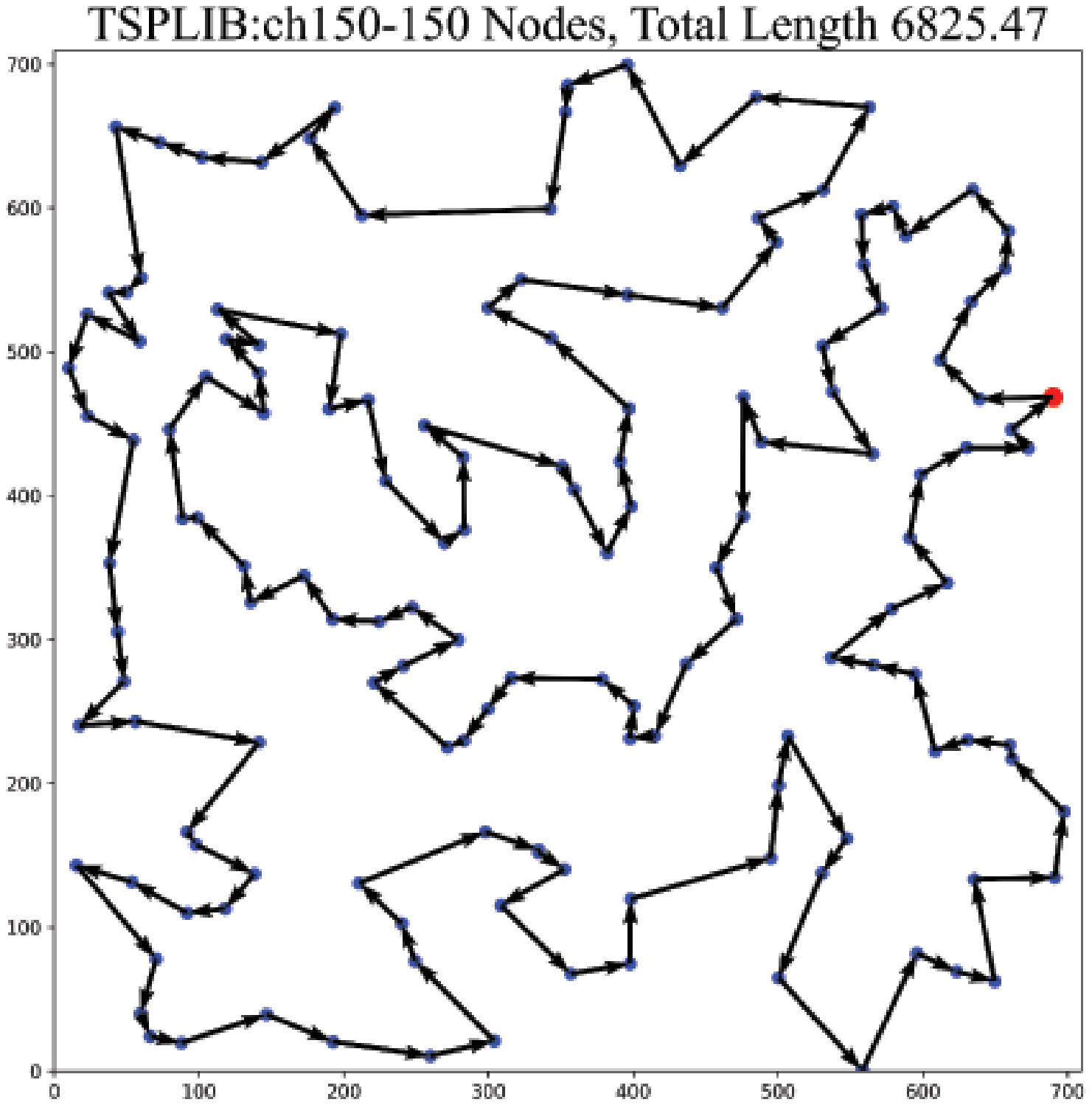}
		}
		\caption{Instances of TSPLIB solutions obtained by our method through \textit{trained-greedy100}.}
		\label{fig:fig12}
	\end{figure}

\begin{figure}[htbp]
		\centering
		\subfigure{
			\label{fig:fig131}
			\includegraphics[width=4.3cm]{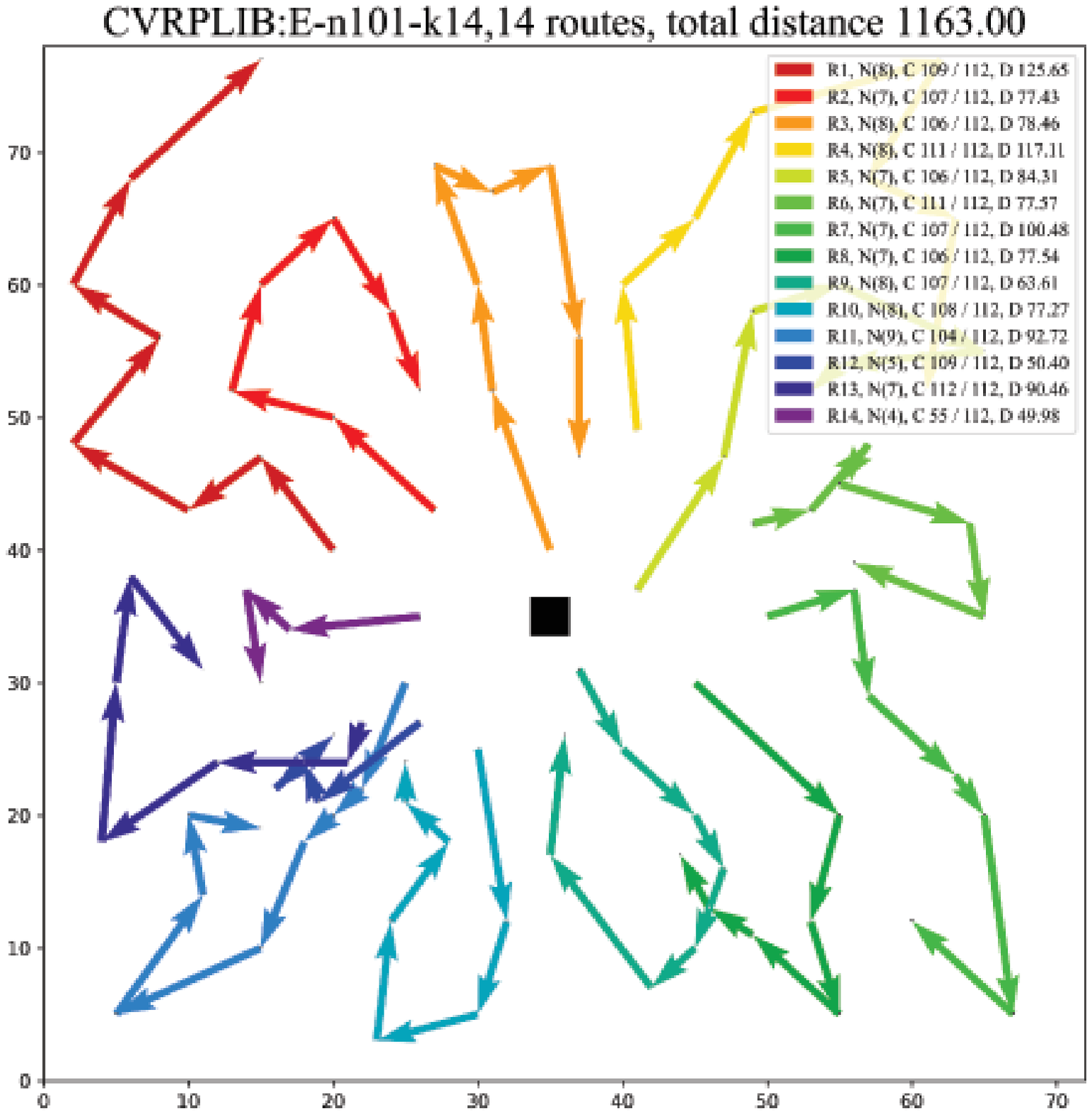}
		}
		\quad
		\subfigure{
			\label{fig:fig132}
			\includegraphics[width=4.3cm]{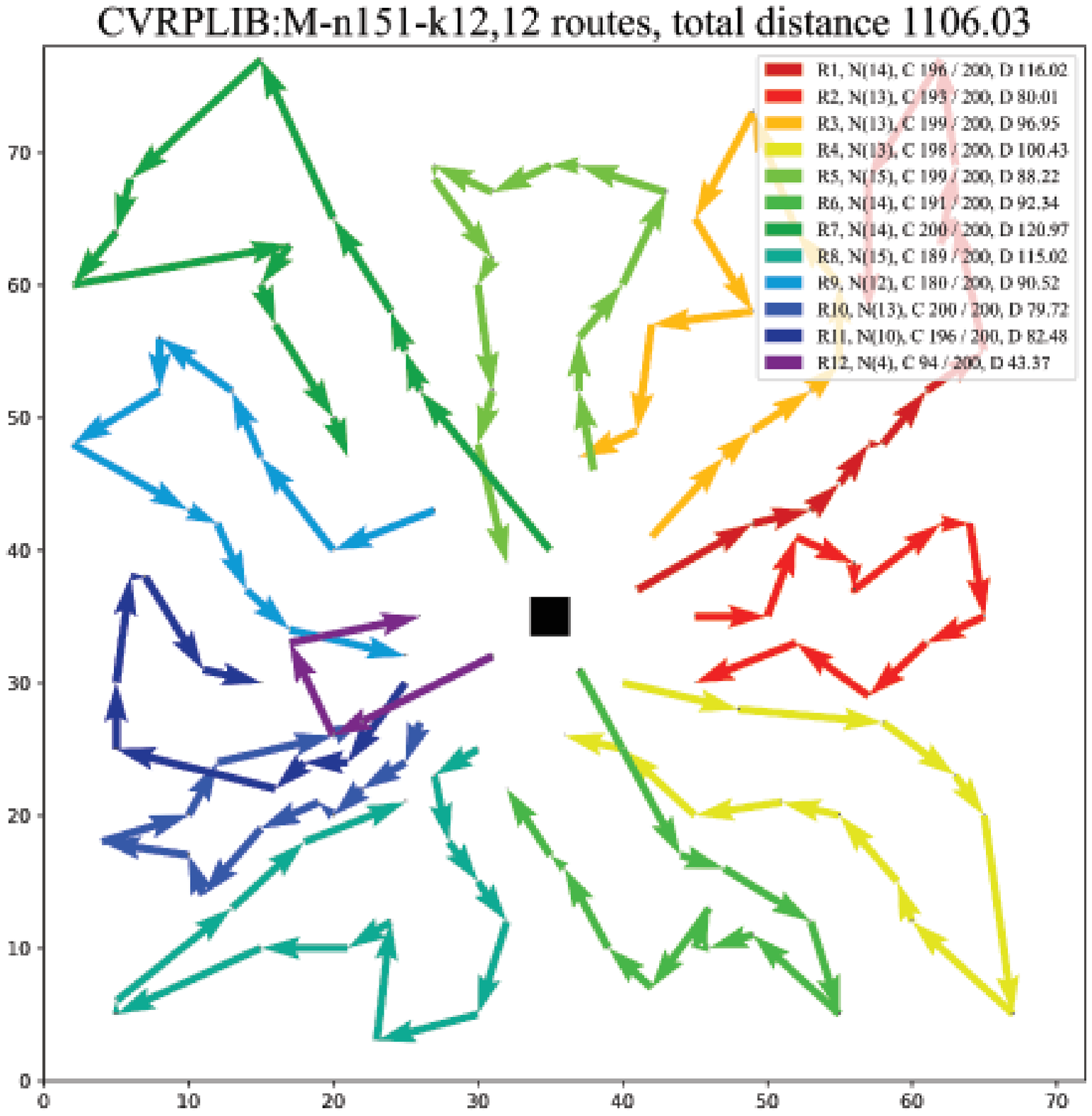}
		}
		\quad
		\subfigure{
			\label{fig:fig133}
			\includegraphics[width=4.3cm]{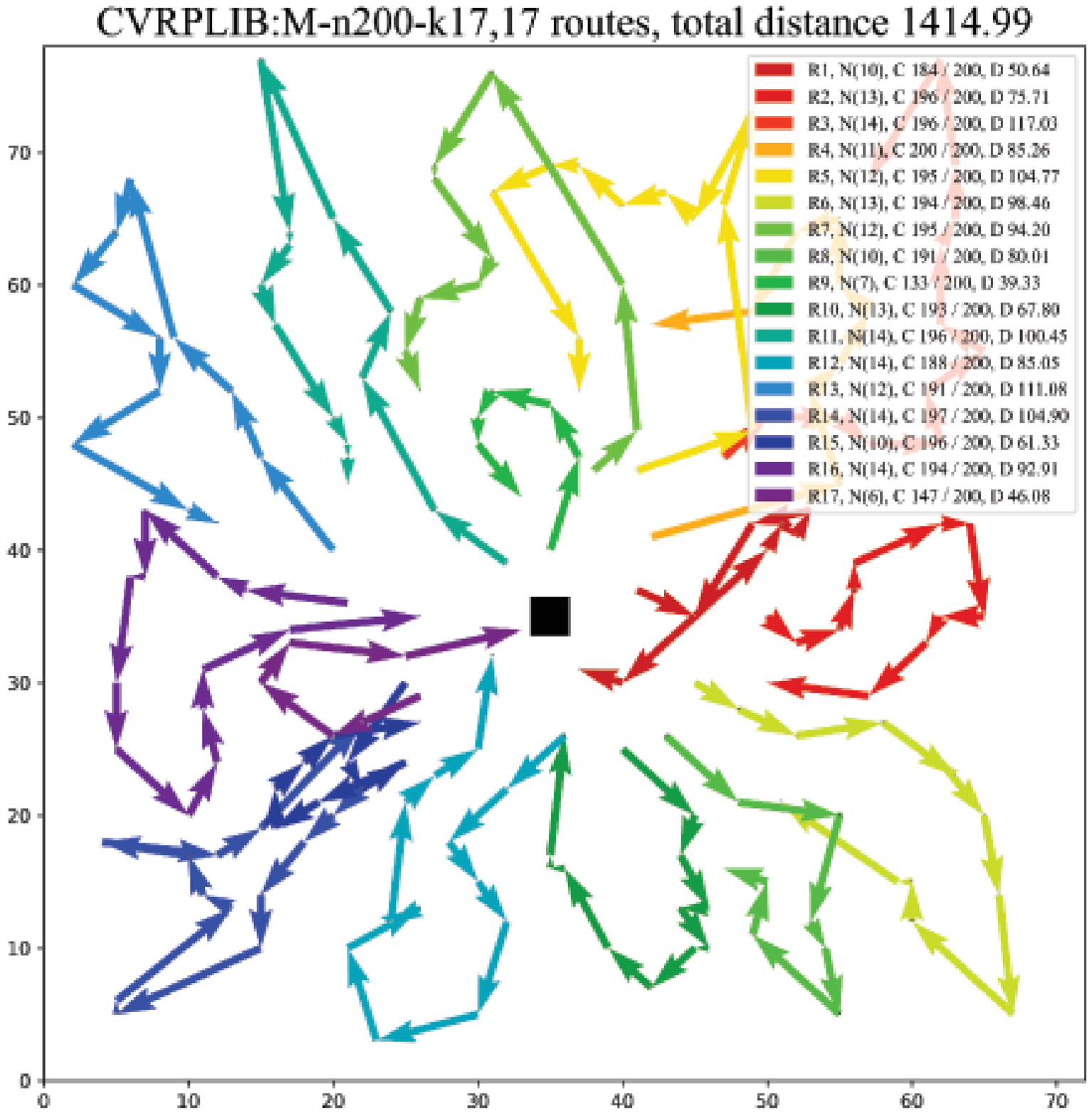}
		}
		\quad
		\subfigure{
			\label{fig:fig134}
			\includegraphics[width=4.3cm]{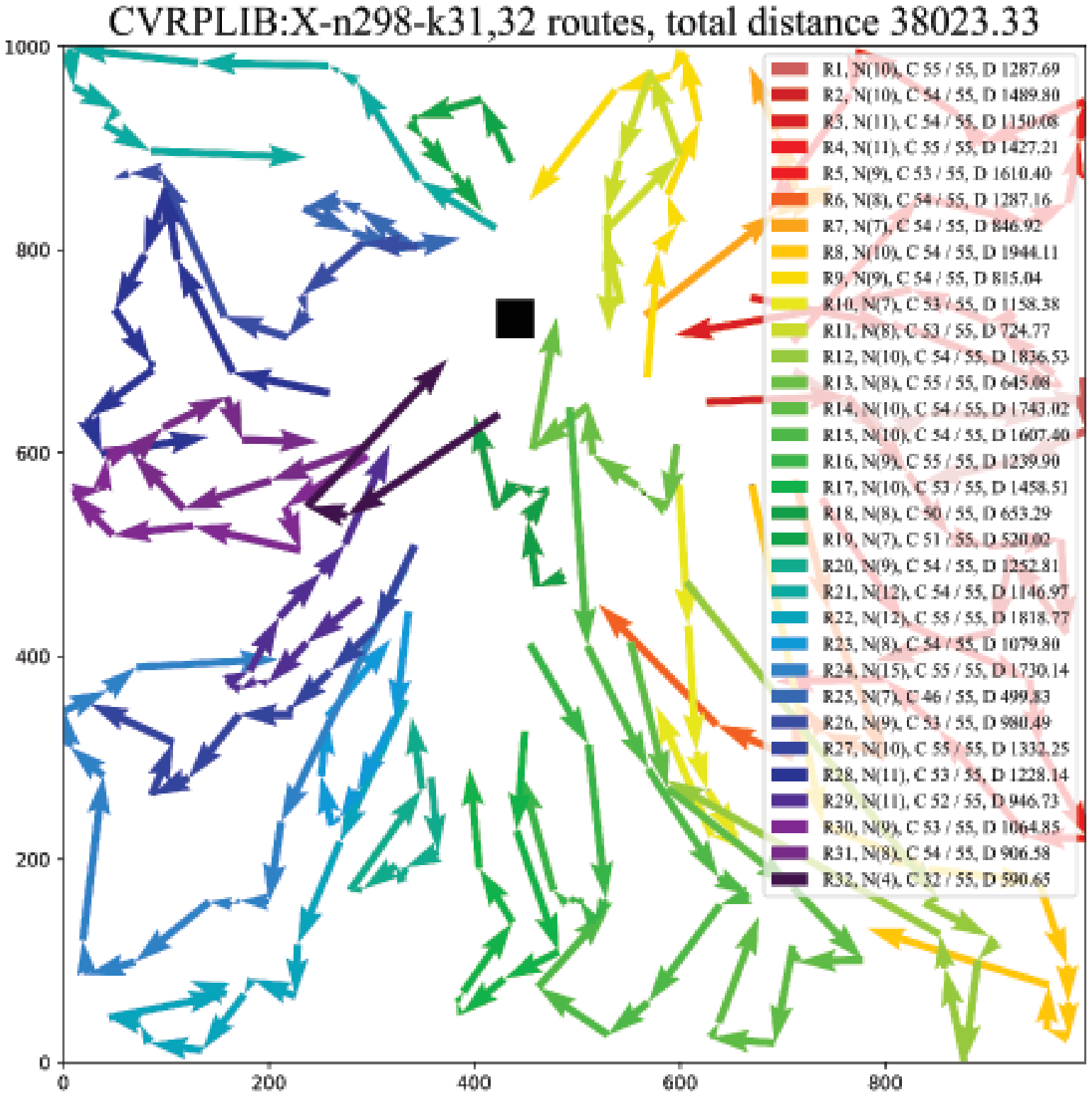}
		}
		\caption{Instances of CVRPLIB solutions obtained by our method through \textit{trained-greedy100}, where, R: Route; N: nodes included in the path; C: all demand of the path/vehicle capacity; and D: the length of the path. Edges from and to depot omitted for clarity.}
		\label{fig:fig13}
	\end{figure}

	Figure \ref{fig:fig12} and Figure \ref{fig:fig13} show the solution of the standard benchmark instance for the TSP and CVRP using our method. The solutions were obtained by \textit{trained-greedy100} as described in Section \ref{se:se5.6}.
\end{appendices}

\end{document}